\documentclass{article} 
\usepackage{iclr2027_conference,times}

\usepackage{amsmath,amsfonts,bm}

\def\eqref#1{equation~\ref{#1}}

\def\1{\bm{1}}

\DeclareMathAlphabet{\mathsfit}{\encodingdefault}{\sfdefault}{m}{sl}
\SetMathAlphabet{\mathsfit}{bold}{\encodingdefault}{\sfdefault}{bx}{n}

\usepackage{hyperref}
\usepackage{url}
\usepackage{soul}
\usepackage{xcolor}
\usepackage{graphicx}
\usepackage{booktabs}
\usepackage{subcaption}
\usepackage{pgfplots}
\usepackage[table,dvipsnames]{xcolor}
\usepackage{multicol}
\usepackage{multirow}
\usepackage{wrapfig}
\usepackage{longtable}

\usepackage[capitalise]{cleveref}

\usepackage[textsize=tiny]{todonotes}

\usepackage{tikz}
\usepackage{pgfplots}
\usetikzlibrary{patterns}
\pgfplotsset{compat=1.18}

\newcommand{\best}[1]{\textbf{#1}}
\newcommand{\second}[1]{\underline{#1}}
\newcommand{\gain}[1]{\textbf{\textcolor{Green}{+#1\%}}}
\newcommand{\drop}[1]{\textbf{\textcolor{BrickRed}{-#1\%}}}
\newcommand{\same}[1]{\textbf{#1\%}}
\definecolor{revblue}{RGB}{0,0,190}

\usepackage{adjustbox}

\title{FlowTool: Controlling Tool Parameter in Image Retouching via Flow Matching}

\author{%
\textbf{Thanh-Long V. Le}\textnormal{\textsuperscript{1,2}\thanks{Work done during Thanh-Long's internship at Adobe Research.}}\quad
\textbf{Steven Walton}\textnormal{\textsuperscript{1}}\quad
\textbf{Seunghyun Yoon}\textnormal{\textsuperscript{1}}\quad
\textbf{Branislav Kveton}\textnormal{\textsuperscript{1}}\\[3pt]
\textbf{Trung Bui}\textnormal{\textsuperscript{1}}\quad
\textbf{Eunho Yang}\textnormal{\textsuperscript{2}}\quad
\textbf{Viet Lai}\textnormal{\textsuperscript{1}\thanks{Project Lead. Correspondence to: \texttt{daclai@adobe.com}}}\\[3pt]
\textsuperscript{1}\,Adobe Research\quad
\textsuperscript{2}\,KAIST AI
}

\iclrfinalcopy 
\begin{document}

\maketitle

\begin{abstract}
Tool-based image editing (image retouching) is commonly formulated with autoregressive multimodal large language models (MLLMs) that sequentially generate reasoning, tool selections, and parameter values. In this work, we present a novel approach to tool-based image editing by framing the task as a flow matching problem. We introduce FlowTool, a framework that directly models the distribution of high-quality tool parameters conditioned on the input image and user instruction using conditional rectified flow. FlowTool combines a vision-language model backbone for multimodal understanding with a Diffusion Transformer parameter generator that transforms Gaussian noise into an editing plan. We train FlowTool with a two-stage supervised flow-matching curriculum, followed by reward-based post-training. Across four benchmarks, FlowTool significantly outperforms specialized MLLM agents and remains competitive with proprietary MLLMs, while reducing latency by at least $50\times$ and using nearly half the memory of the evaluated local baselines.
These results demonstrate that tool-based image editing can be effectively modeled as conditional generation over structured continuous editing parameters, without autoregressive reasoning.
\end{abstract}

\section{Introduction}

Professional tool-based image editing software provides a rich collection of tools for precise and controllable visual manipulation~\citep{retouchiq2026}. However, effectively using these tools often requires substantial domain expertise. Users must translate high-level editing intents, such as \emph{make the image more cinematic''} or \emph{brighten the subject while preserving the background,''} into concrete editing plans: selecting appropriate tools (e.g., saturation, hue, and opacity) and determining their parameter values. 
Recent advances in MLLMs ~\citep{claude,gpt5.6,comanici2025gemini,wang2025internvl3,qwen3.5,guo2025deepseek} have enabled a promising approach to solving the image retouching (IR) problem. Given an image and an instruction, an MLLM generates reasoning trace, tool selections, and numerical parameter values as sequences of discrete tokens~\citep{lin2026jarvisart,dutt2025monetgpt} in an interactive environment~\citep{yao2022react}. 

While this formulation is technically convenient, we argue that MLLMs is a fundamental mismatch for the IR problem for 3 reasons: 
(1) First, IR tools are usually standardized with smooth and continuous linearized scalers for parameters. MLLMs do not yield a high-fidelity, continuous numerical representation sufficient for the sequential decoding of discrete tokens down to numbers \citep{song2025decoding}.
This leads to poor numerical understanding and generation \citep{lovering2025language,golkar2023xval,ni2026numeracy}, which is unfit for IR.
(2) An MLLM predicts an editing plan token by token, effectively modeling each tool or parameter value conditioned on all preceding actions and parameter predictions \cite{vaswani2017attention}. This autoregressive formulation exposes the IR task to cascading errors, as mistakes made early in a long prediction trajectory can propagate and compound over subsequent steps~\citep{zhang2024snowball}. We therefore argue that the desired output is better viewed not as a single deterministic sequence of tokens, but as a sample from a conditional joint distribution over coherent editing actions~\citep{chi2025diffusion}.
(3) Autoregressive agent formulations additionally incur a substantial inference cost. 
They may generate unnecessarily lengthy reasoning and tool-call trajectories, even though these tokens are not the system's final objective. 
This leads to high latency and memory overhead~\citep{Gagrani_2024_CVPR,leviathan2023fast}. 
This limitation are particularly consequential for interactive editing, where users expect near-real-time feedback, and for deployment on resource-constrained devices.

\begin{figure}[t]
\centering
    \fbox{
    \includegraphics[width=0.97\textwidth]{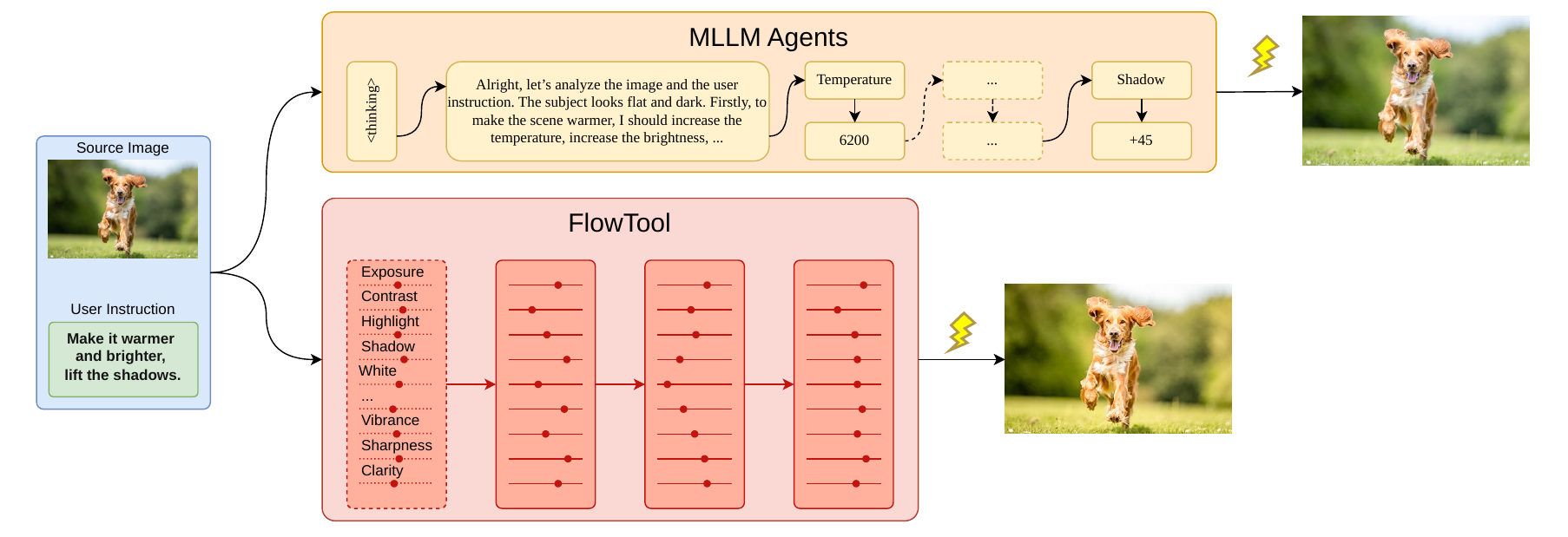}
    }
    \caption{\textbf{FlowTool versus autoregressive agents.} Existing MLLM agents formulate retouching as language generation, autoregressively producing reasoning, tool selections, and discretized numerical parameters. FlowTool instead treats retouching as conditional generation within a structured, continuous parameter space and directly generates tool parameters via conditional flow matching.}
    \vspace{-10pt}
\label{fig:overall}
\end{figure}

Motivated by these observations, we formulate tool-based image retouching as a \emph{conditional generative modeling} problem~\citep{liu2022rectifiedflow,lipman2023flowmatching}. Given an input image $\mathbf{x}$ and user instruction $\mathbf{c}$, let $\mathbf{a(\cdot)}$ denote an editing plan consisting of the selected tools and their parameters. Rather than representing $\mathbf{a(\cdot)}$ as a language sequence, we directly model the conditional distribution
$p(\mathbf{a}\mid\mathbf{x},\mathbf{c})$ \citep{chi2025diffusion,black2024pi_0}.
This formulation preserves the continuous structure of tool parameters, naturally models the conditional joint distribution over the tool parameter space, and eliminates the need for natural-language reasoning and numerical generation during inference.
To this end, we introduce \textbf{FlowTool}, a framework that formulates tool-based image editing as conditional flow matching~\citep{tong2024improving}. 

FlowTool combines a VLM backbone for multimodal understanding with a DiT-based tool parameter generator \citep{peebles2023scalable} for continuous parameter value generation and a tool-presence head for tool selection. We train the model with a two-stage supervised flow-matching curriculum, followed by reward-based post-training \citep{zheng2025diffusionnft} that directly optimizes the quality of rendered edits. Across four benchmarks, MMArt-Bench \citep{lin2026jarvisart}, ArtEdit-Bench \citep{jarvisevo2026}, MIT-Adobe5K \citep{bychkovsky2011fivek}, and FlowTool-Eval, FlowTool significantly outperforms specialized MLLM agents and proprietary MLLMs on reference-based metrics. Under reference-free evaluation of semantic consistency and perceptual quality, FlowTool also consistently outperforms specialized MLLM agents while remaining competitive with proprietary models. Beyond editing quality, directly generating editing parameters yields substantial computational benefits: FlowTool reduces inference latency by at least $50\times$ and requires nearly $2\times$ less memory than autoregressive baselines. These gains make high-quality tool-based image editing practical in interactive environments and on resource-constrained devices.

We make three key contributions as follows: 
\begin{itemize}
    \item First, we recast tool-based image editing (image retouching) from autoregressive language generation into conditional generative modeling over a structured continuous tool parameter space, better matching the continuous, precision-sensitive, and multimodal nature of professional editing parameters. 
    \item Second, we introduce FlowTool, a conditional flow-matching framework that directly generates tool parameters from multimodal image--instruction representations and further improves rendered outcomes through reward-based post-training. \item Third, we conducted extensive evaluation across four benchmarks, demonstrating that FlowTool achieves strong editing quality while dramatically reducing inference latency and memory consumption compared with autoregressive MLLM agents.
\end{itemize}

\section{Related Work}

\textbf{Tool-based image editing.} Early work in tool-based editing predicting editing action to satisfy a general audience without taking into account user intents \citep{hu2018exposure,ke2022harmonizer,ouyang2023rsfnet}. 
Recent advancements in MLLMs have led to models that predict editing actions by analyzing user intent \citep{dutt2025monetgpt,lin2026jarvisart}, critiquing images \citep{retouchiq2026}, and performing iterative editing \citep{jarvisevo2026}. 
These methods adopt an autoregressive model to generate both reasoning, tool sequences, and parameter values token-by-token in a discrete space. 
The development of diffusion models has led to various diffusion-based image editing methods \citep{duan2025diffretouch,hertz2022prompt,liu2025mofrr,wu2023uncovering}. 
While these models can handle a broad range of edits, they are highly compute-intensive and often struggle to preserve subject identity \cite{shi2024dragdiffusion}.
Our work models the editing action space jointly in continuous numerical space through flow matching.

\textbf{Vision-language-action models.}
Vision-language-action (VLA) models extend pretrained multimodal representations with action prediction for language-conditioned control~\citep{pmlr-v229-zitkovich23a,kim2024openvla,liu2025hybridvla,wang2026unified,black2024pi_0,wang2026qwen,li2024cogact,bjorck2025gr00t}. In parallel, diffusion-based decision models have established generative action and trajectory modeling as an alternative to autoregressive prediction. Early work models entire trajectories as denoising targets~\citep{janner2022planning}, with later extensions introducing conditional guidance and inverse-dynamics action recovery~\citep{ajay2022conditional}. More recently, hierarchical diffusion uses language-aligned discrete skill abstractions to condition continuous trajectory generation~\citep{liang2024skilldiffuser}, while recent work explores discrete diffusion over tokenized action chunks for non-autoregressive VLA decoding~\citep{liang2025discrete}.
FlowTool shares the general principle of conditioning a continuous generator on multimodal representations, but applies it to a fundamentally different problem: generating structured image-editing parameters rather than physical control trajectories.

\section{Preliminaries}
\label{sec:prelim_flow_matching}

{\bf Flow Matching.} Flow matching learns a time-dependent vector field that transports samples between data distribution $p_0$ and standard Gaussian prior $p_1=\mathcal{N}(\mathbf{0},\mathbf{I})$ through the Ordinary Differential Equation (ODE) $\frac{d\mathbf{a}_t}{dt}=\mathbf{v}_{\theta}(\mathbf{a}_t,t)$~\citep{lipman2023flowmatching}.
Under the rectified-flow formulation~\citep{liu2022rectifiedflow}, data sample $\mathbf{a}_0\sim p_0$ and Gaussian sample $\mathbf{a}_1\sim p_1$ are connected by the linear path:
\begin{equation}
\mathbf{a}_t=(1-t)\mathbf{a}_0+t\mathbf{a}_1,
\qquad
\mathbf{u}_t=\frac{d\mathbf{a}_t}{dt}
=\mathbf{a}_1-\mathbf{a}_0.
\label{eq:rectified_path}
\end{equation}
The velocity predictor is trained to match this path velocity for $t$ sampled uniformly from $[0,1]$:
\begin{equation}
    \mathcal{L}_{\mathrm{FM}}(\theta)
    =
    \mathbb{E}_{\substack{
        \mathbf{a}_0\sim p_0,
        \mathbf{a}_1\sim p_1,
        t\sim\mathcal{U}[0,1]
    }}
    \left\|
    \mathbf{v}_{\theta}(\mathbf{a}_t,t)
    -
    (\mathbf{a}_1-\mathbf{a}_0)
    \right\|_2^2.
    \label{eq:conditional_fm_loss}
\end{equation} 
At inference time, sampling starts from
$\mathbf{a}_1\sim\mathcal{N}(\mathbf{0},\mathbf{I})$
and integrates the learned ODE from $t=1$ to $t=0$ to recover a sample from the data distribution. In FlowTool, $\mathbf{a}$ corresponds to the continuous tool-parameter representation, and the velocity field is additionally conditioned on the input image and user instruction, as described in \cref{sec:method}.

{\bf Reinforcement Learning for Flow Models.} DiffusionNFT~\citep{zheng2025diffusionnft} performs reward-based post-training through the forward flow-matching process. For condition $\mathbf{y}$, let
$\pi_{\mathrm{old}}(\cdot\mid\mathbf{y})$ denote the rollout policy before the current policy update, and let
$\hat{\mathbf{a}}_0^{(k)}\sim\pi_{\mathrm{old}}(\cdot\mid\mathbf{y})$
denote the $k$-th rollout sampled from this policy, with reward $R^{(k)}$.
Given $K$ rollouts and the reward-normalization factor $Z_{\mathbf{y}}>0$, the reward is converted to an optimality weight
$r^{(k)}\in[0,1]$ as:
\begin{equation}
r^{(k)}
=
\frac{1}{2}
+
\frac{1}{2}
\operatorname{clip}
\left(
\frac{R^{(k)}-\bar{R}}{Z_{\mathbf{y}}},
-1,1
\right),
\qquad
\bar{R}=\frac{1}{K}\sum_{j=1}^{K}R^{(j)}.
\label{eq:nft_optimality}
\end{equation}

For each rollout, we sample
$\mathbf{a}_1\sim\mathcal{N}(\mathbf{0},\mathbf{I})$ and
$t\sim\mathcal{U}[0,1]$, and construct
$\hat{\mathbf{a}}_t=(1-t)\hat{\mathbf{a}}_0+t\mathbf{a}_1$
with target velocity
$\hat{\mathbf{u}}_t=\mathbf{a}_1-\hat{\mathbf{a}}_0$.
DiffusionNFT defines
\begin{equation}
\begin{aligned}
\mathbf{v}^{+}_{\theta}(\hat{\mathbf{a}}_t,t\mid\mathbf{y})
&=(1-\beta)\mathbf{v}_{\mathrm{old}}(\hat{\mathbf{a}}_t,t\mid\mathbf{y})
+\beta\mathbf{v}_{\theta}(\hat{\mathbf{a}}_t,t\mid\mathbf{y}),\\
\mathbf{v}^{-}_{\theta}(\hat{\mathbf{a}}_t,t\mid\mathbf{y})
&=(1+\beta)\mathbf{v}_{\mathrm{old}}(\hat{\mathbf{a}}_t,t\mid\mathbf{y})
-\beta\mathbf{v}_{\theta}(\hat{\mathbf{a}}_t,t\mid\mathbf{y}),
\end{aligned}
\label{eq:nft_implicit_policies}
\end{equation}

and optimizes
\begin{equation}
    \mathcal{L}_{\mathrm{NFT}}
    =
    \mathbb{E}_{\substack{
    \mathbf{y},\,\hat{\mathbf{a}}_0\sim\pi_{\mathrm{old}},
    \mathbf{a}_1\sim\mathcal{N}(\mathbf{0},\mathbf{I}),\,t\sim\mathcal{U}[0,1]
    }}
    \left[
    r\|\mathbf{v}^{+}_{\theta}-\hat{\mathbf{u}}_t\|_2^2
    +
    (1-r)\|\mathbf{v}^{-}_{\theta}-\hat{\mathbf{u}}_t\|_2^2
    \right].
\label{eq:nft_loss}
\end{equation}
The positive and negative branches, respectively, encourage high-reward rollouts and discourage low-reward ones.
In FlowTool, $\mathbf{y}$ corresponds to the source image and user instruction, and the reward is computed from the image rendered using the predicted editing plan (See \cref{sec:rl}).

\section{Methodology}
\label{sec:method}
\subsection{Problem Formulation}
\label{sec:problem_formulation}

We consider image-editing renderer \(\mathcal{R}\) equipped with a collection of editing tools whose behavior is controlled by continuous parameters. Following previous works, we assume that the renderer applies a complex plan with multiple tools using its own static optimal tool order \citep{lin2026jarvisart,retouchiq2026}.
The continuous parameter space is defined as:
$\mathcal{T} = [\ell_1,u_1]\times\cdots\times[\ell_D,u_D]$ where $D$ is the number of continuous tool parameters, and $[\ell_d,u_d]$ specifies the valid \emph{native} range of the $d$-th parameter. Thus, an element \(T_i\in\mathcal{T}\) specifies a valid configuration of the renderer's tools and their associated parameter values.

Given source image $\mathbf{x}$ and natural-language editing instruction $\mathbf{c}$, our goal is to generate an editing plan that can be executed by $\mathcal{R}$ to produce an output image that satisfies the user's intent. The editing plan determines which tools to invoke and specifies their corresponding parameters within $\mathcal{T}$, conditioned on both the visual content of $\mathbf{x}$ and the editing intent expressed by $\mathbf{c}$.

We normalize the $i$-th parameter value $a^i_\mathrm{raw}$ to [-1,1] range and represent the 
editing plan as:
\begin{equation}
    \mathbf{a}=[a^{1},\ldots,a^{D}]^\top\in[-1,1]^D, \hspace{1cm}
    a^{i}=
    2\frac{a^i_\mathrm{raw}-\ell_i}{u_i-\ell_i}-1.
    \label{eq:action_vector}
\end{equation}

Default values for all the tools are also normalized as $\mathbf{d}^{norm}$. 
Since an editing plan typically activates only a subset of the available tools, we associate $\mathbf{a}$ with tool-presence mask
$\mathbf{m}\in\{0,1\}^{D}$, where $m^i=1$ indicates that the $i$-th tool is active.
The corresponding editing plan is:
\begin{equation}
    \mathcal{P}(\mathbf{a},\mathbf{m})
    =
    \text{denorm}(\mathbf{m} \odot \mathbf{a} + (1-\mathbf{m}) \odot \mathbf{d}^{norm}),
    \label{eq:executable_plan}
\end{equation}
where the element-wise operator $\operatorname{denorm}_i$ maps the normalized $a^i$ back to its native range $[\ell_i,u_i]$.

Let $\mathbf{y}=(\mathbf{x},\mathbf{c})$ be the multimodal condition. 
Because multiple parameter configurations may produce valid edits for the same image-instruction pair, we model the conditional distribution
$p_0(\mathbf{a}_0\mid\mathbf{y})$
rather than regressing to a single deterministic solution.

For
$\mathbf{a}_0\sim p_0(\cdot\mid\mathbf{y})$ and Gaussian noise
$\boldsymbol{\mathbf{a}_1}\sim\mathcal{N}(\mathbf{0},\mathbf{I})$, we define
\begin{equation}
\mathbf{a}_t
=
(1-t)\mathbf{a}_0+t\boldsymbol{\mathbf{a}_1},
\qquad t\in[0,1],
\label{eq:action_flow_path}
\end{equation}
with target velocity $\boldsymbol{\mathbf{a}_1}-\mathbf{a}_0$.
FlowTool learns the conditional velocity field
$\mathbf{v}_{\theta}(\mathbf{a}_t,t\mid\mathbf{y})$
over the continuous tool parameter space, while a dedicated
\emph{tool-presence head} predicts the tool-presence mask
$\mathbf{m}$ conditioned on the same multimodal representation.

At inference time, we sample
$\mathbf{a}_1\sim\mathcal{N}(\mathbf{0},\mathbf{I})$
and integrate the learned ODE from $t=1$ to $t=0$ to obtain
$\hat{\mathbf{a}}_0$. The tool-presence head predicts
$\hat{\mathbf{m}}$, and together
$(\hat{\mathbf{a}}_0,\hat{\mathbf{m}})$ defines the editing plan
$\mathcal{P}(\hat{\mathbf{a}}_0,\hat{\mathbf{m}})$, and the final image is produced as
$
\hat{\mathbf{x}}
=
\mathcal{R}\!\left(
\mathbf{x},
\mathcal{P}(\hat{\mathbf{a}}_0,\hat{\mathbf{m}})
\right).
\label{eq:rendered_output}
$
Thus, FlowTool directly generates a structured editing plan in continuous parameter space rather than serializing tool selections, parameter values, and intermediate reasoning into an autoregressive language sequence. 

\subsection{Model Architecture}
\label{sec:architecture}
\begin{figure*}[h]
    \centering
    \includegraphics[width=0.95\textwidth]{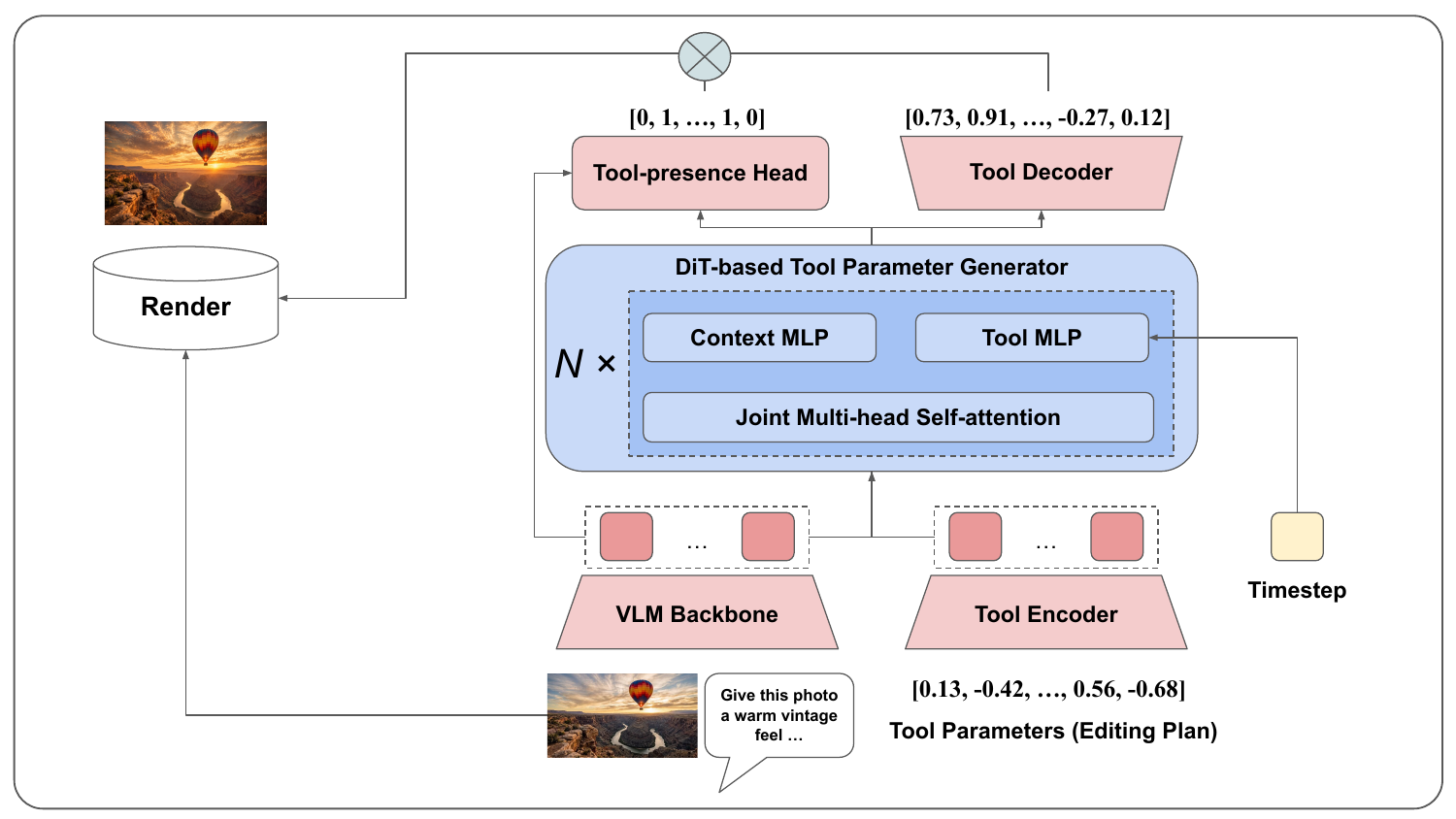}
    \caption{
    \textbf{Overview of the FlowTool architecture.}
    A vision-language model (VLM) encodes the input image and editing instruction into multimodal conditioning features, while a DiT-based tool parameter generator transforms noisy tool parameters into the final editing plan through conditional rectified flow. The resulting tool parameters are then executed by the image-editing engine.
    }
    \label{fig:flowtool_architecture}
\end{figure*}
As shown in \cref{fig:flowtool_architecture}, FlowTool consists of three main components: a vision-language model (VLM) backbone for multimodal understanding, a DiT-based tool parameter generator for continuous parameter prediction, and a tool-presence head for determining which tool (editing parameters) should be activated. Together, these components map an input image-instruction pair to an editing plan.

\textbf{Vision-Language Backbone.}
Given source image $\mathbf{x}$ and editing instruction $\mathbf{c}$, the VLM produces multimodal tokens $\mathbf{H}=\operatorname{VLM}(\mathbf{x},\mathbf{c})$,
which encode both the visual information of the source image and the editing intent. This multimodal representation provides the condition for both parameter generation and tool-presence prediction.

\textbf{DiT-based Tool Parameter Generator.}
The parameter generator models the conditional flow over the continuous editing parameters. Given noisy parameter vector $\mathbf{a}_t\in\mathbb{R}^{D}$, each scalar parameter is projected into the DiT hidden space to form a sequence of parameter representations, $\mathbf{P}_t=\operatorname{Embed}_{\mathrm{param}}(\mathbf{a}_t)$.
Multimodal tokens $\mathbf{H}$ and parameter representations $\mathbf{P}_t$ are concatenated and processed by a stack of two-stream DiT blocks,
$
    \left(
    \mathbf{H}^{\ell+1},
    \mathbf{P}_t^{\ell+1}
    \right)
    =
    \mathcal{B}_{\ell}
    \left(
    \mathbf{H}^{\ell},
    \mathbf{P}_t^{\ell},
    t
    \right),
$
where joint attention allows each parameter representation to condition on both the multimodal context and the remaining editing parameters, while the flow timestep $t$ is injected through adaptive normalization. The final parameter representations are projected to the conditional velocity field:
\begin{equation}
    \mathbf{v}_{\theta}
    (\mathbf{a}_t,t\mid\mathbf{x},\mathbf{c})
    =
    \operatorname{Head}_{\mathrm{vel}}
    \left(
    \mathbf{P}_t^{N}
    \right)
    \in\mathbb{R}^{D}.
\end{equation}
Integrating this velocity field from $t=1$ to $t=0$ transforms Gaussian noise into the clean tool parameters $\hat{\mathbf{a}}_0$.

\textbf{Tool-Presence Head.}
Since a user instruction typically activates only a subset of the available tools, FlowTool additionally predicts a tool-presence mask $\hat{\mathbf{m}}\in\{0,1\}^{D}$. The tool-presence head takes multimodal representation $\mathbf{H}$ as input and predicts whether each parameter should be included in the editing plan. Predicted parameter vector $\hat{\mathbf{a}}_0$ and tool-presence mask $\hat{\mathbf{m}}$ jointly define the editing plan
$\mathcal{P}(\hat{\mathbf{a}}_0,\hat{\mathbf{m}})$ introduced in \cref{sec:problem_formulation}.
\subsection{Data Preparation}
\label{sec:data_preparation}

We curate the dataset in which each sample is
$(\mathbf{x},\mathbf{x}^{*},\mathbf{c},\mathbf{a}_0,\mathbf{m})$,
where $\mathbf{x}$ and $\mathbf{x}^{*}$ are the source and expert-edited images, respectively,
$\mathbf{c}$ is the synthetic user instruction, and
$(\mathbf{a}_0,\mathbf{m})$ is the ground-truth editing plan with normalized tool parameter value $\mathbf{a}_0$ and tool-presence mask $\mathbf{m}$.

\textbf{Expert Editing Data Curation.}
Using our internal platform, we curate high-quality editing samples consisting of a source image, an edited image, and the tool parameter settings used by experts to produce the edit. We discard incomplete records and map the recorded settings to our fixed tool catalog. The parameter values are normalized to obtain $\mathbf{a}_0$, with unused tools assigned their default parameter values, while the parameters present in the expert edit define tool-presence mask $\mathbf{m}$.

\textbf{Instruction Synthesis.}
Our curated dataset does not contain the user instructions. To address this, we use a pretrained VLM to synthesize the instructions. Given source image $\mathbf{x}$, expert-edited image $\mathbf{x}^{*}$, and recorded parameter values $\mathbf{a_0}$, the VLM infers the intended visual transformation and generates corresponding instruction $\mathbf{c}$. We generate multiple instruction variants for each edit and sample one variant per training example in each epoch. Unlike autoregressive MLLM-based editing agents, FlowTool requires no reasoning traces, reducing the burden of data generation.

\subsection{Training}
\label{sec:training}

FlowTool is trained in two phases: we first learn the distribution of expert editing plans through a two-stage supervised flow-matching training (SFT phase), and then apply reward-based post-training to directly optimize the rendered editing results (RL phase). Each phase serves a different purpose: the SFT phase establishes reliable tool-parameter generation and tool-presence prediction, while the RL phase allows the model to move beyond exact imitation of the demonstrated expert tool parameters.

\subsubsection{Supervised Flow-Matching Training}
\label{sec:sft}

\textbf{Training Objective.} Given training tuple $(\mathbf{x},\mathbf{c},\mathbf{a}_0,\mathbf{m})$,
we sample $\boldsymbol{\mathbf{a}_1}\sim\mathcal{N}(\mathbf{0},\mathbf{I})$
and construct $\mathbf{a}_t$ following \cref{eq:action_flow_path}. The DiT-based parameter generator is trained to predict target velocity $\mathbf{u}_t=\boldsymbol{\mathbf{a}_1}-\mathbf{a}_0$. Since only tools present in the expert editing plan should contribute to parameter regression, we compute the flow-matching loss over the active tools indicated by mask $\mathbf{m}$:
\begin{equation}
    \mathcal{L}_{\mathrm{FM}}
    =
    \mathbb{E}
    \frac{
    \left\|
    \sqrt{\mathbf{m}} \odot
    \left(
    \mathbf{v}_{\theta}(\mathbf{a}_t,t \mid y)
    -\mathbf{u}_t
    \right)
    \right\|_2^2
    }{
    \max\left(1,\mathbf{1}^{\top}\mathbf{m}\right)
    }.
\label{eq:sft_masked_fm}
\end{equation}

In parallel, the tool-presence head predicts whether each parameter is active in the editing plan. We train it using a binary cross-entropy objective:
\begin{equation}
    \mathcal{L}_{\mathrm{pres}}
    =
    -\frac{1}{D}
    \sum_{i=1}^{D}
    \left[
    m^{i}\log \hat{m}^{i}
    +
    (1-m^{i})\log(1-\hat{m}^{i})
    \right],
\label{eq:presence_loss}
\end{equation}
where $\hat{m}^{i}$ is the predicted presence probability for the $i$-th tool.
The complete supervised objective is the sum of regular flow matching loss and the weighted cross-entropy loss for tool masking prediction parameterized by $\lambda_{pres}$:
\begin{equation}
    \mathcal{L}_{\mathrm{SFT}} = \mathcal{L}_{\mathrm{FM}} + \lambda_{\mathrm{pres}}\mathcal{L}_{\mathrm{pres}}.
\label{eq:sft_total_loss}
\end{equation}
\textbf{Two-stage SFT Curriculum.}
The VLM backbone is already pretrained, whereas the DiT parameter generator and tool-presence head are randomly initialized. Jointly optimizing all components from the beginning can cause unstable updates from the randomly initialized modules to propagate into the VLM, corrupting its pretrained multimodal representations before the new components have learned meaningful task structure. We therefore adopt a two-stage curriculum. We first freeze the VLM and train only the parameter generator and tool-presence head, allowing the newly introduced modules to first acquire the tool-parameter distribution and tool-activation patterns under stable pretrained conditioning. Then, we enable LoRA adapters in the VLM and jointly optimize all trainable components. This second stage allows the VLM representations and the prediction modules to adapt to one another, refining the multimodal features toward fine-grained editing decisions while preserving the knowledge acquired during the first stage.

\subsubsection{Reward-Based Post-Training}
\label{sec:rl}

\textbf{Intuition.} The SFT phase encourages FlowTool to reproduce the demonstrated expert editing plans. However, the expert parameter setting represents only one possible solution, and different parameter configurations can produce edits that satisfy the same user request. We therefore further optimize the SFT model using DiffusionNFT~\citep{zheng2025diffusionnft}, as introduced in \cref{sec:prelim_flow_matching}. For each image--instruction pair, the model samples a group of editing plans, which are executed by the renderer to obtain edited images
$\{\hat{\mathbf{x}}^{(k)}\}_{k=1}^{G}$.
The resulting images are scored and used to construct the group-relative rewards for training.

\textbf{Reward Design.}
Our total reward is a weighted sum of reference-based reward $\lambda_\mathrm{ref}$ and reference-free reward $\lambda_\mathrm{vlm}$:
\begin{equation}
    R^{(k)} =
    \lambda_{\mathrm{ref}}R_{\mathrm{ref}}^{(k)} +
    \lambda_{\mathrm{vlm}}R_{\mathrm{vlm}}^{(k)}.
\label{eq:composite_render_reward}
\end{equation} 
The reference-based reward measures similarity to expert-edited target
$\mathbf{x}^{*}$ using the negative pixel-wise $L_1$ distance. The negative sign converts the distance into a reward, such that outputs closer to the expert rendition receive higher scores as follows:
\begin{equation}
R_{\mathrm{ref}}^{(k)}=-\|\hat{\mathbf{x}}^{(k)}-\mathbf{x}^{*}\|_1.
\label{eq:reference_render_reward}
\end{equation}

Because matching a single expert rendition does not fully capture whether an edit satisfies the user's request, we additionally employ a VLM-as-a-judge reward. Given original image $\mathbf{x}$, rendered output $\hat{\mathbf{x}}^{(k)}$, and user instruction $\mathbf{c}$, the VLM judge evaluates how well the transformation from $\mathbf{x}$ to $\hat{\mathbf{x}}^{(k)}$ fulfills the editing instruction:
\begin{equation}
R_{\mathrm{vlm}}^{(k)}=\operatorname{VLMJudge}(\mathbf{x},\hat{\mathbf{x}}^{(k)},\mathbf{c}).
\label{eq:vlm_render_reward}
\end{equation}
Together, the two rewards encourage FlowTool to remain consistent with expert editing outcomes while directly optimizing whether the rendered result fulfills the user's instructions.
\section{Experiments}
\subsection{Experimental Settings}
\label{sec:experimental_settings}

\textbf{Models and Training Data.}
We use Qwen3.5-4B~\citep{qwen3.5} as the VLM backbone of the FlowTool model. During reward-based post-training, we utilize Qwen3.8-27B~\citep{qwen38} as the VLM judge. Our training corpus is constructed from an internally curated collection of expert editing records. Following \cref{sec:data_preparation}, we synthesize user instructions using Gemma4-31B~\citep{team2026gemma}, resulting in 496,693 training examples. Finally, FlowTool uses three ODE steps at inference time by default.

\textbf{Benchmarks.}
We evaluate FlowTool on three established tool-based image-editing benchmarks, MMArt-Bench~\citep{lin2026jarvisart}, ArtEdit-Bench~\citep{jarvisevo2026}, and MIT-Adobe5K~\citep{bychkovsky2011fivek}. We additionally evaluate on \textbf{FlowTool-Eval}, which contains 300 held-out examples from our internally curated dataset, with no overlap with the training data. 
The number of tools and the complete tool catalog are provided in \cref{app:editing_tools}.

\textbf{Evaluation Metrics.}
Following standard practice in prior work~\citep{lin2026jarvisart,retouchiq2026}, we report both reference-based and reference-free metrics. For reference-based evaluation, we measure agreement with the expert-edited target using PSNR, SSIM, and LPIPS~\citep{zhang2018unreasonable} on MIT-Adobe5K, and $L_1$ and $L_2$ on other benchmarks. For reference-free evaluation, we report Semantic Consistency (SC), Perceptual Quality (PQ), and Overall Score (O)~\citep{ku2023viescore} using GPT-5~\citep{gpt5} as the judge.

\textbf{Baselines.}
We compare FlowTool with two groups of baselines. Firstly, we evaluate proprietary MLLMs, including GPT-5.6 Sol, Gemini~3.1 Pro, Claude Sonnet 5, and Claude Opus 4.8, prompting them to generate editing plans without task-specific training. Secondly, we compare against specialized MLLM agents: JarvisArt~\citep{lin2026jarvisart} and RetouchIQ~\citep{retouchiq2026}.
\subsection{Results \& Analysis}
\label{sec:result}

\definecolor{mygreen}{RGB}{1, 165, 81}
\definecolor{myred}{RGB}{181, 51, 29}

\begin{table*}[t]
\centering
\caption{
\textbf{Performance comparison on 4 datasets}.
$\downarrow$ indicates lower is better and $\uparrow$ means higher is better.
The best and second-best for each column are shown in \textbf{bold} and
\underline{underlined}, respectively. The $\Delta$ rows report the relative percentage improvement (in \textcolor{Green}{green}) and regression (in \textcolor{BrickRed}{red}) of the best-performing FlowTool variant (SFT or RL) against the strongest general/specialized MLLMs.
}
\vspace{-6pt}

\label{tab:main_results}
\footnotesize
\setlength{\tabcolsep}{2.0pt}
\renewcommand{\arraystretch}{1.1}
\resizebox{\textwidth}{!}{%
\begin{tabular}{l ccccc ccccc ccccc ccc}
    \toprule
    \multirow{2}{*}{\bf Method}
    & \multicolumn{5}{c}{\textbf{MMArt-Bench}}
    & \multicolumn{5}{c}{\textbf{FlowTool-Eval}}
    & \multicolumn{5}{c}{\textbf{ArtEdit-Bench}}
    & \multicolumn{3}{c}{\textbf{MIT-Adobe5K}}
    \\
    \cmidrule(lr){2-6}
    \cmidrule(lr){7-11}
    \cmidrule(lr){12-16}
    \cmidrule(lr){17-19}
    &
    \textbf{L1} $\downarrow$ &
    \textbf{L2} $\downarrow$ &
    \textbf{SC} $\uparrow$ &
    \textbf{PQ} $\uparrow$ &
    \textbf{O} $\uparrow$
    &
    \textbf{L1} $\downarrow$ &
    \textbf{L2} $\downarrow$ &
    \textbf{SC} $\uparrow$ &
    \textbf{PQ} $\uparrow$ &
    \textbf{O} $\uparrow$
    &
    \textbf{L1} $\downarrow$ &
    \textbf{L2} $\downarrow$ &
    \textbf{SC} $\uparrow$ &
    \textbf{PQ} $\uparrow$ &
    \textbf{O} $\uparrow$
    &
    \textbf{PSNR} $\uparrow$ &
    \textbf{SSIM} $\uparrow$ &
    \textbf{LPIPS} $\downarrow$
    \\
    \midrule
    GPT-5.6 Sol
    & 10.77 & 22.68 & \best{8.34} & \second{9.38} & \second{8.80}
    & 8.81 & 16.44 & \second{8.27} & 9.55 & \best{8.84}
    & 8.73 & 16.60 & \second{8.37} & 9.53 & 8.88
    & 17.25 & \best{0.52} & 0.38
    \\
    Gemini 3.1 Pro
    & 12.16 & 27.65 & 8.14 & 9.22 & 8.63
    & 10.23 & 21.10 & \best{8.30} & 9.41 & \second{8.81}
    & 9.96 & 21.82 & 8.31 & 9.39 & 8.78
    & 16.88 & 0.45 & 0.42
    \\
    Claude Sonnet 5
    & 10.92 & 22.57 & 8.20 & 9.35 & 8.72
    & 9.47 & 18.47 & 7.68 & 9.56 & 8.50
    & 8.77 & 16.53 & 8.34 & 9.50 & 8.86
    & 17.58 & 0.47 & 0.41
    \\
    Claude Opus 4.8
    & 12.13 & 27.48 & \second{8.26} & 9.28 & 8.73
    & 9.38 & 17.95 & 8.19 & 9.48 & 8.78
    & 8.97 & 16.99 & \best{8.44} & 9.47 & \second{8.90}
    & 17.56 & 0.46 & \second{0.37}
    \\
    \midrule
    JarvisArt
    & 11.64 & 28.69 & 7.31 & 9.25 & 8.13
    & 11.06 & 25.12 & 7.31 & 9.44 & 8.19
    & 9.56 & 22.72 & 7.87 & 9.46 & 8.56
    & 17.39 & 0.47 & 0.40
    \\
    RetouchIQ
    & 17.94 & 60.80 & 6.42 & 9.01 & 7.45
    & 16.83 & 58.64 & 7.01 & 9.16 & 7.92
    & 15.24 & 48.83 & 7.41 & 9.07 & 8.12
    & 14.16 & 0.39 & 0.52
    \\
    \midrule
    \rowcolor{gray!12}
    \textbf{FlowTool-SFT}
    & \best{9.37} & \best{17.78} & 7.96 & \best{9.59} & 8.69
    & \second{8.24} & \second{14.65} & 7.53 & \best{9.65} & 8.41
    & \second{7.90} & \second{13.92} & 8.13 & \best{9.78} & 8.87
    & \second{17.83} & \best{0.52} & \best{0.36}
    \\

    \rowcolor{gray!12}
    \textbf{FlowTool-RL}
    & \second{9.42} & \second{17.98} & 8.16 & \best{9.59} & \best{8.81}
    & \best{7.87} & \best{13.21} & 8.04 & \second{9.64} & 8.76
    & \best{7.65} & \best{12.95} & 8.34 & \second{9.75} & \best{8.99}
    & \best{17.90} & \second{0.51} & \best{0.36}
    \\

    \addlinespace[1pt]
    
    \textit{$\Delta$ vs. General}
    & \gain{13.0} & \gain{21.2} & \drop{2.2} & \gain{2.2} & \gain{0.1}
    & \gain{10.7} & \gain{19.6} & \drop{3.1} & \gain{0.9} & \drop{0.9}
    & \gain{12.4} & \gain{21.7} & \drop{1.2} & \gain{2.6} & \gain{1.0}
    & \gain{1.8} & \same{0.0} & \gain{2.7}
    \\
    
    \textit{$\Delta$ vs. Specialized}
    & \gain{19.5} & \gain{38.0} & \gain{11.6} & \gain{3.7} & \gain{8.4}
    & \gain{28.8} & \gain{47.4} & \gain{10.0} & \gain{2.2} & \gain{7.0}
    & \gain{20.0} & \gain{43.0} & \gain{6.0} & \gain{3.4} & \gain{5.0}
    & \gain{2.9} & \gain{10.6} & \gain{10.0}
    \\

    \bottomrule
\end{tabular}%
}
\end{table*}
\newcommand{\qualimg}[1]{
    \vspace{-7pt}
    \includegraphics[width=0.135\textwidth]{#1}
}

\newcommand{\qualimgcrop}[3]{%
    \vspace{-7pt}%
    \resizebox{0.135\textwidth}{!}{%
        \clipbox*{0 {#1\height} {\width} {#2\height}}{%
            \includegraphics{#3}%
        }%
    }%
}

\newcommand{\qualinst}[1]{%
\multicolumn{7}{p{0.97\textwidth}}{\centering \vspace{-12pt}\scriptsize #1}%
}

\begin{figure}[!h]
    \centering
    \setlength{\tabcolsep}{1pt}
    \renewcommand{\arraystretch}{1.05}

    \begin{adjustbox}{width=\textwidth}
    \begin{tabular}{
        >{\centering\arraybackslash}m{0.135\textwidth}
        >{\centering\arraybackslash}m{0.135\textwidth}
        >{\centering\arraybackslash}m{0.135\textwidth}
        >{\centering\arraybackslash}m{0.135\textwidth}
        >{\centering\arraybackslash}m{0.135\textwidth}
        >{\centering\arraybackslash}m{0.135\textwidth}
        >{\centering\arraybackslash}m{0.135\textwidth}}
        \scriptsize\textbf{Input} &
        \scriptsize\textbf{Human Edit} &
        \scriptsize\textbf{FlowTool (Ours)} &
        \scriptsize\textbf{GPT-5.6 Sol} &
        \scriptsize\textbf{Gemini 3.1 Pro} &
        \scriptsize\textbf{JarvisArt} &
        \scriptsize\textbf{RetouchIQ} \\[6pt]
        
        \qualimgcrop{0.3}{0.95}{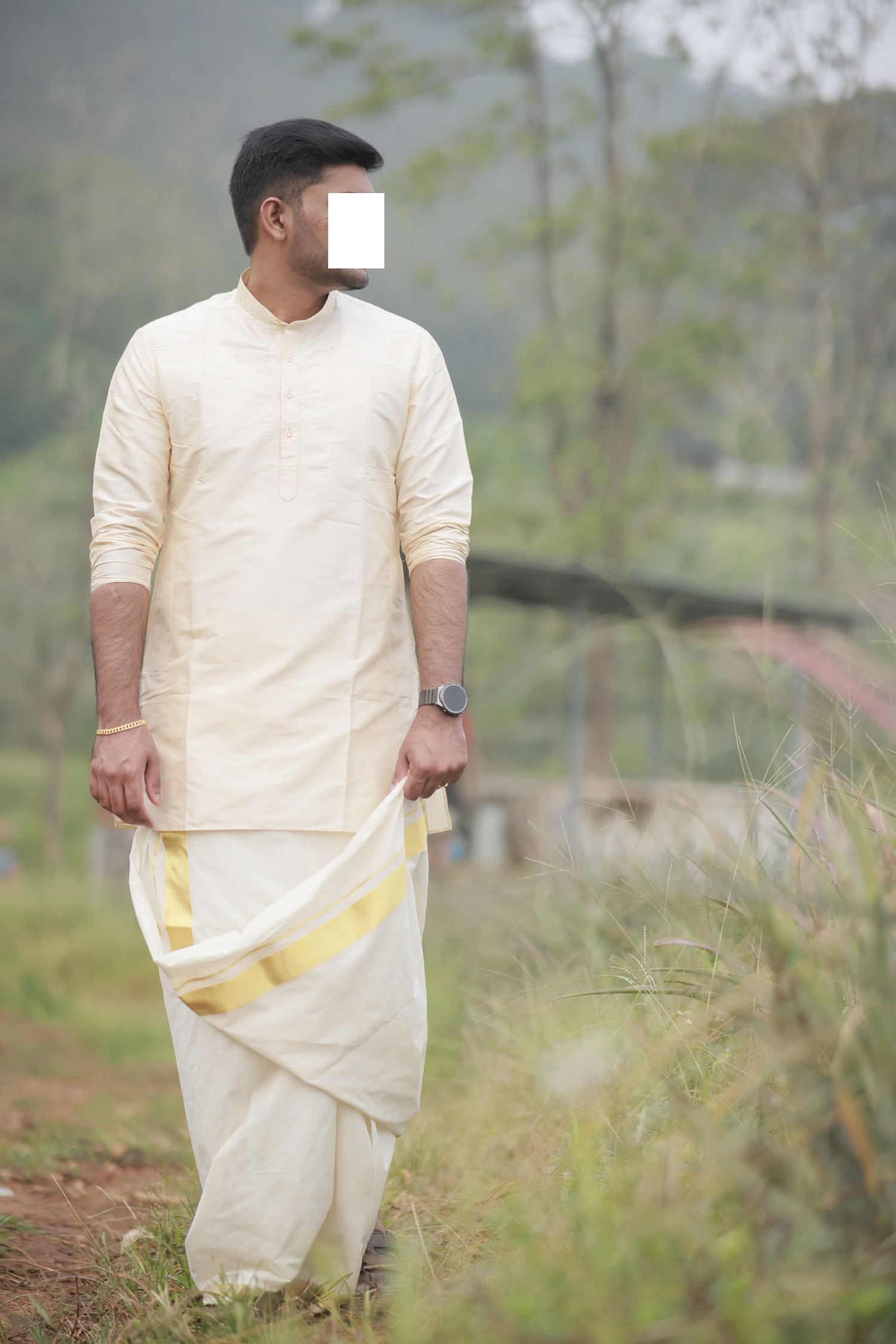} &
        \qualimgcrop{0.3}{0.95}{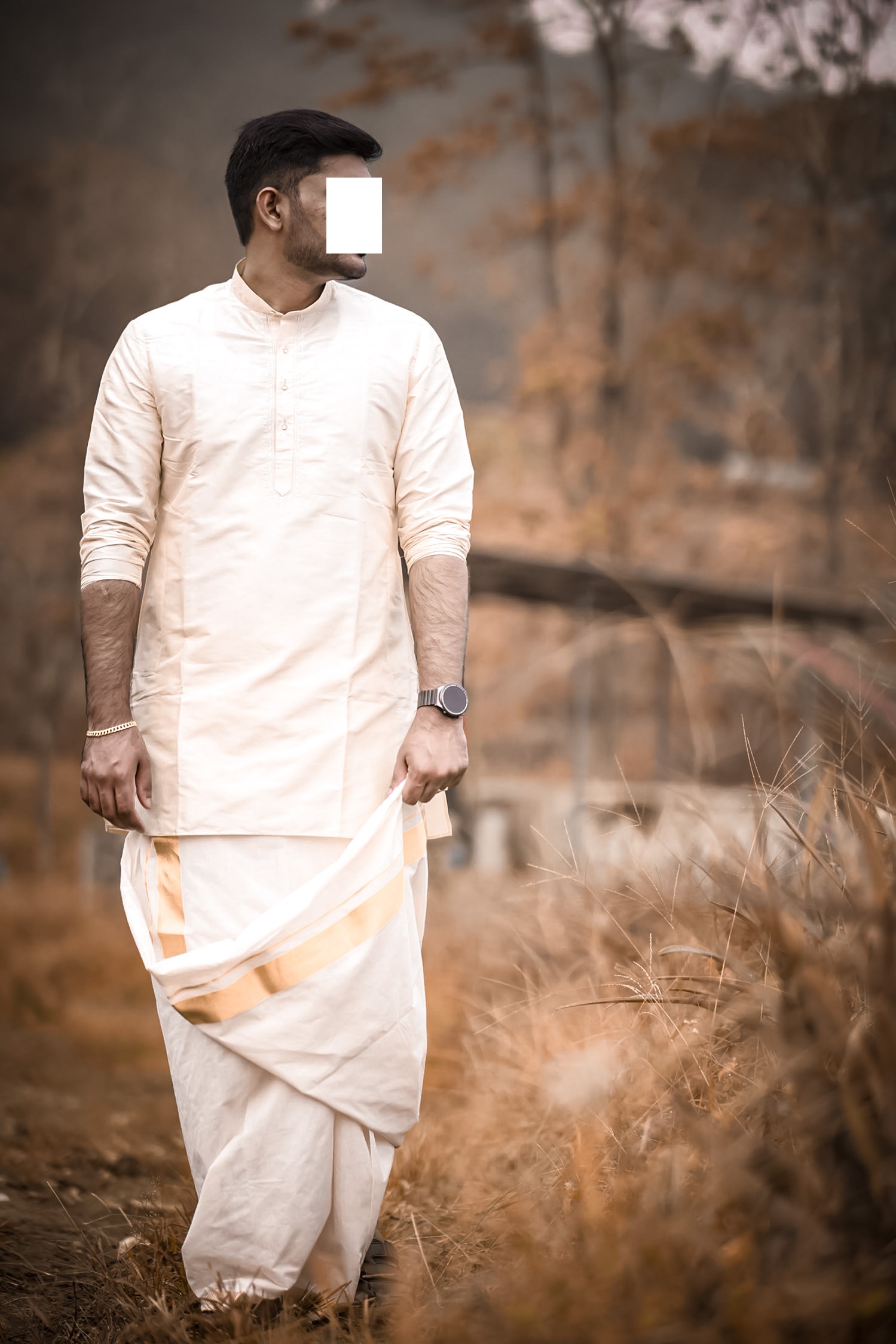} &
        \qualimgcrop{0.3}{0.95}{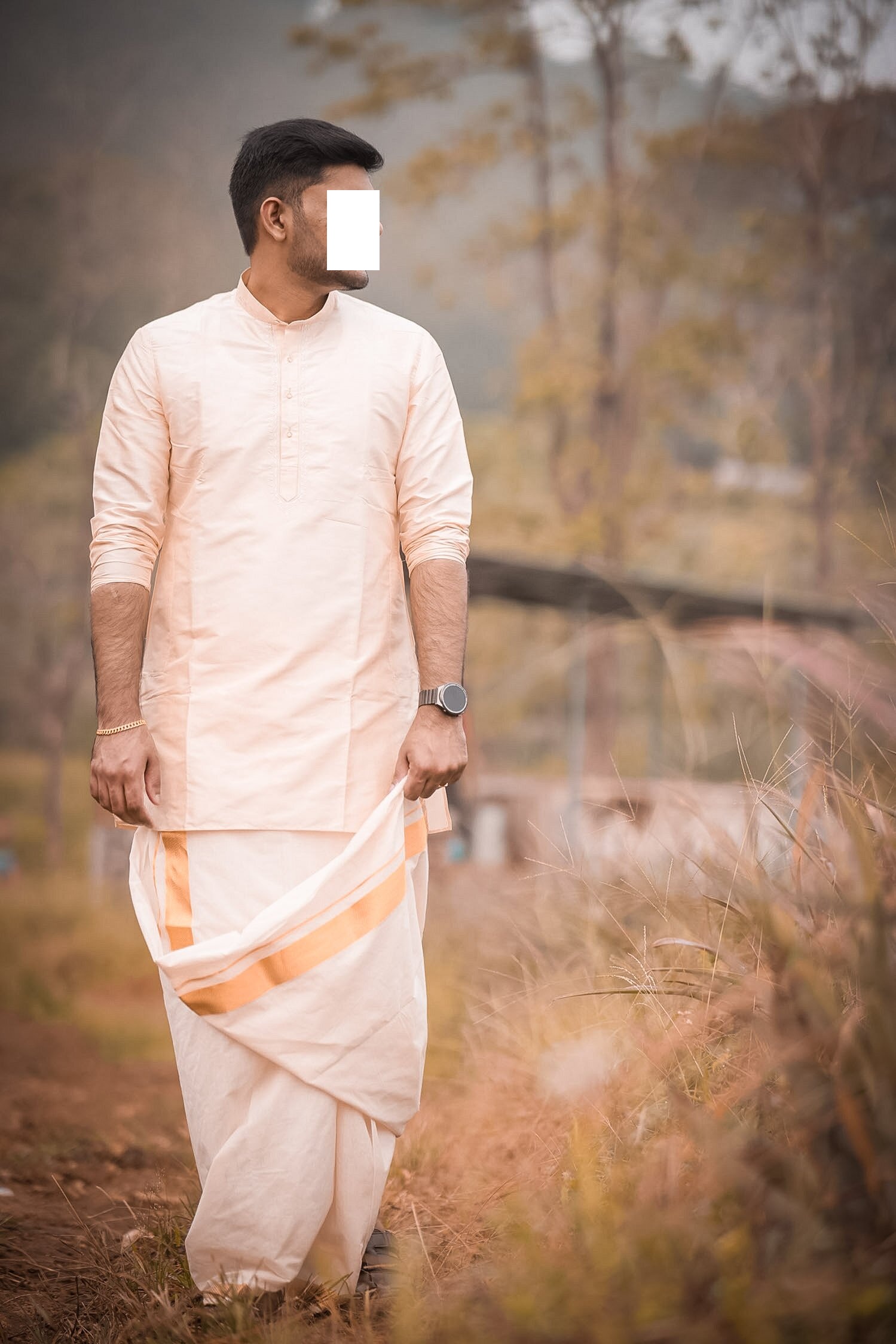} &
        \qualimgcrop{0.3}{0.95}{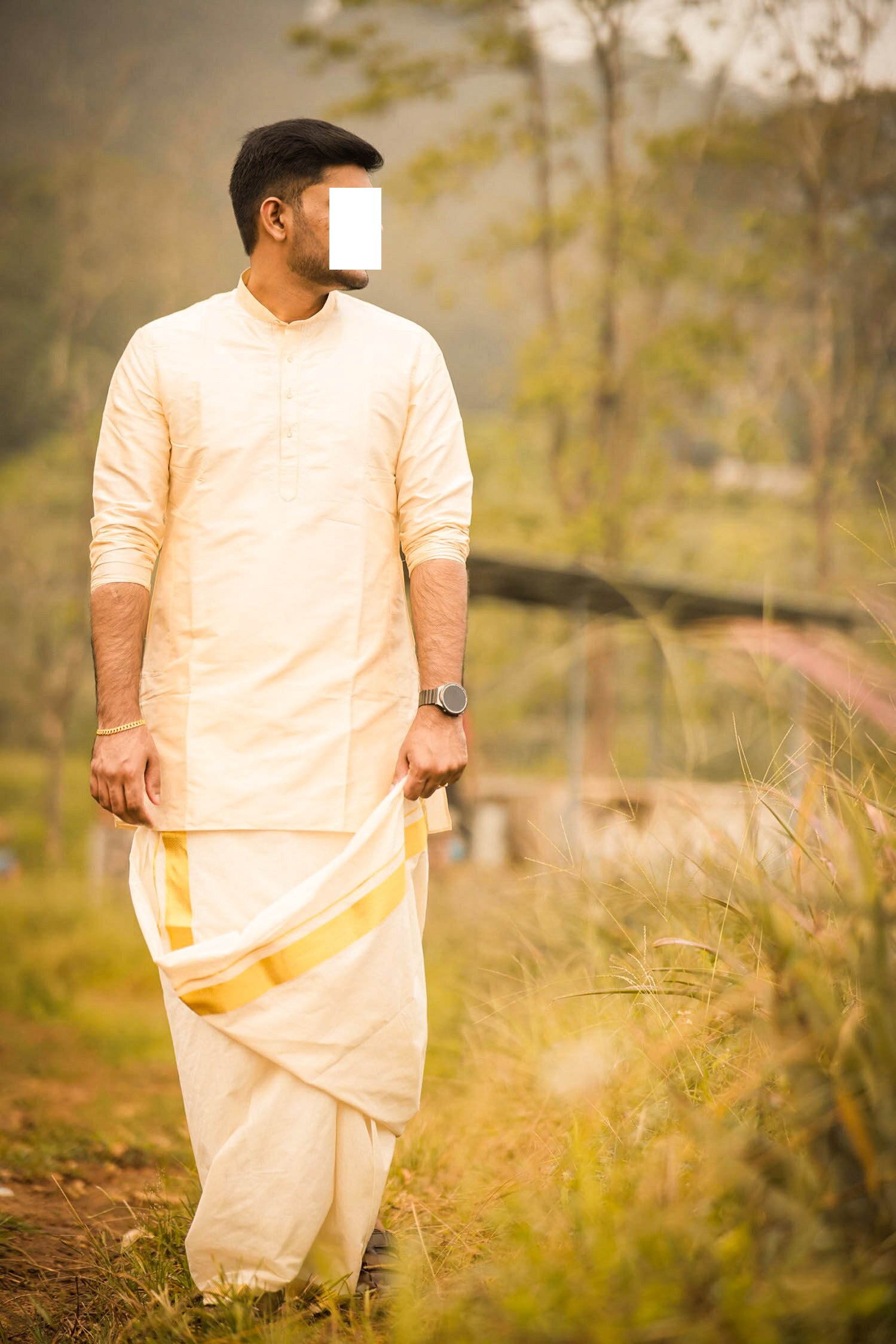} &
        \qualimgcrop{0.3}{0.95}{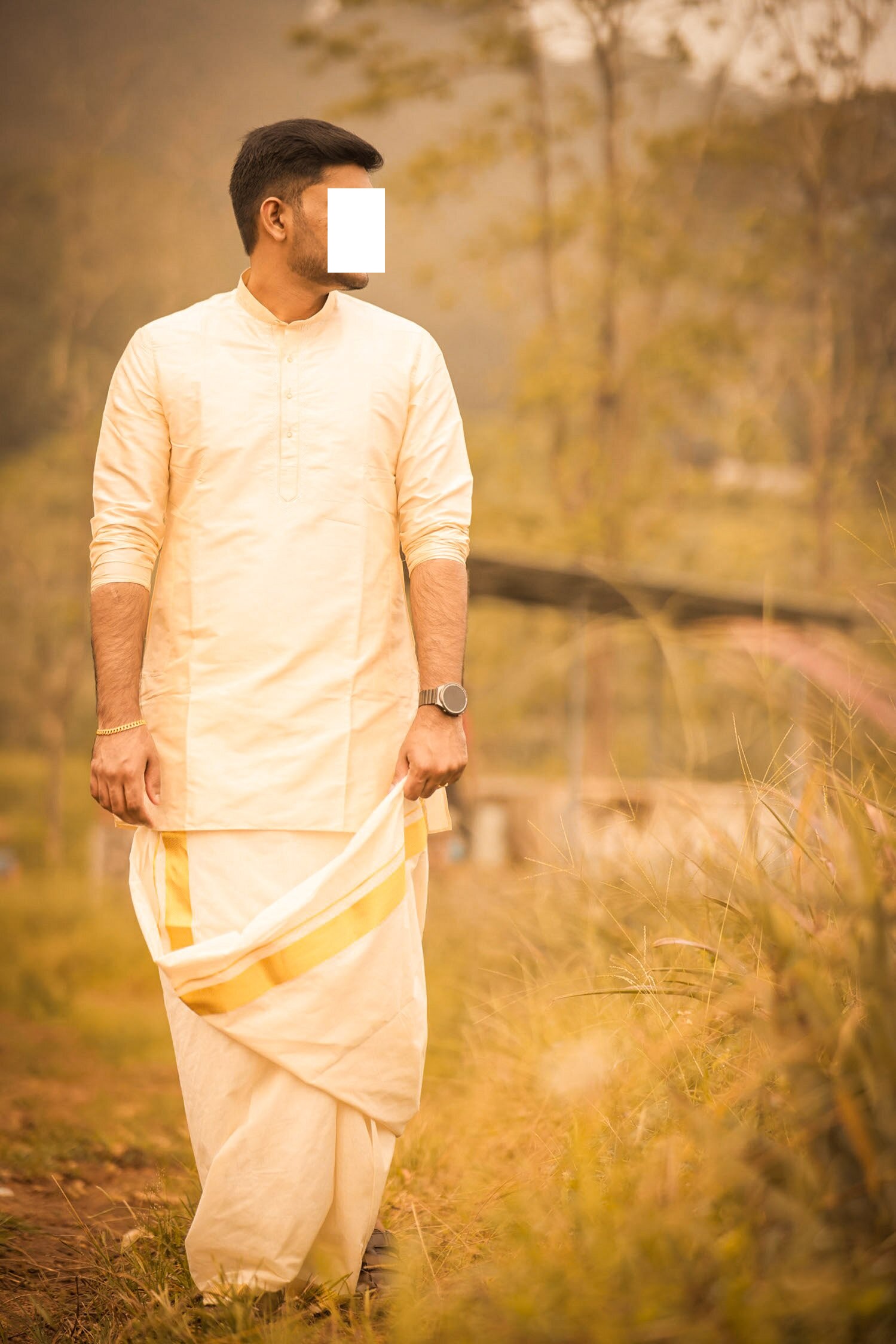} &
        \qualimgcrop{0.3}{0.95}{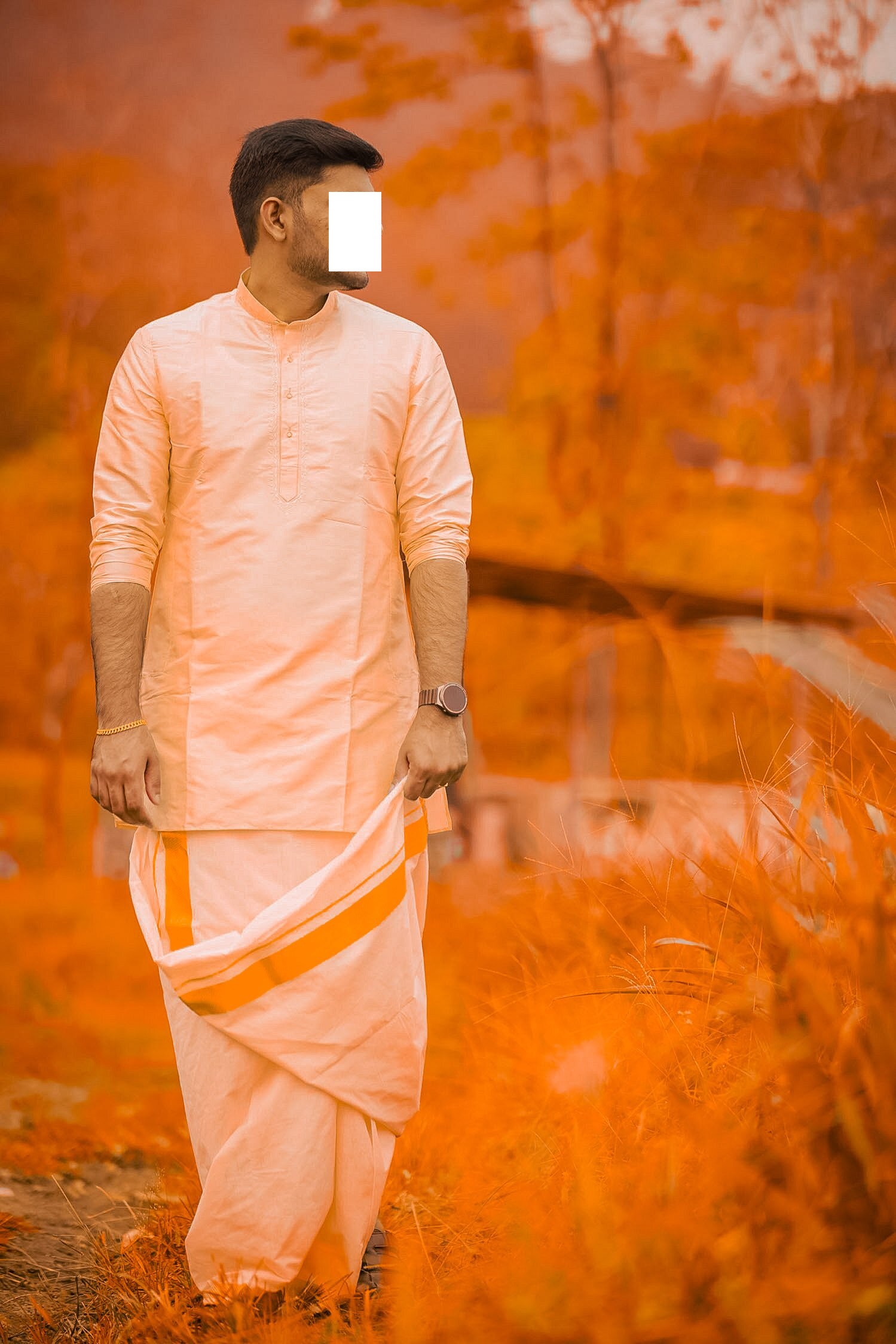} &
        \qualimgcrop{0.3}{0.95}{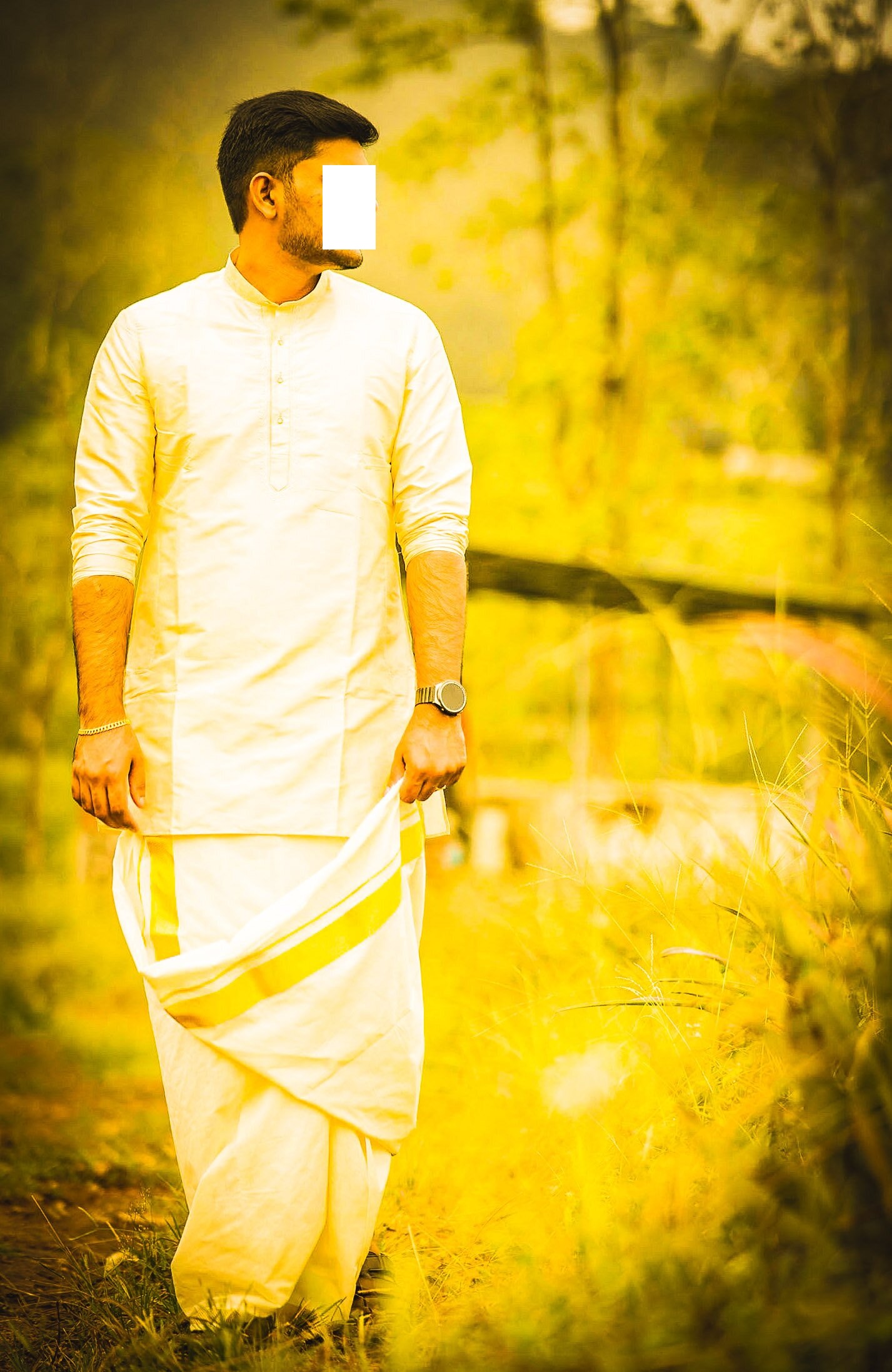} \\
        \qualinst{\input{images/qual1/instruction}
        } \\ [10pt]
        
        \qualimgcrop{0.3}{0.95}{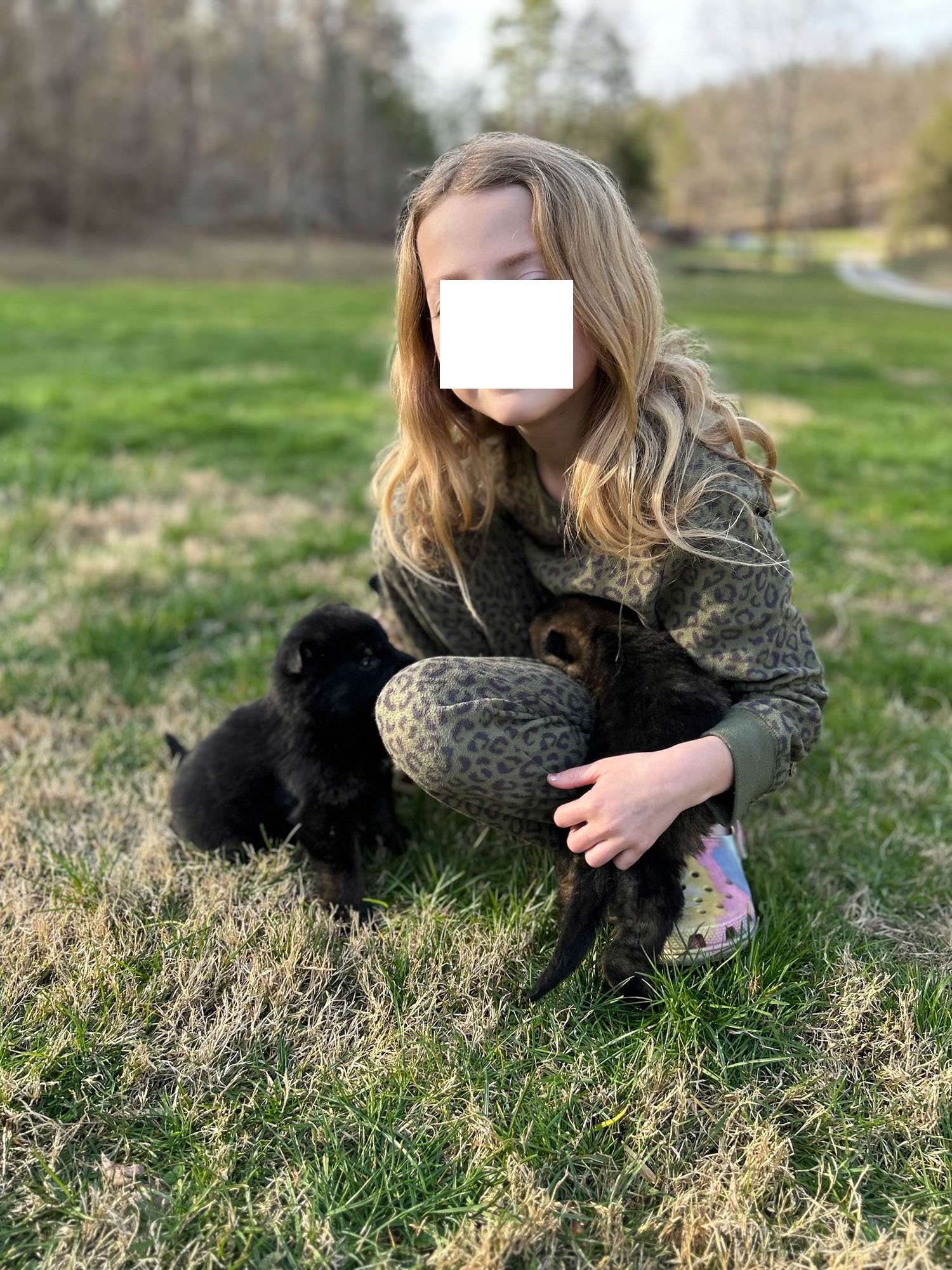} &
        \qualimgcrop{0.3}{0.95}{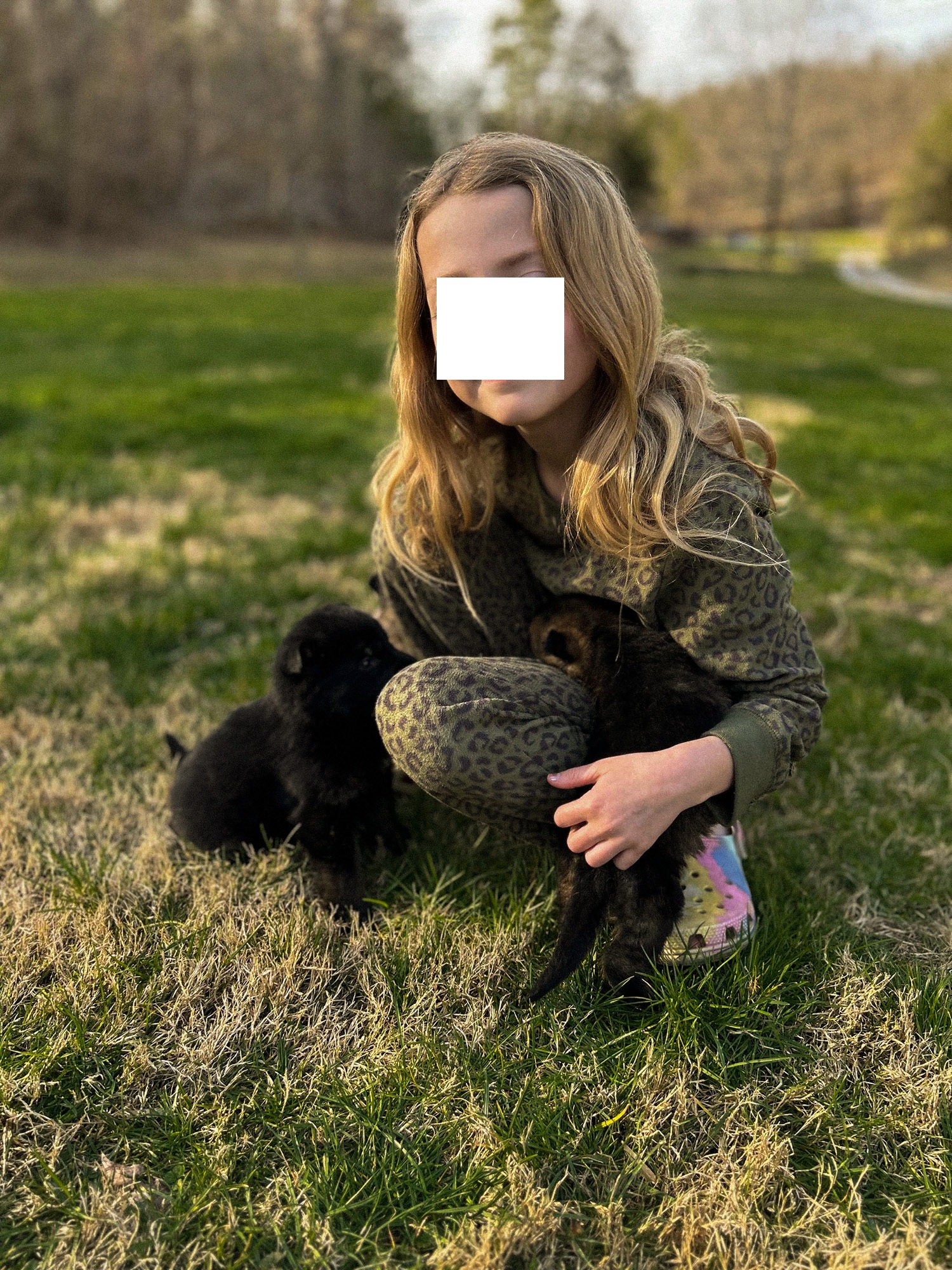} &
        \qualimgcrop{0.3}{0.95}{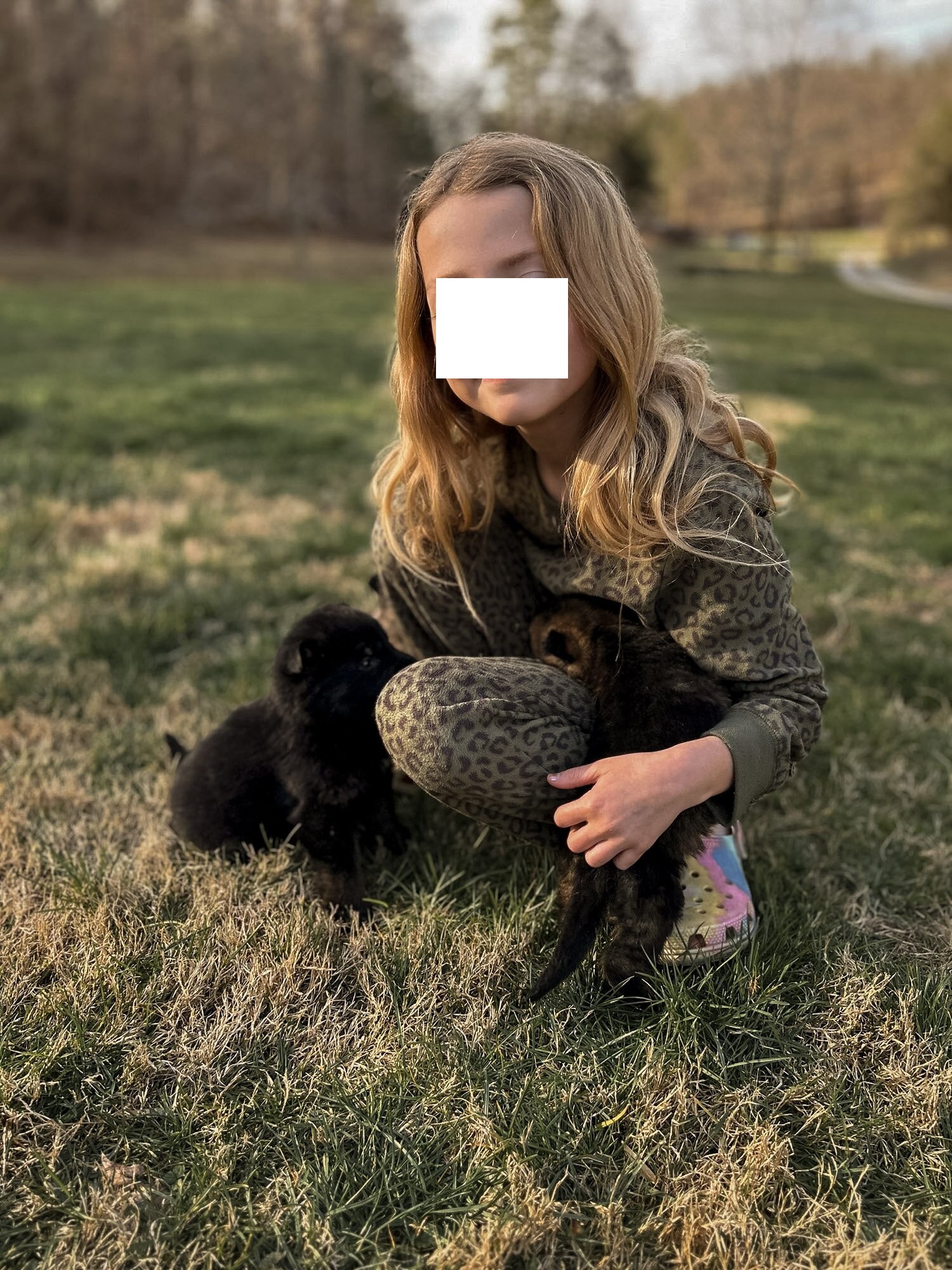} &
        \qualimgcrop{0.3}{0.95}{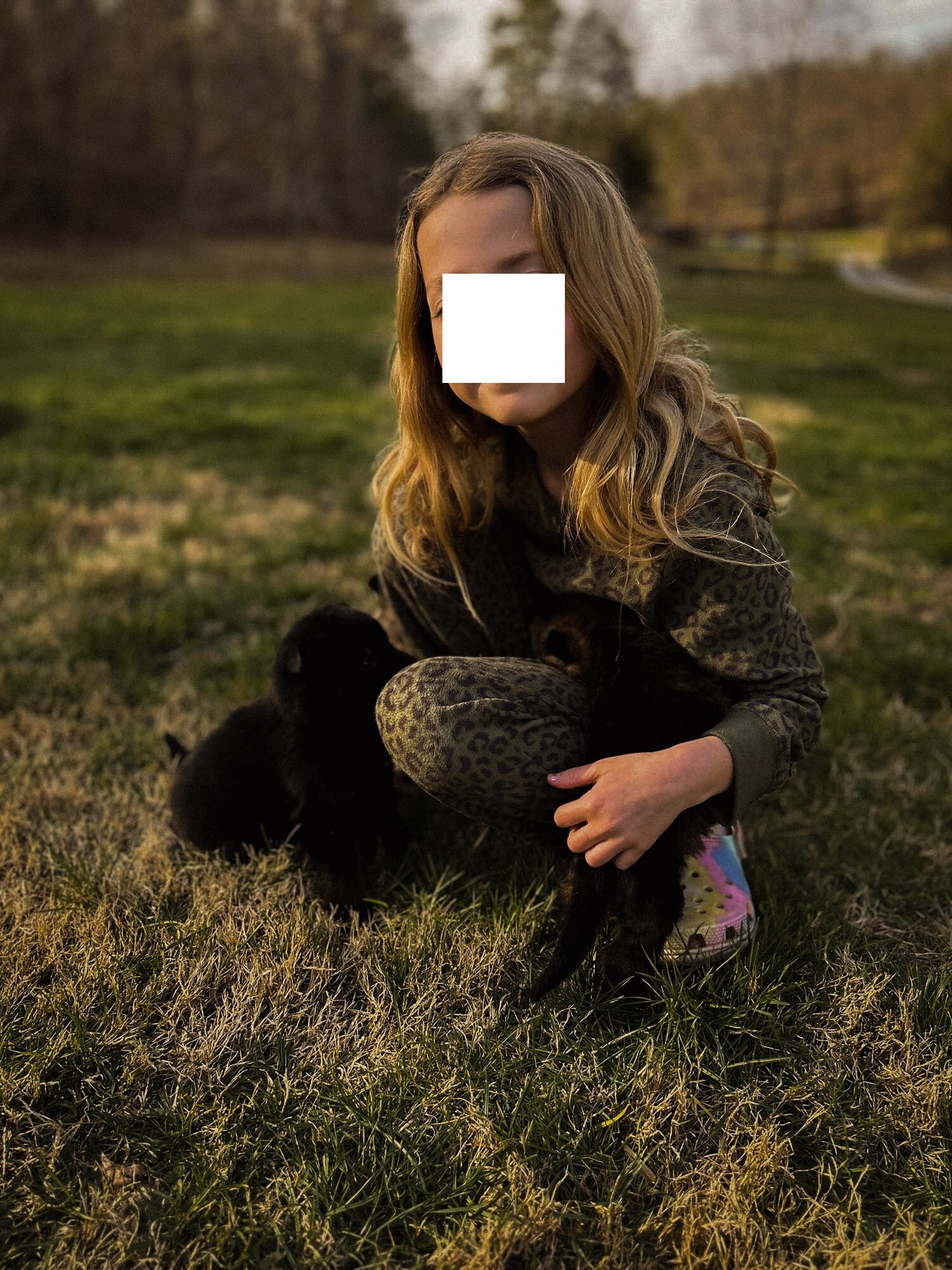} &
        \qualimgcrop{0.3}{0.95}{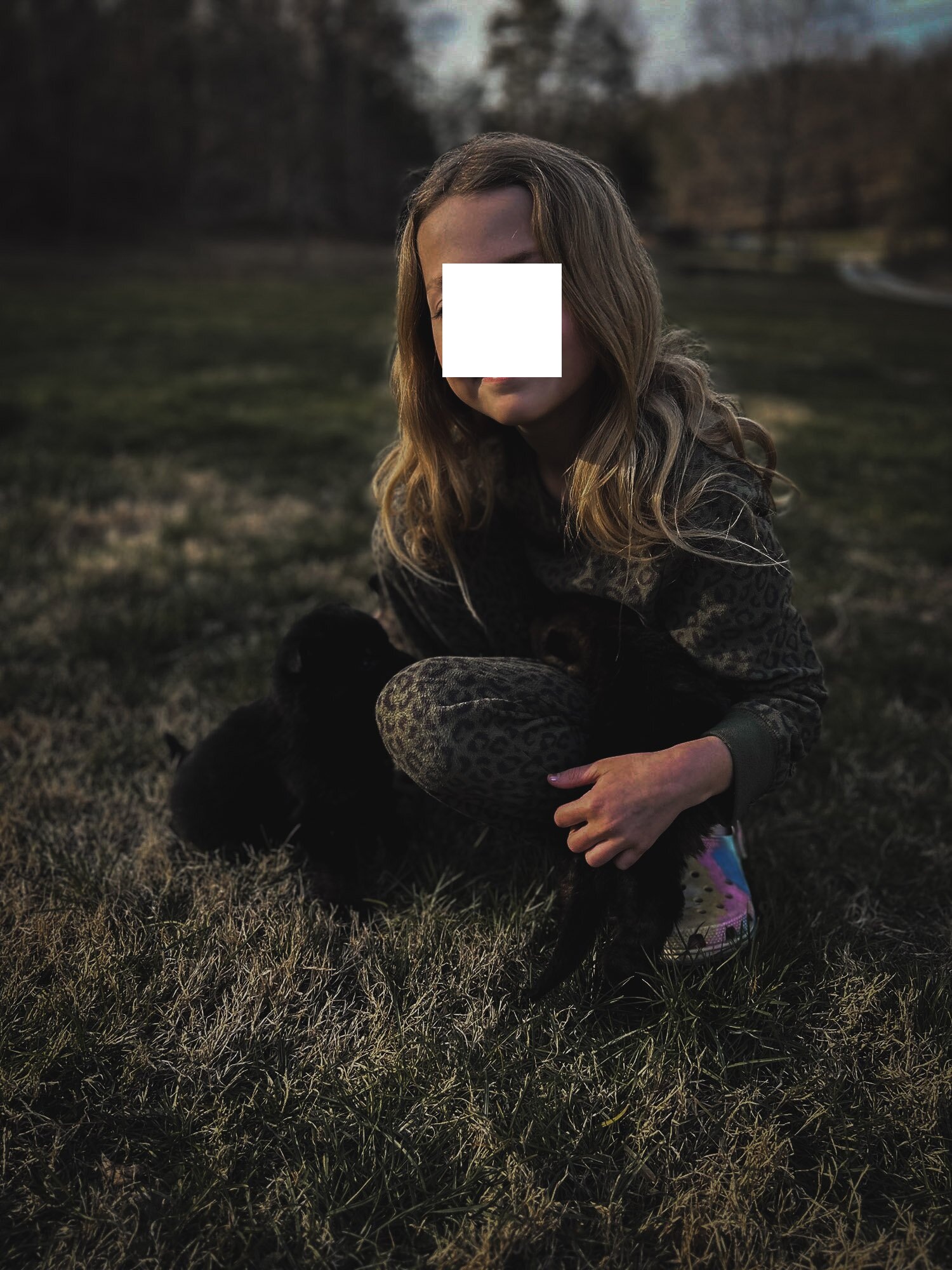} &
        \qualimgcrop{0.3}{0.95}{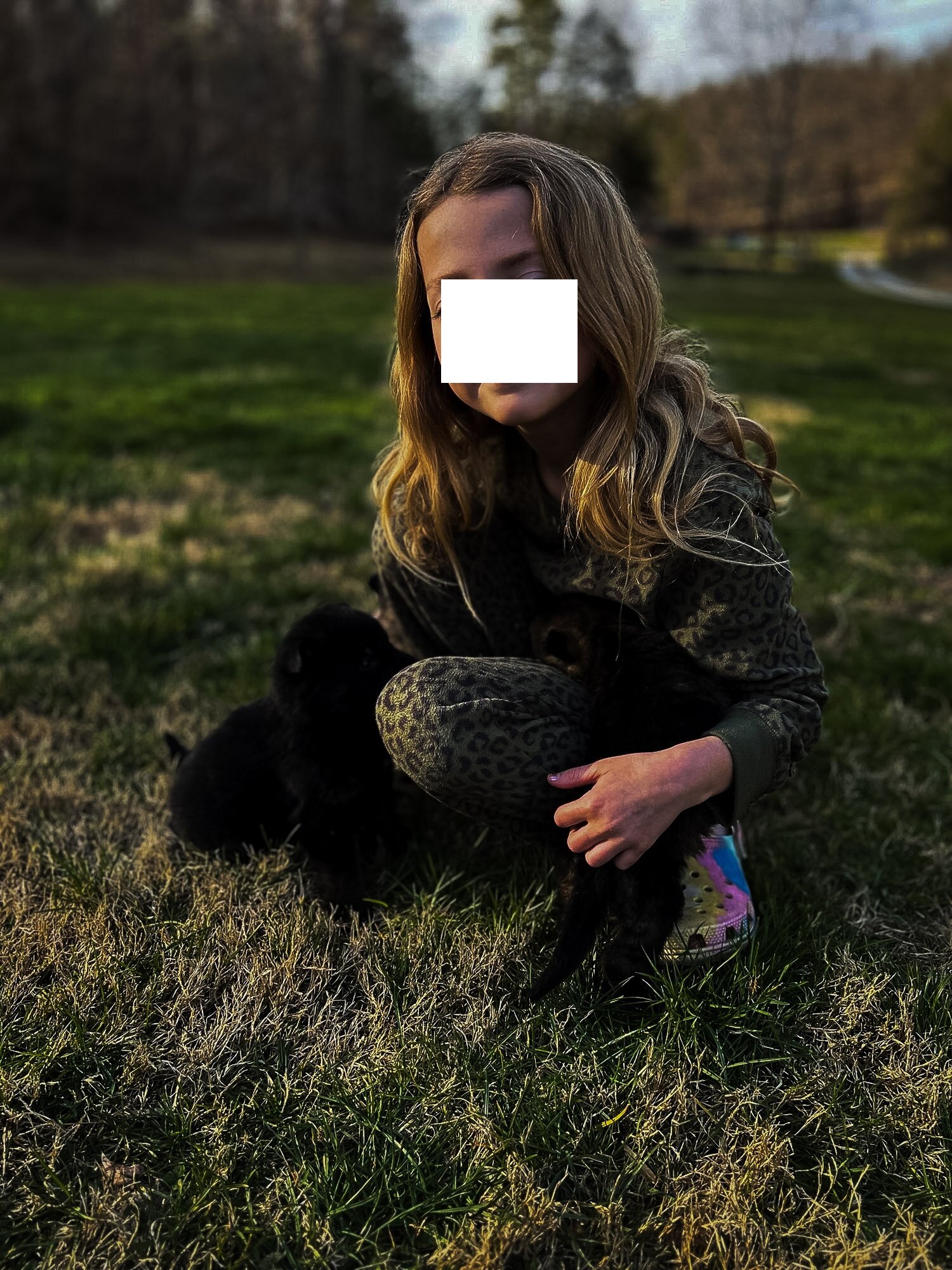} &
        \qualimgcrop{0.3}{0.95}{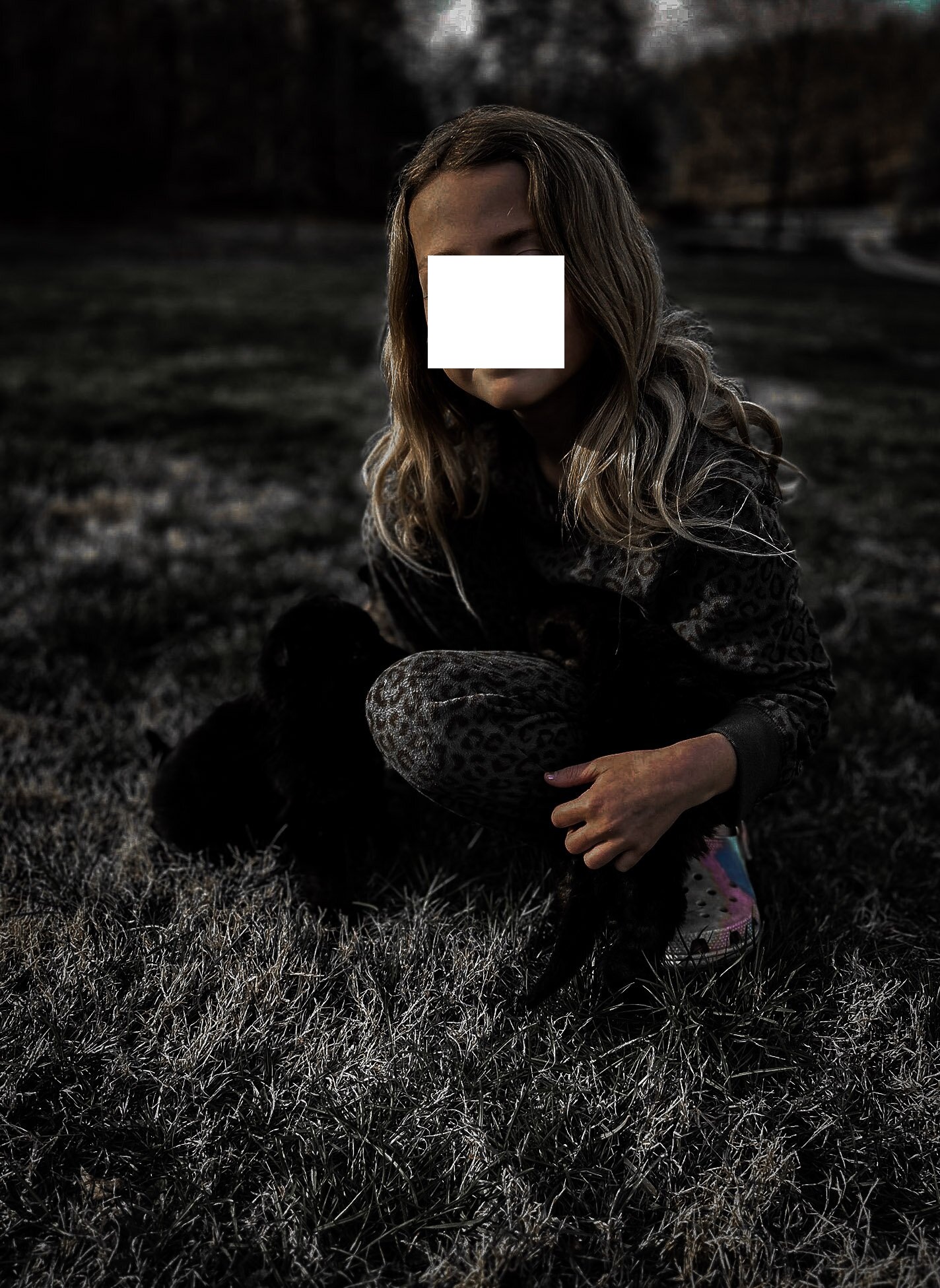}  \\
        \qualinst{\input{images/qual2/instruction}} \\[10pt]
        
        \qualimgcrop{0.2}{0.8}{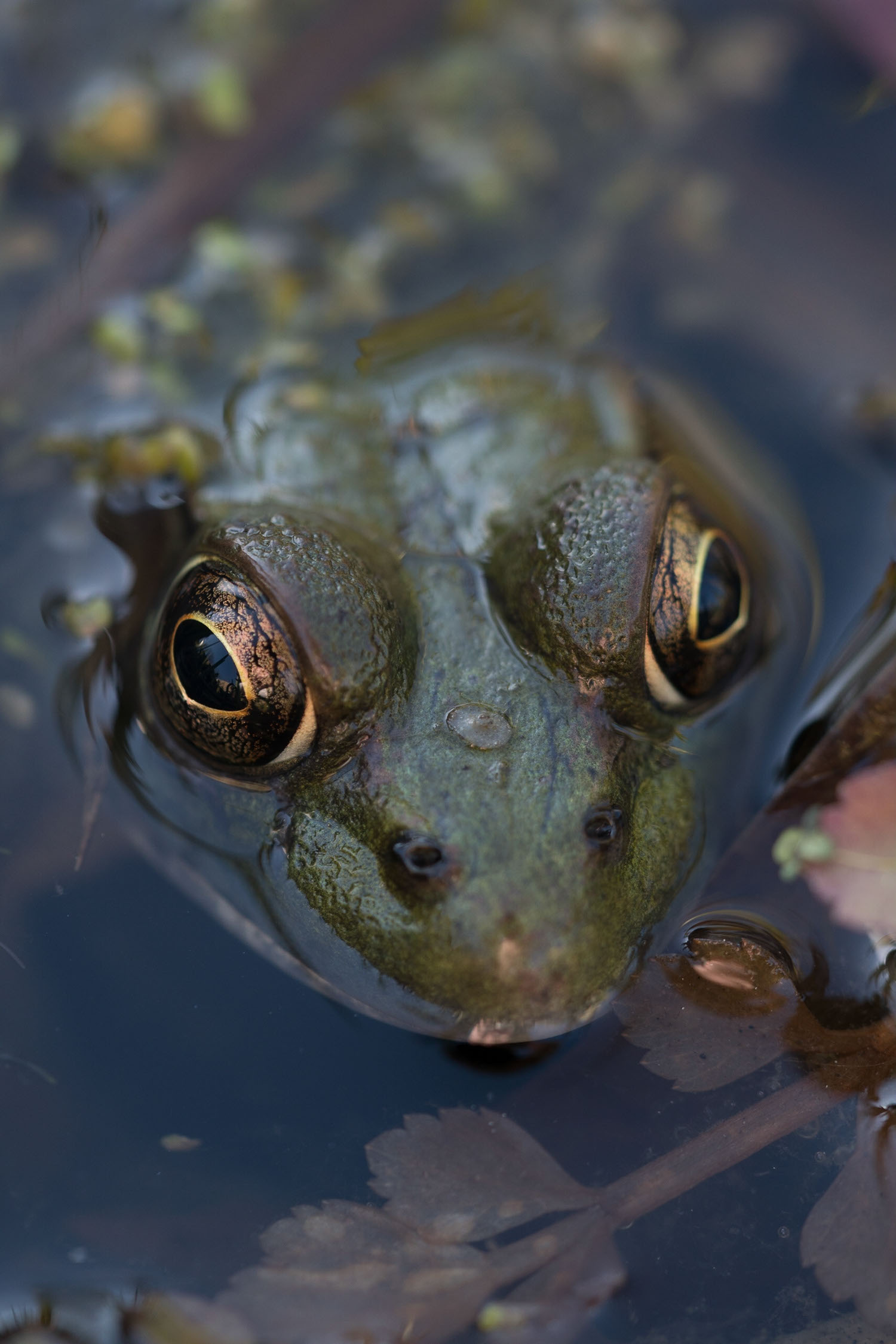} &
        \qualimgcrop{0.2}{0.8}{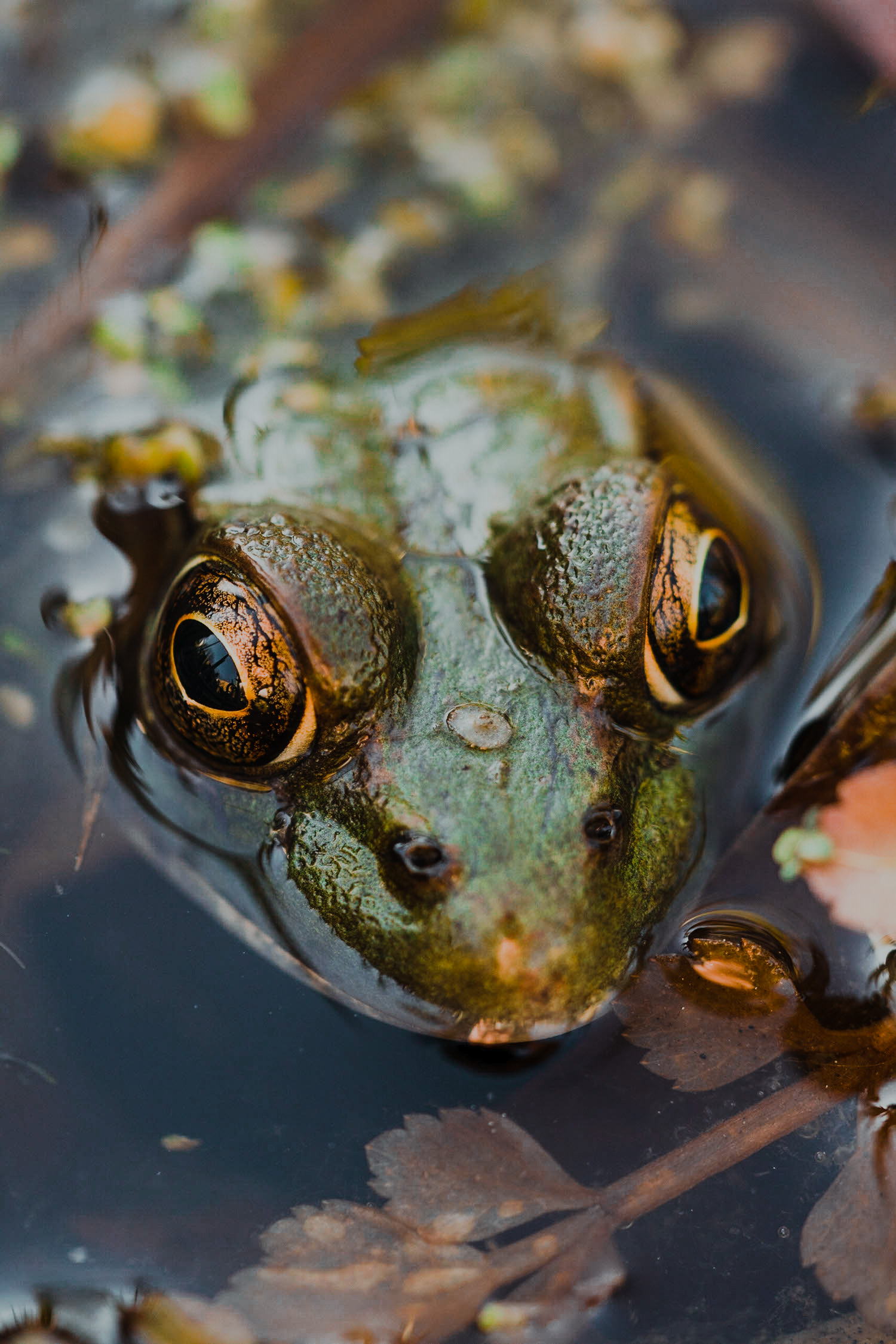} &
        \qualimgcrop{0.2}{0.8}{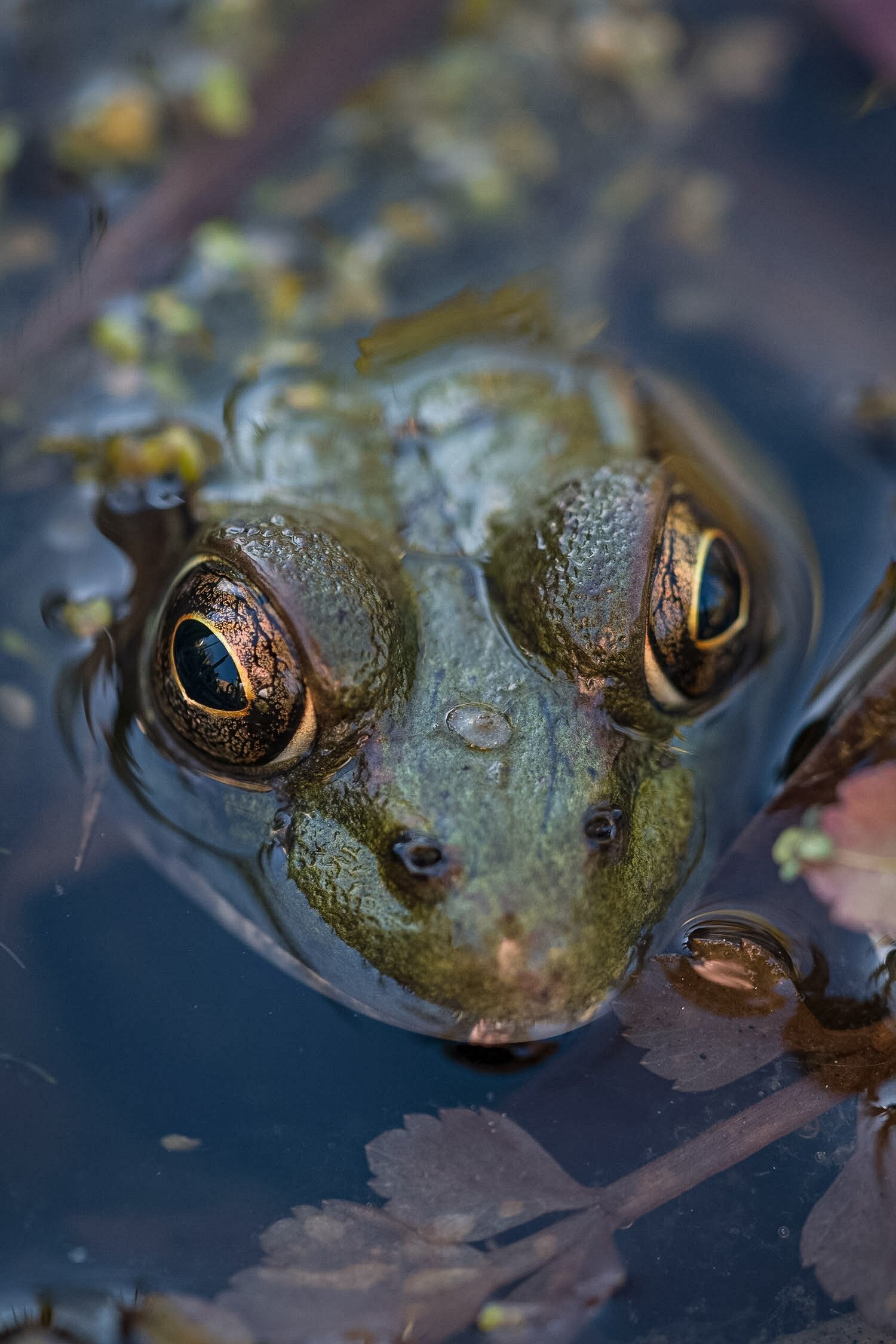} &
        \qualimgcrop{0.2}{0.8}{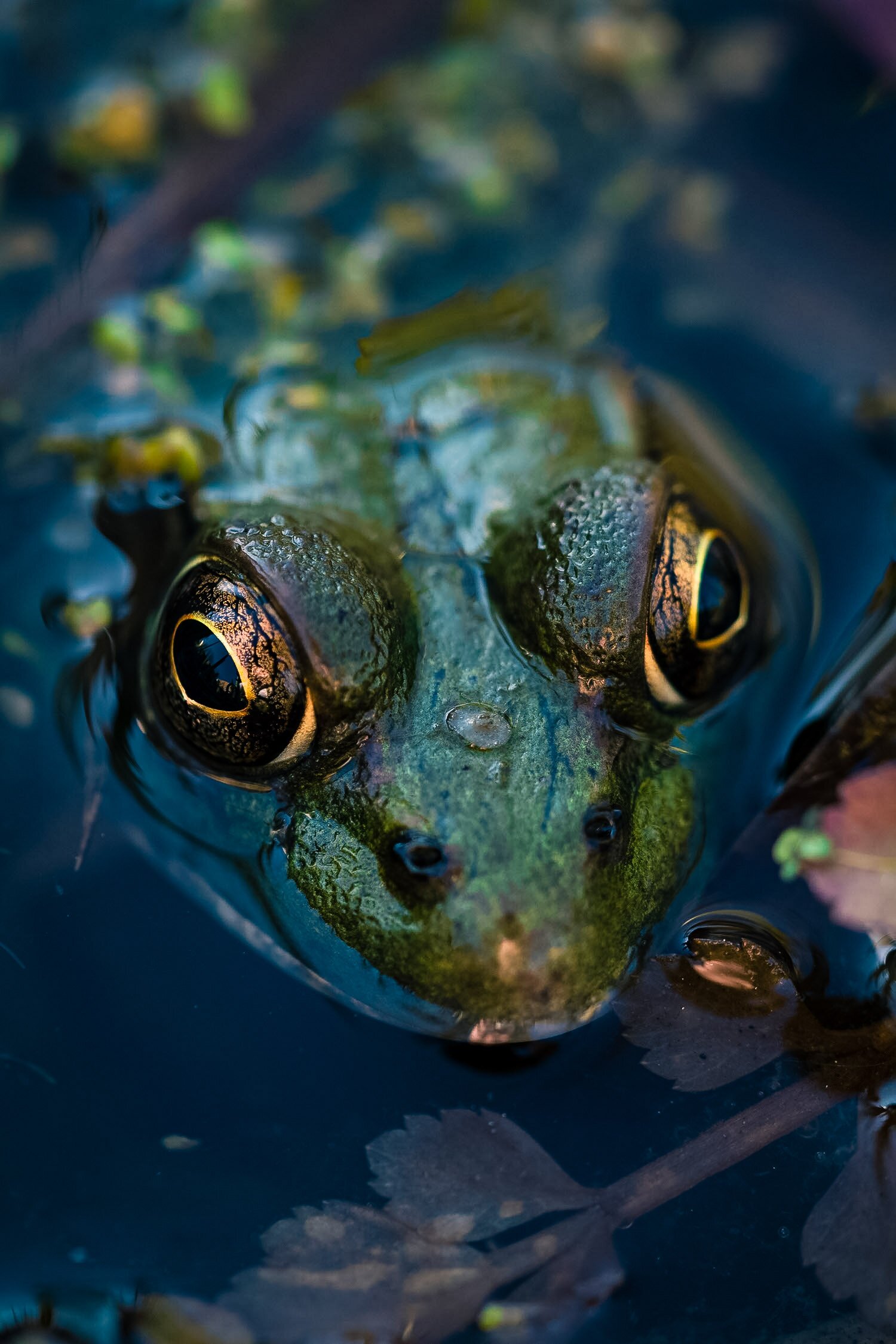} &
        \qualimgcrop{0.2}{0.8}{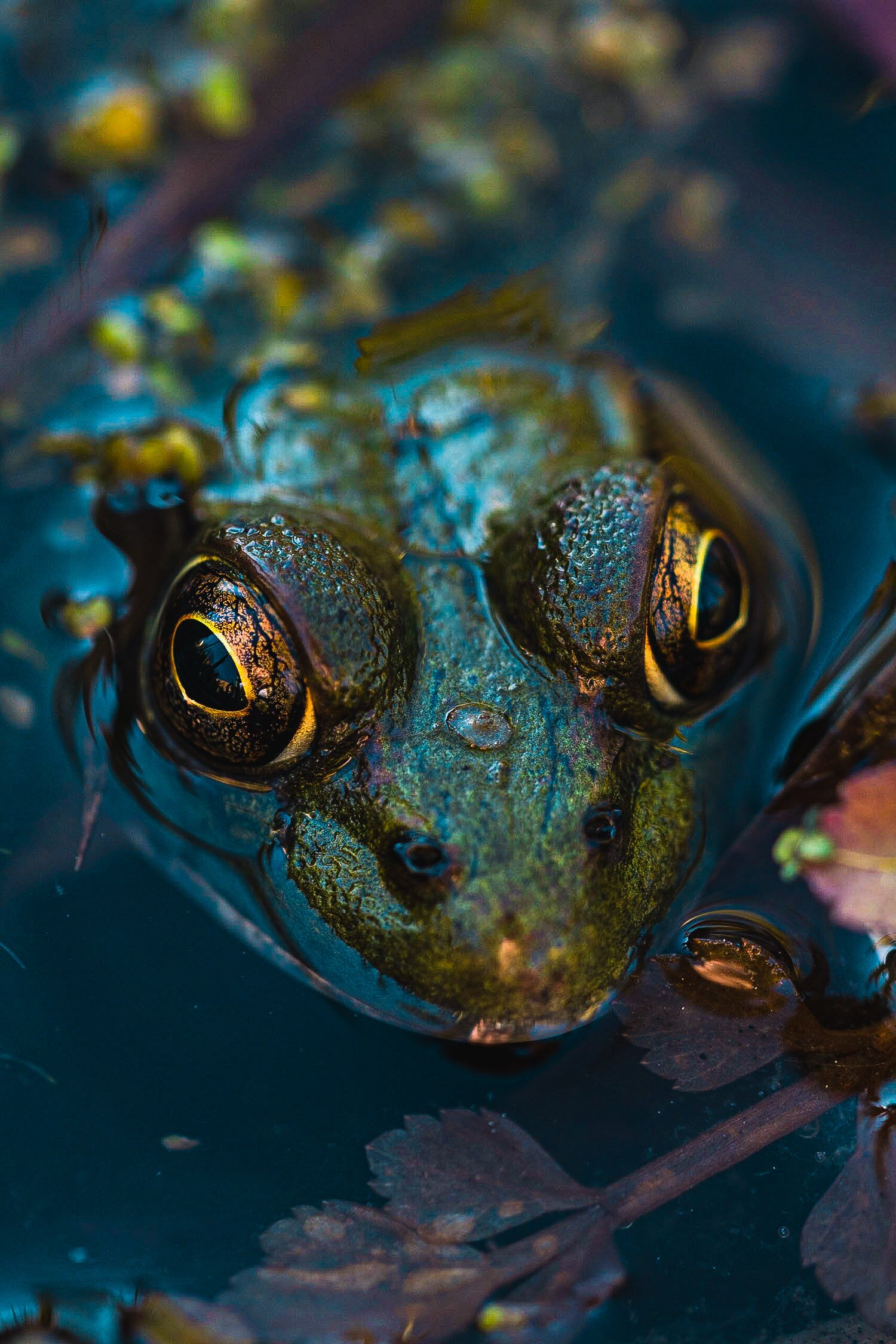} &
        \qualimgcrop{0.2}{0.8}{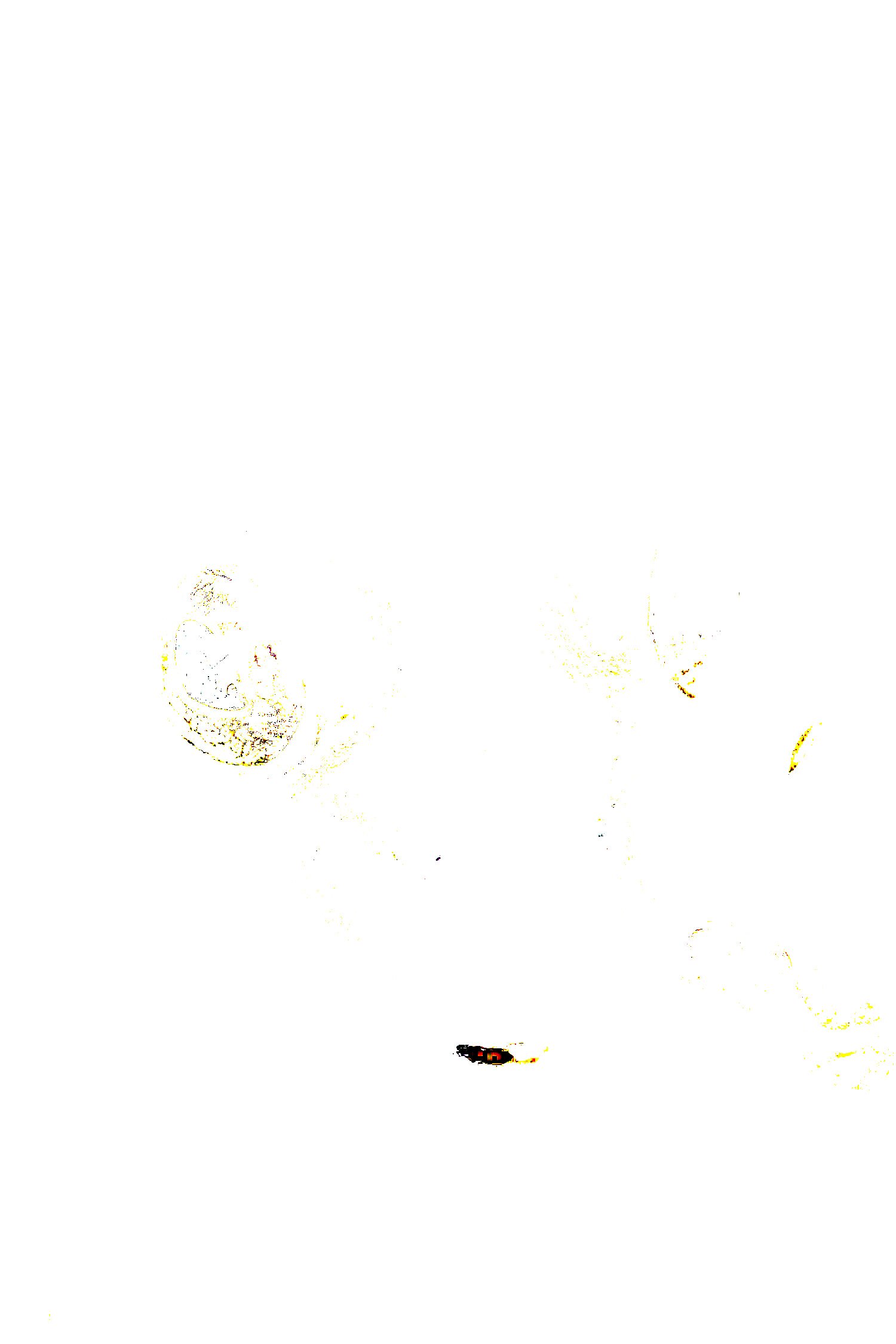} &
        \qualimgcrop{0.1}{0.8}{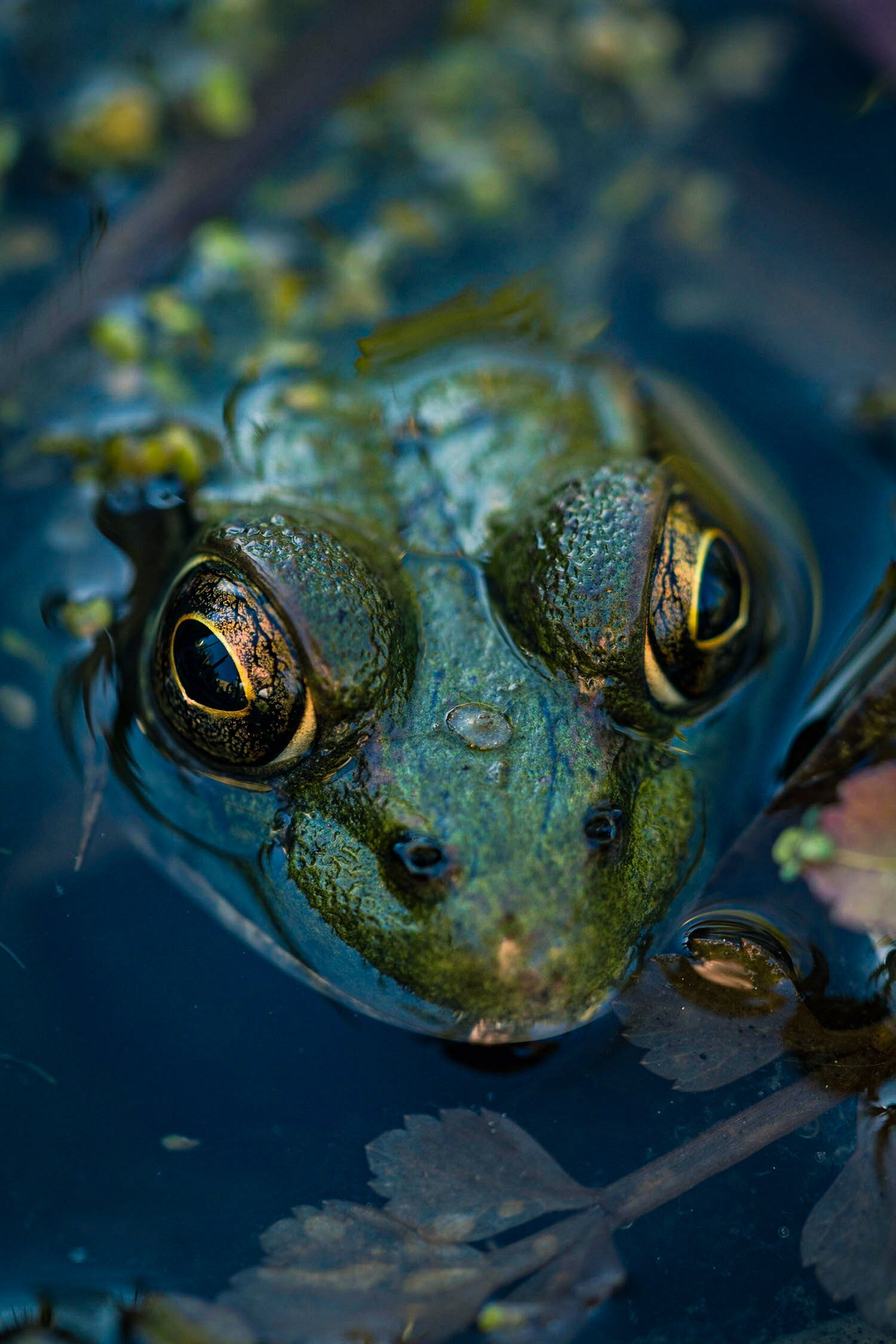}  \\
        \qualinst{\input{images/qual3/instruction}} \\[10pt]
        
        \qualimg{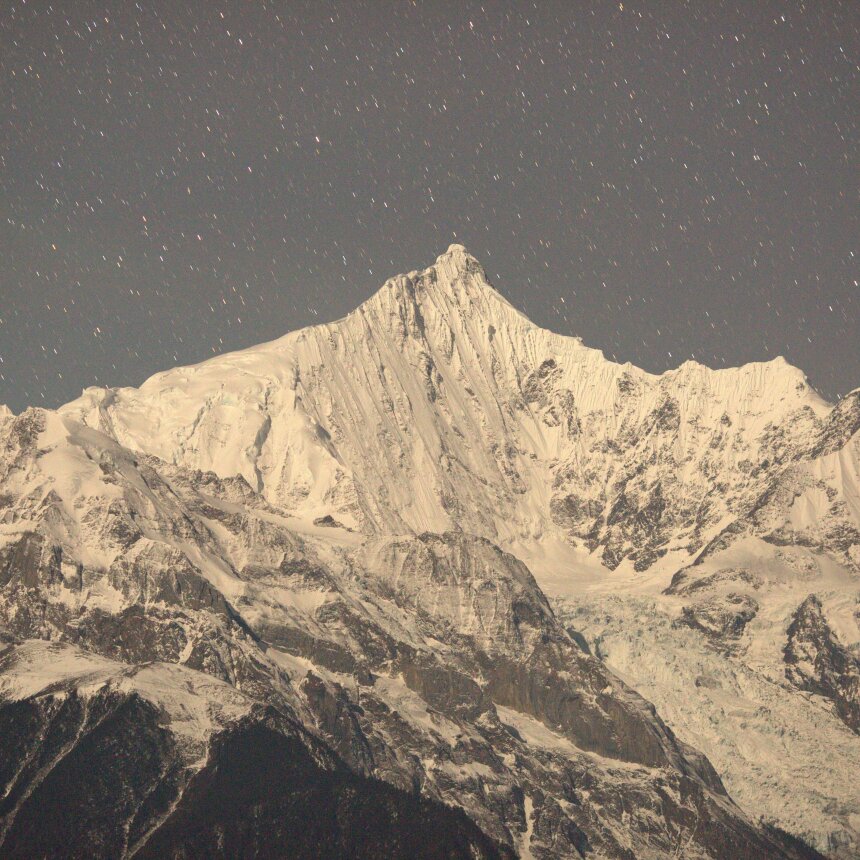} &
        \qualimg{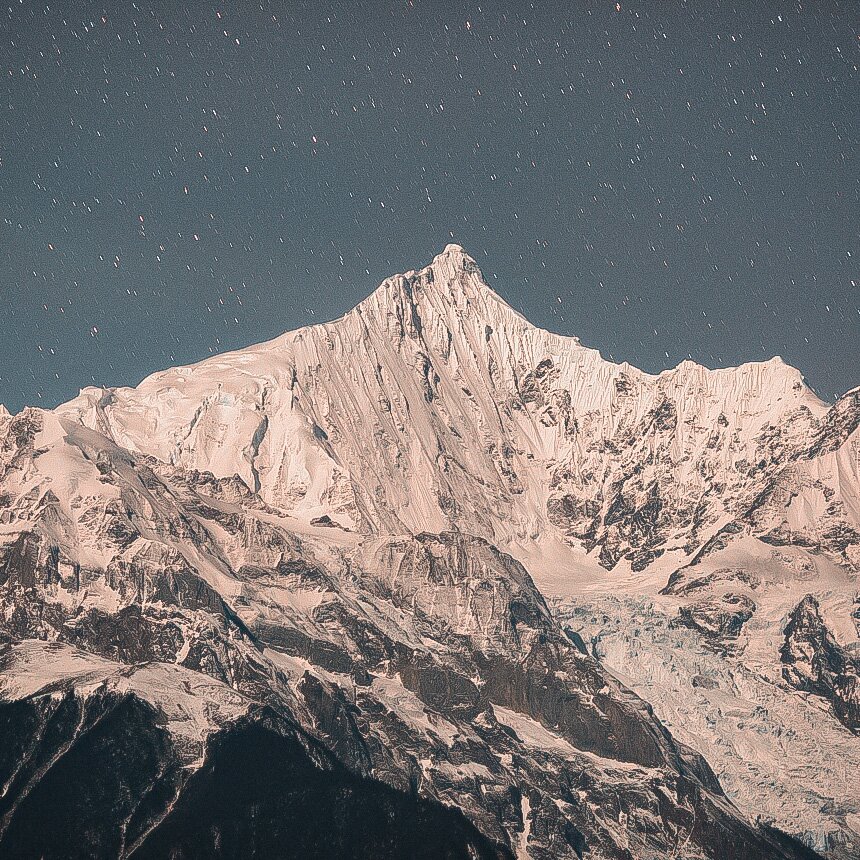} &
        \qualimg{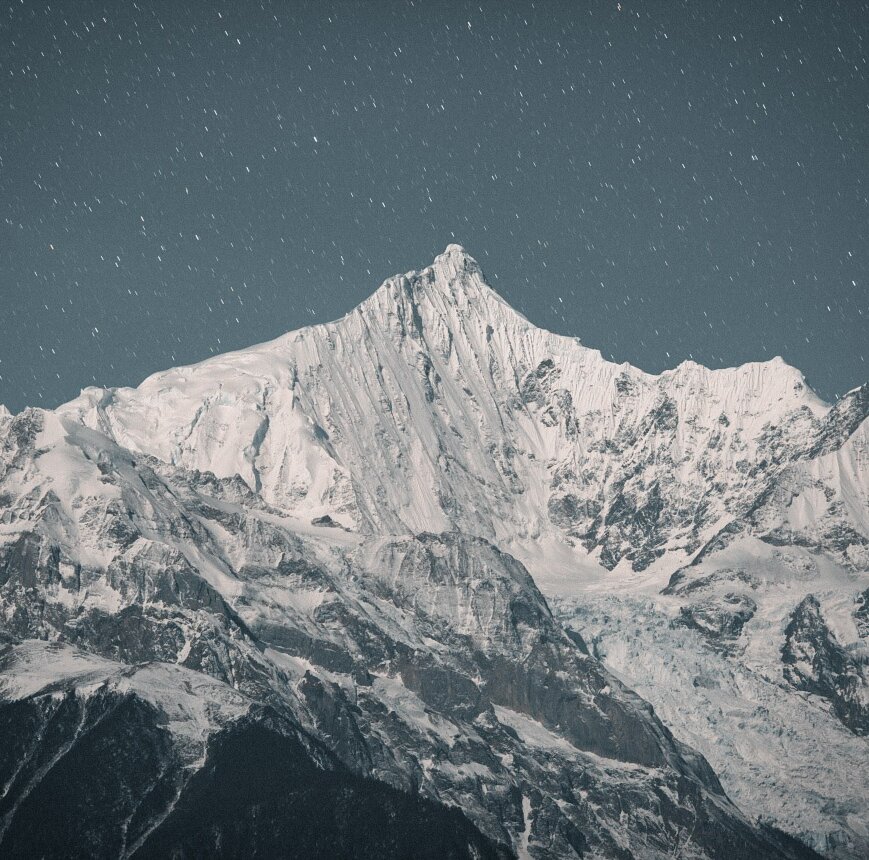} &
        \qualimg{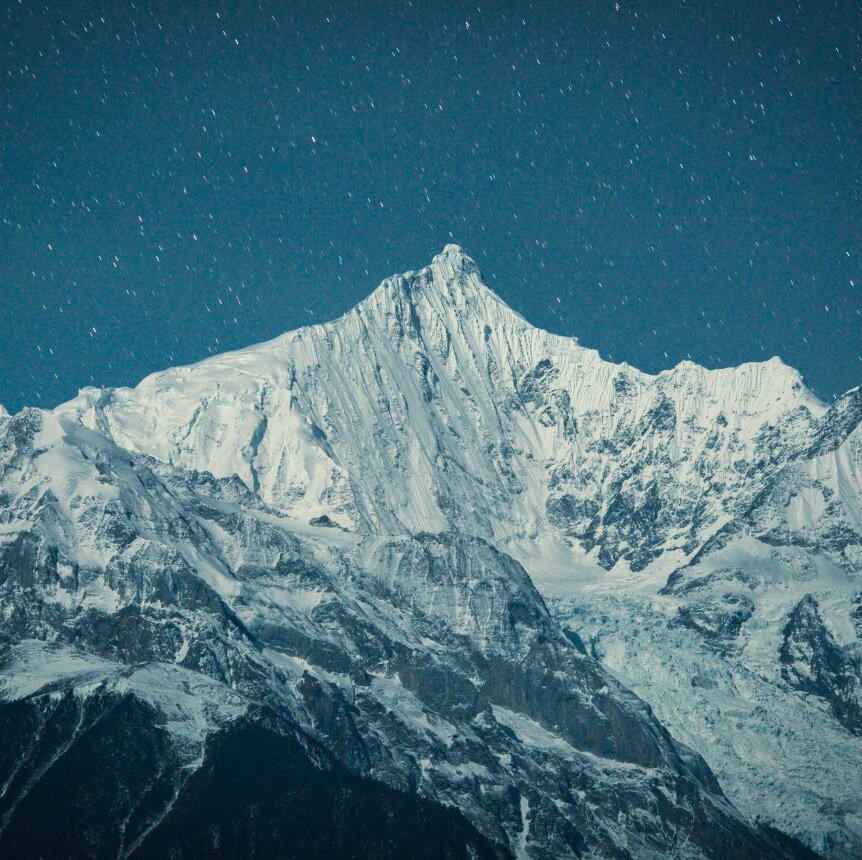} &
        \qualimg{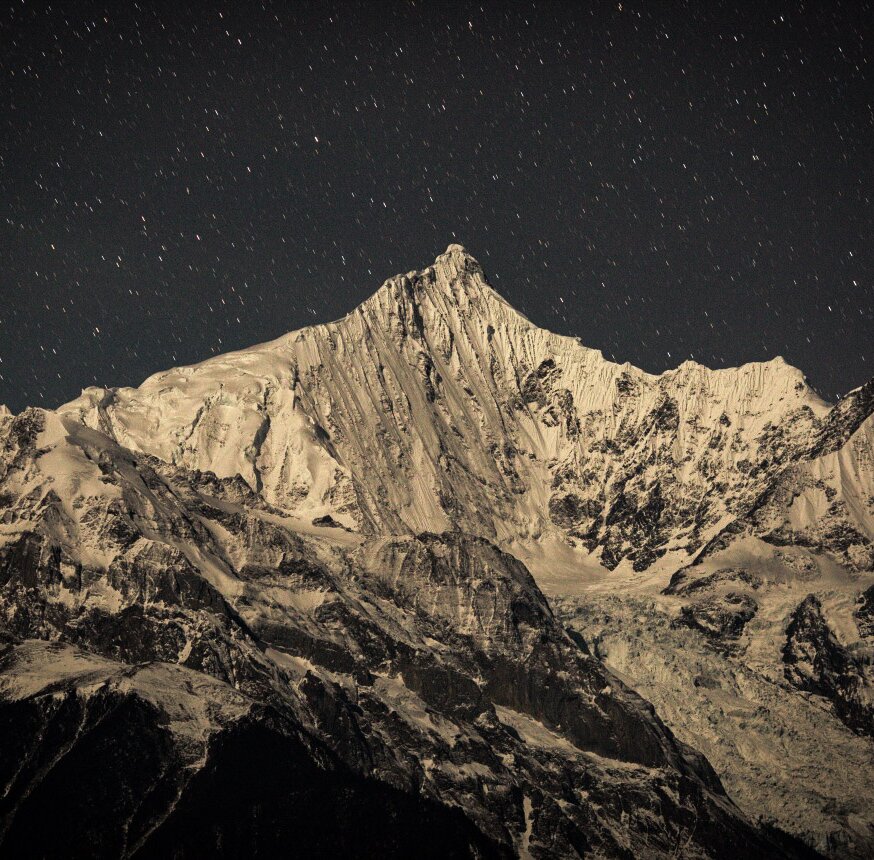} &
        \qualimg{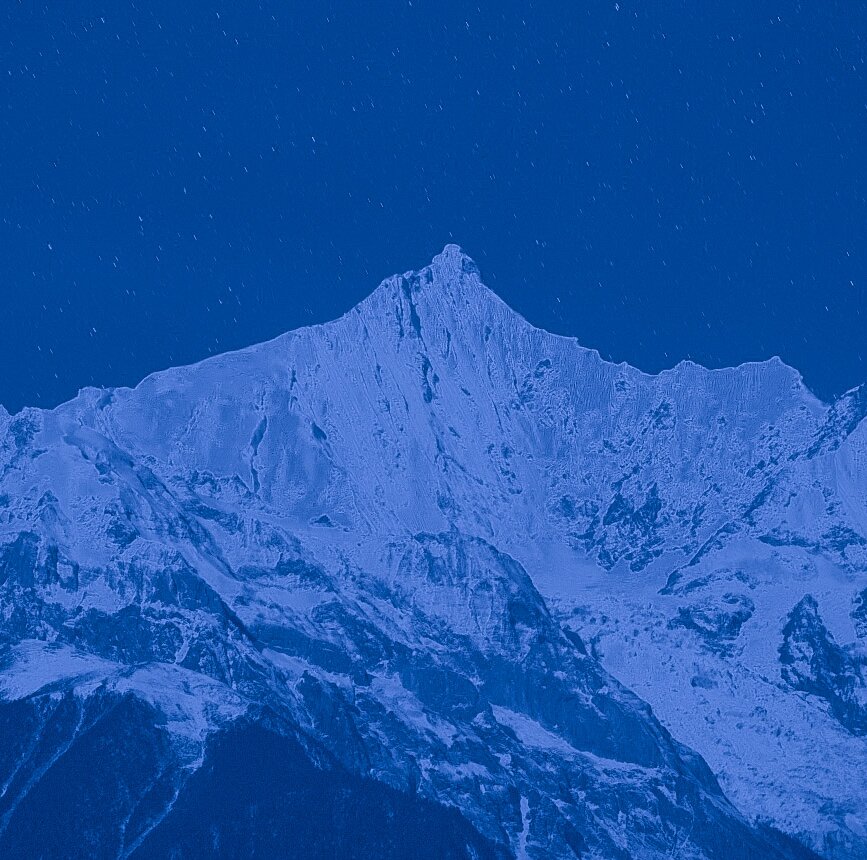} &
        \qualimg{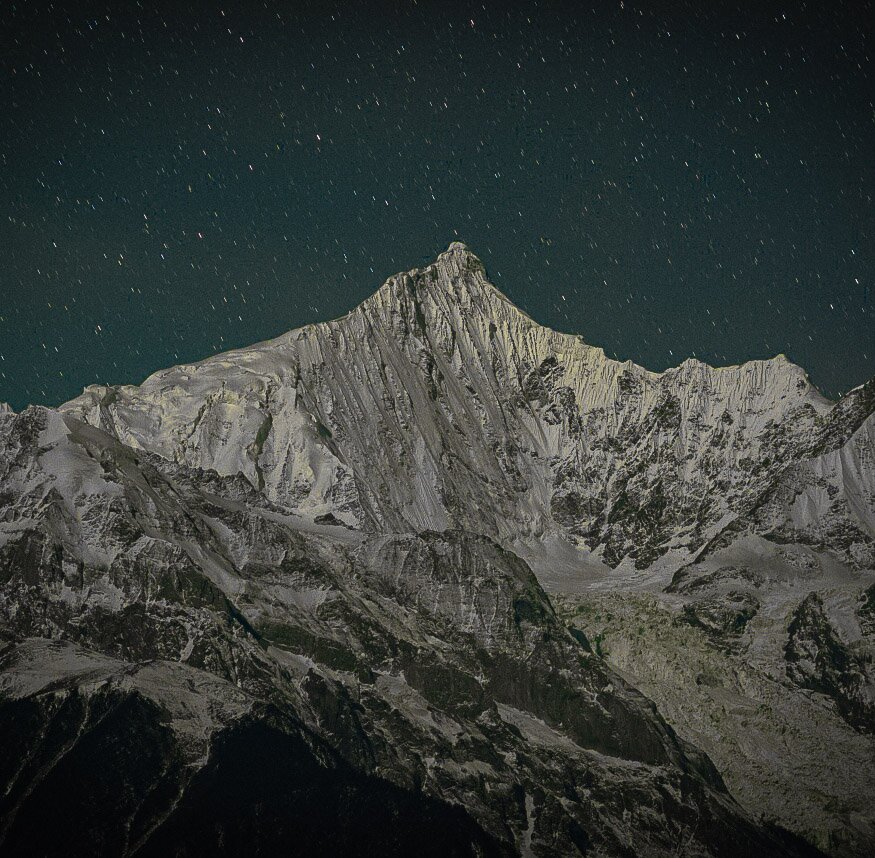}  \\
        \qualinst{\input{images/qual4/instruction}} \\[3pt]

        \qualimgcrop{0.15}{0.8}{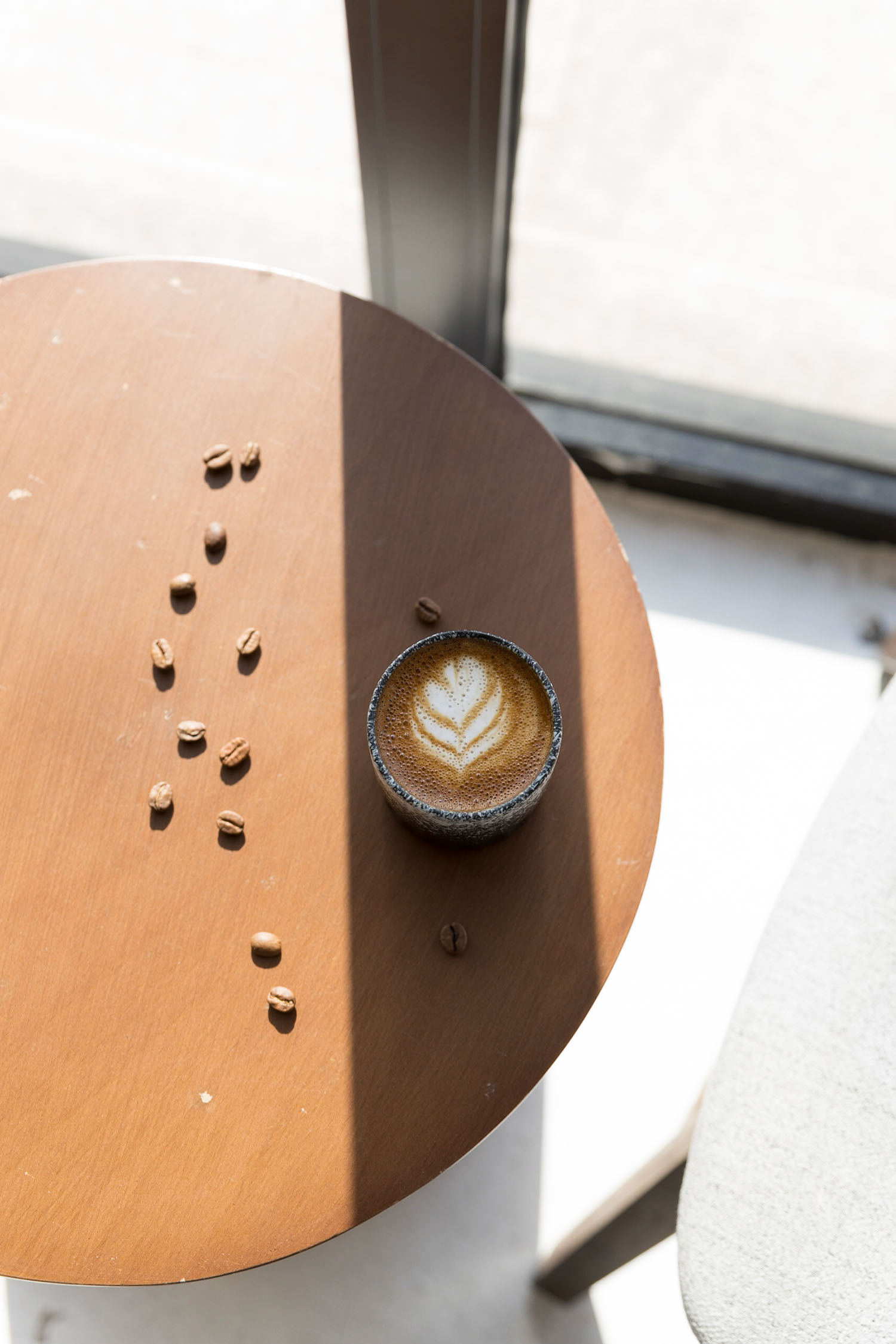} &
        \qualimgcrop{0.15}{0.8}{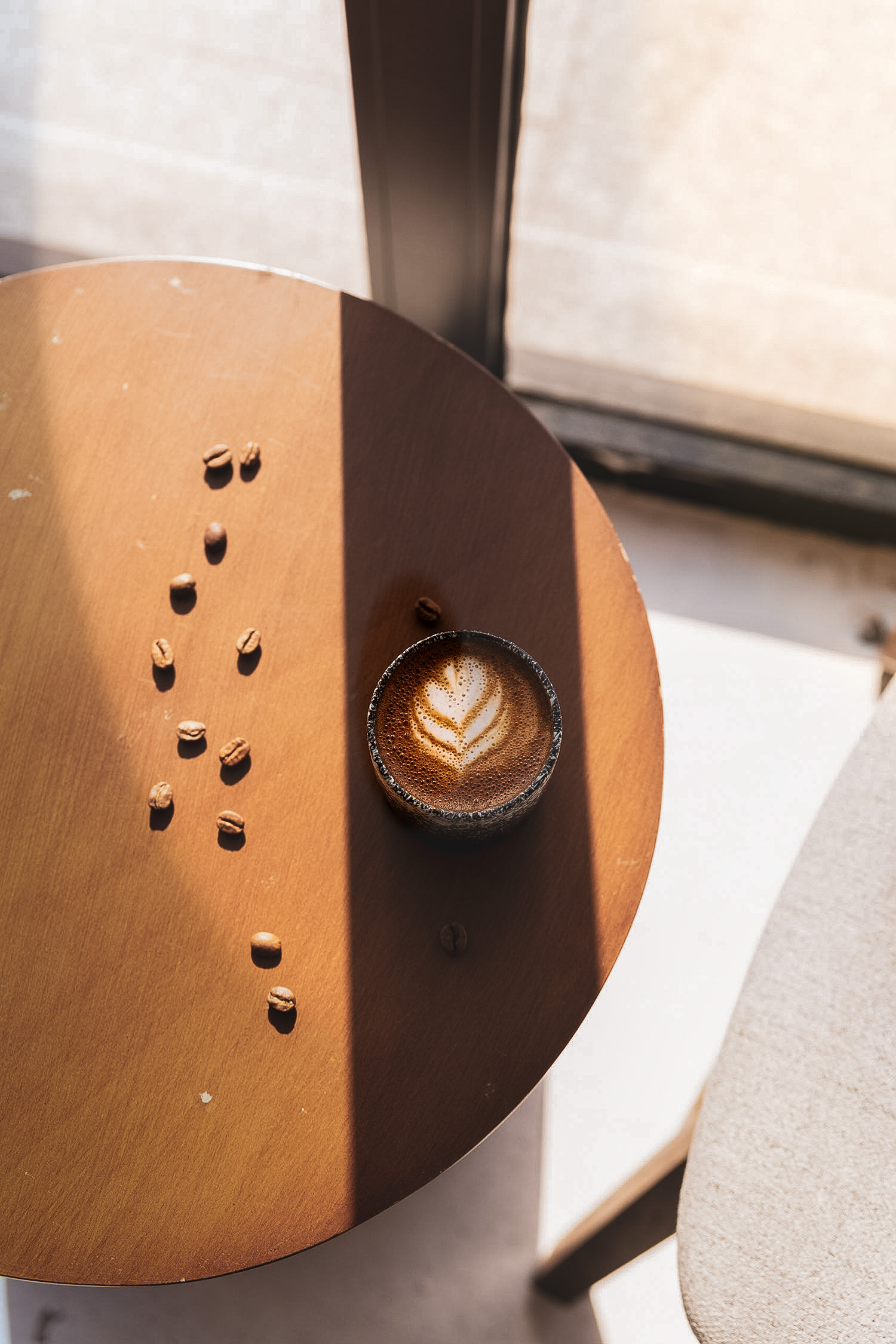} &
        \qualimgcrop{0.15}{0.8}{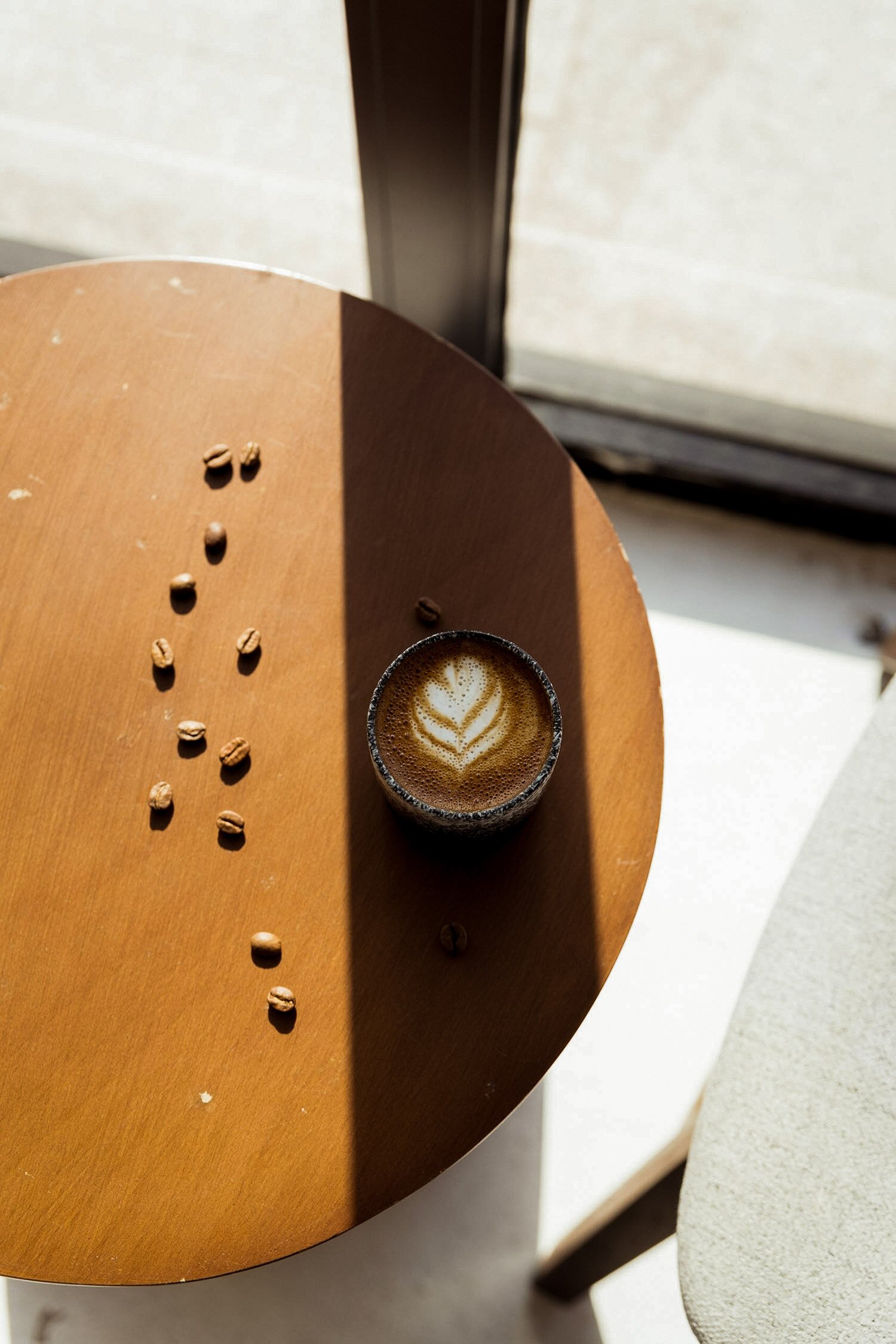} &
        \qualimgcrop{0.15}{0.8}{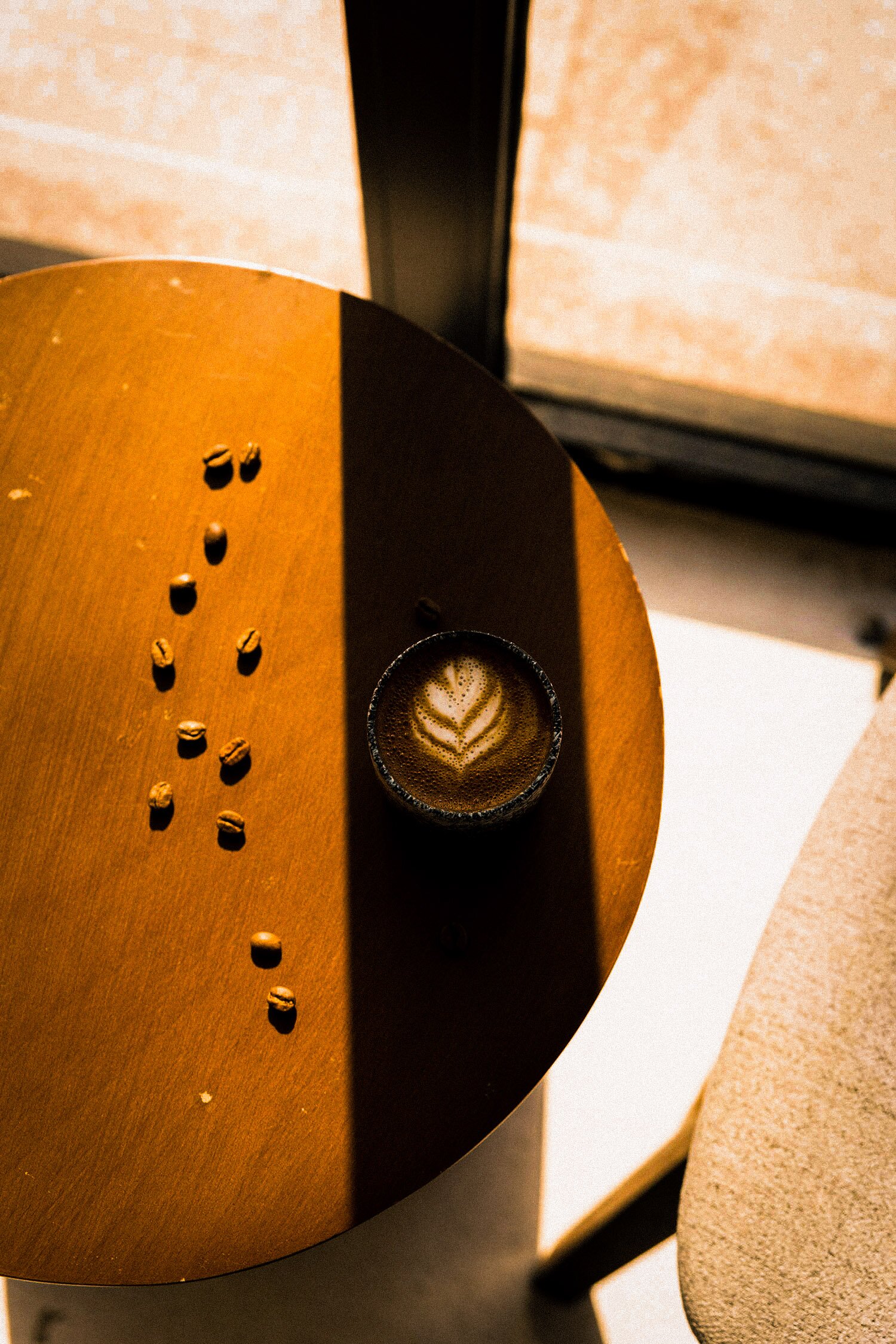} &
        \qualimgcrop{0.15}{0.8}{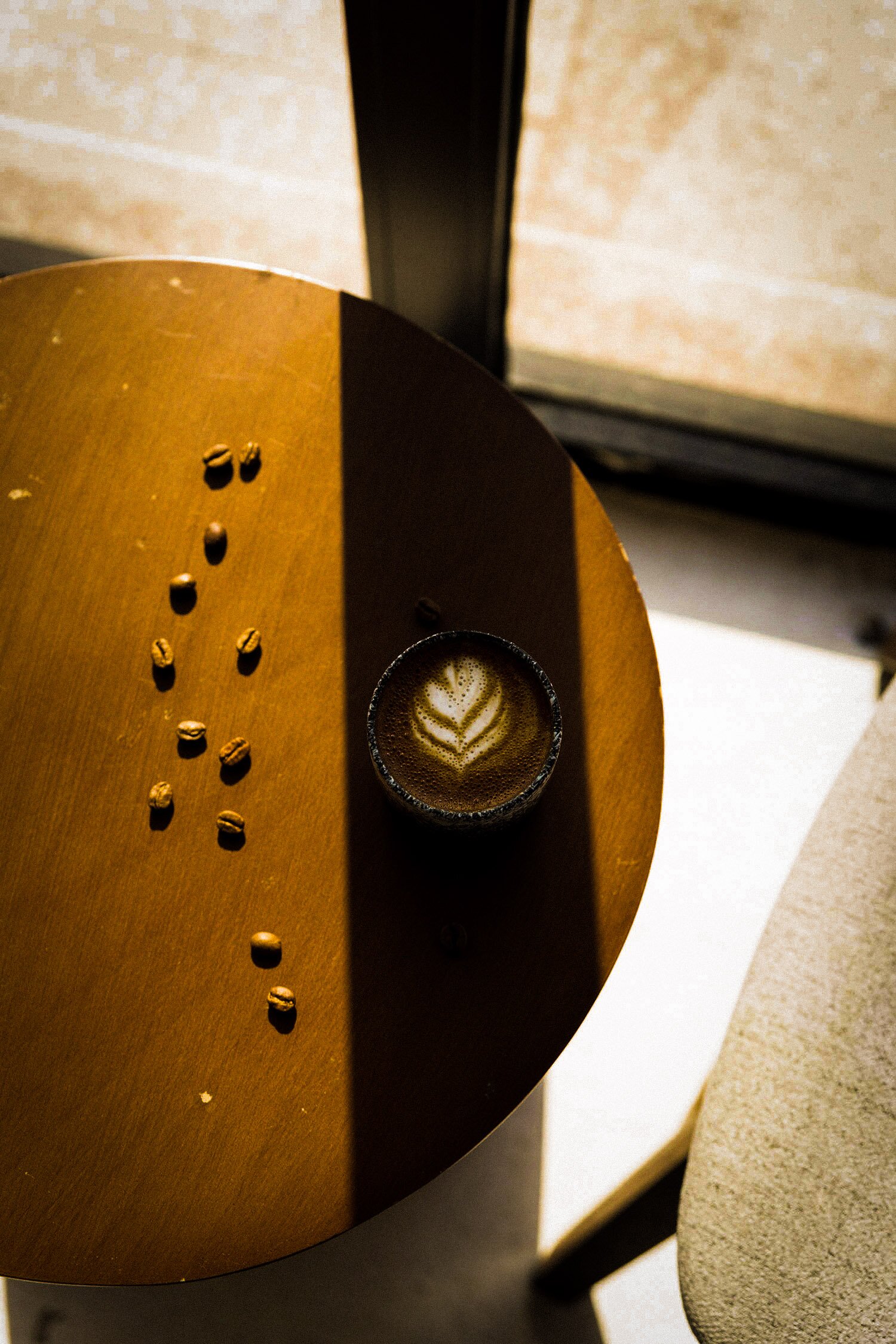} &
        \qualimgcrop{0.15}{0.8}{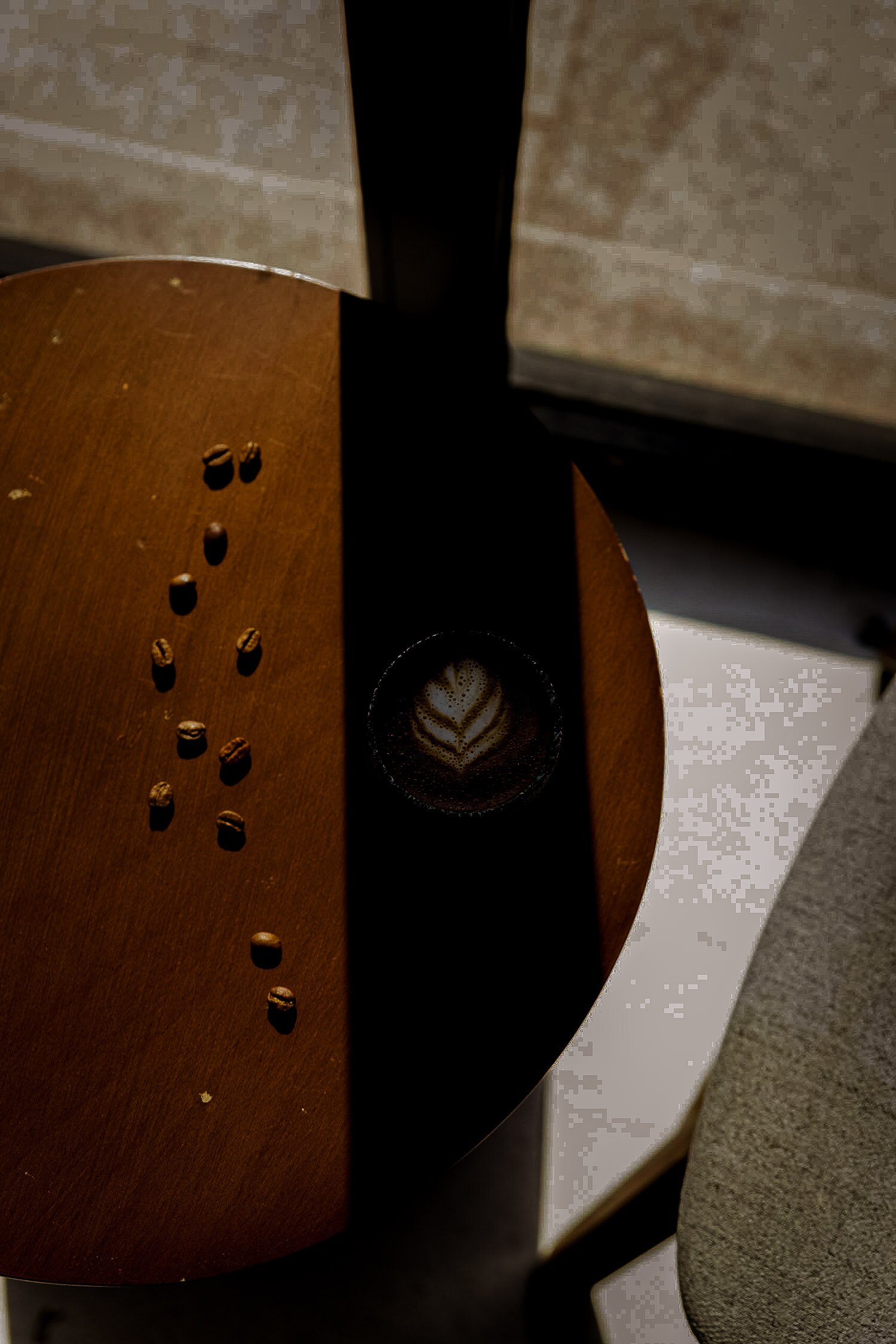} &
        \qualimgcrop{0.15}{0.8}{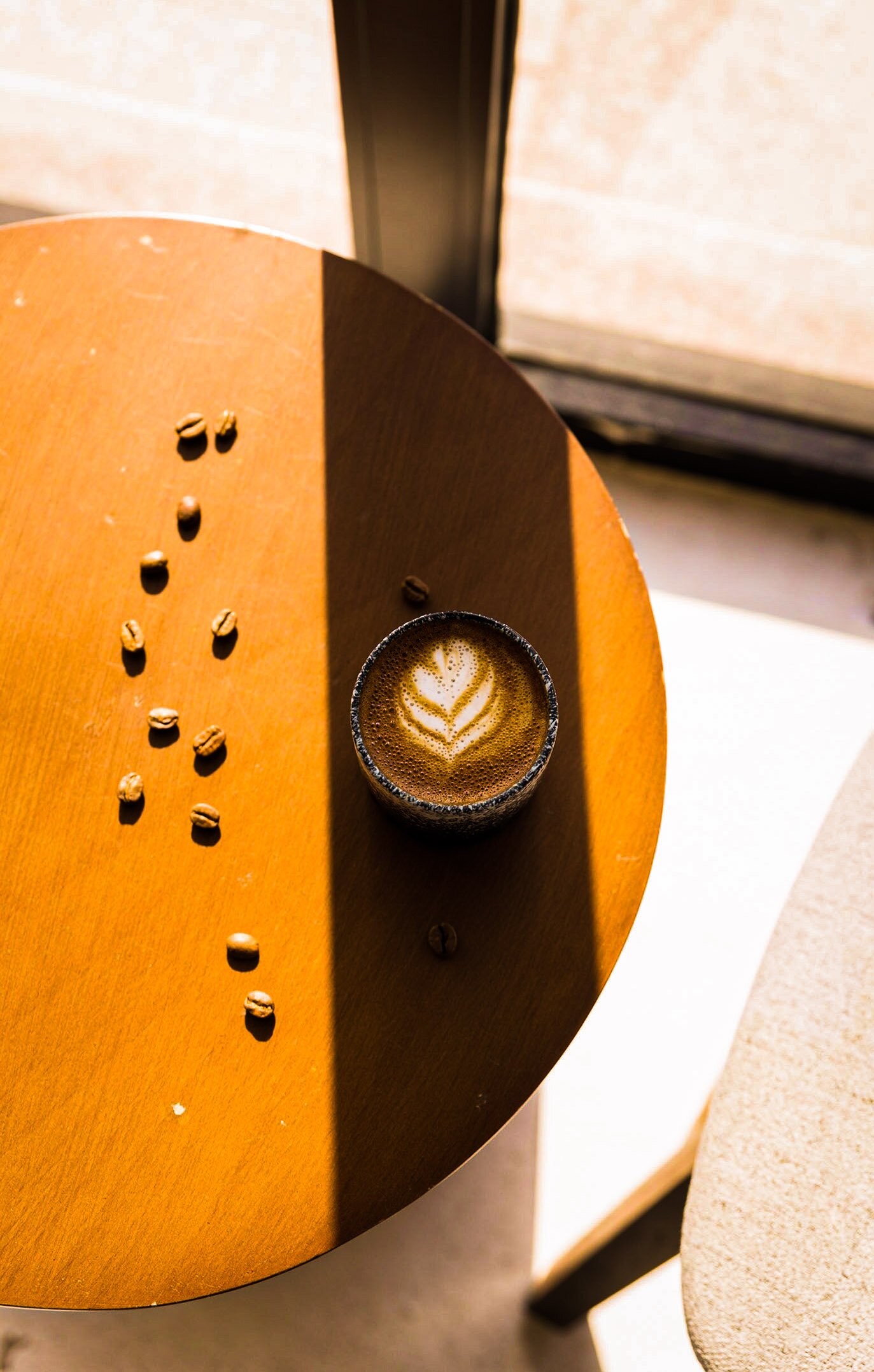}  \\
        \qualinst{\input{images/qual6/instruction}} \\
    \end{tabular}
    \end{adjustbox}
    \vspace{-5pt}
    \caption{\textbf{Qualitative comparison between different methods}. For each example, we show the input image, the user edit, and the outputs produced by FlowTool, proprietary MLLMs, and specialized MLLM agents.  The editing instruction for each example is shown below each row.}
    \vspace{-10pt}
\label{fig:qualitative_comparison}
\end{figure}
\textbf{Quantitative Comparison.}
As shown in \cref{tab:main_results}, FlowTool consistently outperforms specialized MLLM editing agents across all benchmarks, with the largest gains on reference-based metrics: compared with the strongest specialized baseline, it reduces $L_1$ by up to $28.8\%$ and $L_2$ by up to $47.4\%$. FlowTool also surpasses the frontier models, reducing $L_1$ and $L_2$ by up to $13.0\%$ and $21.7\%$, respectively, while achieving stronger perceptual quality (PQ) and remaining competitive in semantic consistency (SC). Reward-based post-training further improves instruction alignment, with FlowTool-RL increasing SC and the overall score across all three instruction-driven benchmarks. On MIT-Adobe5K, FlowTool outperforms all baselines.

\textbf{Qualitative Analysis.}
Fig.~\ref{fig:qualitative_comparison} reveals clear differences in the editing behavior of the compared methods. Specialized MLLM agents tend to apply extremely aggressive transformations, often producing excessive changes in exposure, color, or contrast that deviate from the intended edit. This behavior is particularly evident for JarvisArt in the frog example, where severe overexposure essentially destroys the image content. Similar tendencies can be observed in other examples, where the specialized agents produce overly saturated, excessively dark, or strongly color-shifted outputs. Gemini~3.1 Pro generally produces more reasonable edits than the specialized agents, but its behavior remains less consistent across instructions; for example, some outputs deviate noticeably from the requested color tone or editing intensity. In contrast, FlowTool produces edits that closely resemble the corresponding human edits while faithfully following the requested changes in tone, color, contrast, and overall appearance.
Our outputs preserve the underlying image structure and avoid the extreme transformations observed in the specialized agents.
Finally, FlowTool achieves visual quality comparable to GPT-5.6 Sol across these examples, consistent with the quantitative findings.

\begin{figure}[t]
    \centering
    \begin{minipage}[t]{0.61\textwidth}
    \centering
    \includegraphics[width=\linewidth]{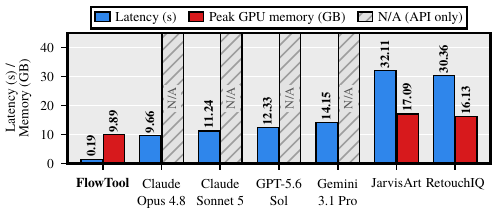}
    \vspace{-18pt}
    \caption{Comparison of inference latency and peak GPU memory across 200 examples. Latency measures editing-plan generation for FlowTool and specialized MLLMs, and end-to-end API response time for general MLLMs. GPU memory is reported only for models deployed locally.}
    \label{fig:efficiency_comparison}
    \end{minipage}\hfill
    \begin{minipage}[t]{0.37\textwidth}
    \centering
    \includegraphics[width=\linewidth]{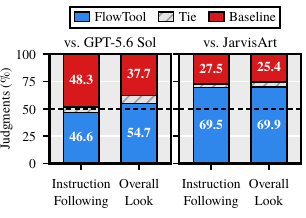}
    \vspace{-18pt}
    \caption{Human preferences for FlowTool versus GPT-5.6 Sol and JarvisArt on two aspects. Bars show the percentage of judgments favoring each system or indicating a tie.}
    \label{fig:user_preference}
    \end{minipage}
\end{figure}

\textbf{Efficiency Comparison.}
We randomly sample 200 examples in total, with 50 examples drawn from each evaluation benchmark, and perform single-sample inference on a NVIDIA A100 GPU. For locally deployed models, we measure the time required to generate the editing plan and record peak GPU memory usage during inference. Proprietary MLLMs are accessed through their APIs under the same evaluation setup; therefore, we report end-to-end API response time as latency, while their GPU memory usage is unavailable. As shown in \cref{fig:efficiency_comparison}, FlowTool is consistently faster than all compared methods, achieving over $50\times$ lower latency than the fastest baselines, while requiring nearly $2\times$ lower peak GPU memory than the specialized MLLM agents. Importantly, these efficiency gains are achieved while maintaining strong editing performance across our benchmarks, highlighting the computational advantage of directly generating structured editing parameters with FlowTool rather than relying on large autoregressive MLLMs and long reasoning trajectories.

\begin{wraptable}{r}{0.62\textwidth}
\vspace{-12pt}
\centering
\caption{User preference results, summarized with individual judgment counts and preferences by case. $p$ denotes the two-sided sign-test $p$-value for preference result by case.}
\label{tab:user_study}
\vspace{-6pt}
\small
\setlength{\tabcolsep}{3pt}
\resizebox{\linewidth}{!}{%
\begin{tabular}{ll ccc c@{\hspace{4pt}}c@{\hspace{4pt}}c c}
\toprule
\multirow{2}{*}{\textbf{Competitor}} & \multirow{2}{*}{\textbf{Criterion}} & \multicolumn{3}{c}{\textbf{Judgments}} & \multicolumn{4}{c}{\textbf{Preferences by Case}} \\
\cmidrule(lr){3-5} \cmidrule(lr){6-9}
 &  & \textbf{Win} & \textbf{Tie} & \textbf{Loss} & \textbf{Win} & \textbf{Tie} & \textbf{Loss} & $\boldsymbol{p}$ \\
\midrule
\multirow{2}{*}{GPT-5.6 Sol}
& Instruction Following & 246 & 27 & 255 & 9 & 2 & 13 & 0.52 \\
& Overall Look     & 289 & 40 & 199 & 16 & 1 & 7 & 0.09 \\
\midrule
\multirow{2}{*}{JarvisArt}
& Instruction Following & 367 & 16 & 145 & 20 & 0 & 4 & 0.002 \\
& Overall Look     & 369 & 25 & 134 & 19 & 0 & 5 & 0.007 \\
\bottomrule
\vspace{-16pt}

\end{tabular}%

}
\end{wraptable}

\textbf{User Preference Study.}
We conduct a user preference study with 22 participants across 24 editing cases, comparing FlowTool with GPT-5.6 Sol and JarvisArt. For each comparison, participants view the instruction, the input image, and two anonymized edits presented in random order, then separately assess instruction following and overall look, with an option to indicate no preference. This results in 528 judgments per competitor and criterion. As shown in \cref{fig:user_preference}, participants prefer FlowTool over JarvisArt in approximately 70\% of judgments on both criteria. When preferences are aggregated by case, FlowTool is favored in 20 of 24 cases for instruction following and 19 of 24 for overall look; both advantages are statistically significant under two-sided sign tests (\cref{tab:user_study}). Agreement across users is also strong: 21 of 22 participants favor FlowTool for instruction following, and all 22 favor it for overall look. Against GPT-5.6 Sol, FlowTool receives more preferences for overall look (54.7\% vs.\ 37.7\%), while instruction-following preferences are closely matched (46.6\% vs.\ 48.3\%).
Participants' written rationales highlight FlowTool's natural colors, balanced tones, and preservation of image detail as reasons for their choices. These results provide direct human evidence of FlowTool's editing quality, demonstrating a clear preference over JarvisArt and competitive performance against GPT-5.6 Sol. The full study protocol, detailed results, and analysis of user preference study are provided in Appendix~\ref{app:user}.
\subsection{Ablation Study}
To analyze the effect of FlowTool's individual components and design choices, we conduct all ablation experiments on MMArt-Bench. We also include additional ablation experiments in Appendix~\ref{app:more_ablation}, where we compare flow matching with deterministic regression and control for differences in training data using an autoregressive Qwen3.5-4B baseline trained on the same dataset.

\begin{wrapfigure}[11]{r}{0.43\linewidth}
\centering
\vspace{-1.0\baselineskip}              

\vspace{-3pt}
\includegraphics[width=\linewidth]{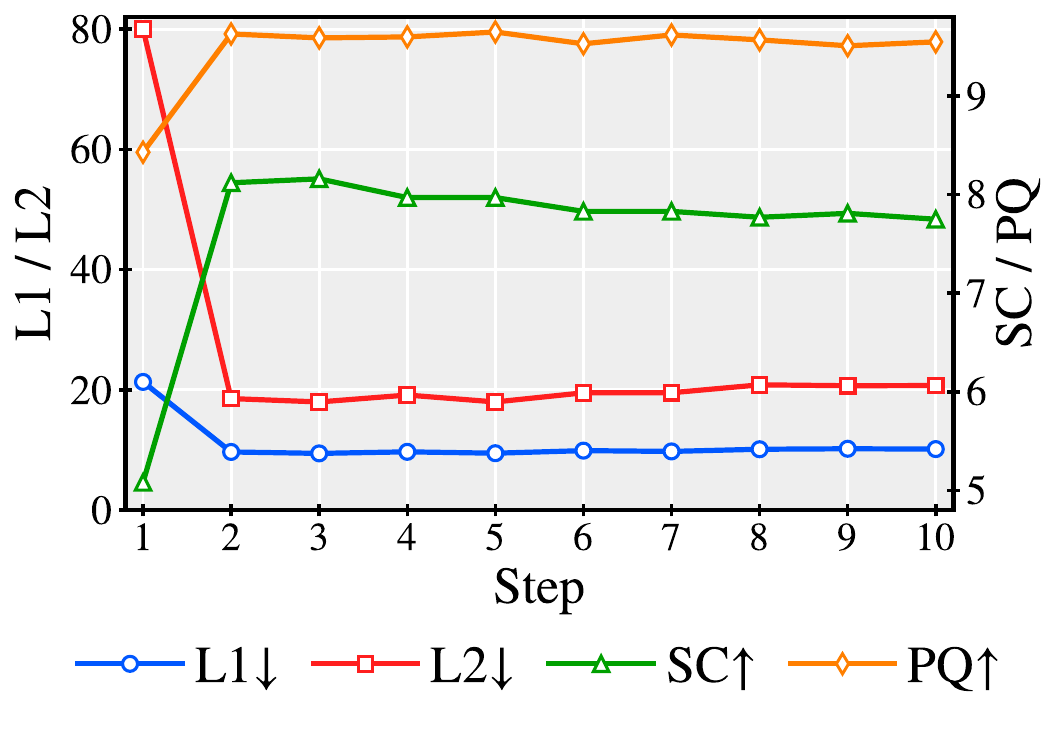}
\vspace{-20pt}
\caption{Effect of \#ODE steps.}
\label{fig:flowtool_step_analysis}
\end{wrapfigure}
\textbf{Effect of ODE Integration Steps.}
\cref{fig:flowtool_step_analysis} shows that increasing the number of steps from one to two substantially improves performance across reference-based and reference-free metrics. 
Performance improves at three steps, providing the best balance across $L_1$, $L_2$, SC, and PQ.
Increasing the number of integration steps beyond this point brings little additional benefit.
These results indicate that FlowTool can accurately generate editing parameters with only a few ODE steps, whereas a single step does not yield a sufficiently good editing plan.

\begin{wraptable}{r}{0.45\textwidth}
\centering
\caption{Ablation of the tool-presence head.}
\vspace{-8pt}
\label{tab:ablation_mask}
\small
\resizebox{\linewidth}{!}{%
\begin{tabular}{lccccc}
    \toprule
    \textbf{Setting}
    & \textbf{L1} $\downarrow$
    & \textbf{L2} $\downarrow$
    & \textbf{SC} $\uparrow$
    & \textbf{PQ} $\uparrow$
    & \textbf{O} $\uparrow$ \\
    \midrule
    w/o Head
    & 9.56 & 18.37 & 7.04 & \best{9.61} & 8.07 \\
    w/ Head
    & \best{9.37} & \best{17.78} & \best{7.96} & 9.59 & \best{8.69} \\
    \bottomrule
\end{tabular}%
}
\end{wraptable}

\textbf{Tool-Presence Modeling.}
\cref{tab:ablation_mask} shows that removing the tool-presence head degrades performance, for both $L_1$ and $L_2$, while causing a pronounced drop in SC and the overall score.
This demonstrates that explicitly modeling which parameters should be activated is important for producing accurate editing plans.

\begin{wraptable}{r}{0.52\textwidth}
\vspace{-12pt}
\centering
\caption{Ablation of the two-stage SFT curriculum.}
\vspace{-8pt}

\label{tab:ablation_training_strategy}
\small
\resizebox{\linewidth}{!}{%
\begin{tabular}{lccccc}
    \toprule
    \textbf{SFT Schedule}
    & \textbf{L1} $\downarrow$
    & \textbf{L2} $\downarrow$
    & \textbf{SC} $\uparrow$
    & \textbf{PQ} $\uparrow$
    & \textbf{O} $\uparrow$ \\
    \midrule
    Stage 1 only
    & 9.63 & 18.92 & 7.73 & 9.55 & 8.50 \\
    Stage 2 only
    & 9.90 & 20.10 & 5.99 & \best{9.69} & 7.42 \\
    Stage 1 $\rightarrow$ Stage 2
    & \best{9.37} & \best{17.78} & \best{7.96} & 9.59 & \best{8.69} \\
    \bottomrule
\end{tabular}
}
\vspace{-8pt}
\end{wraptable}
\textbf{Two-Stage SFT Curriculum.}
As shown in \cref{tab:ablation_training_strategy}, jointly training all components from the beginning yields substantially worse performance, particularly for SC and the overall score. We attribute this degradation to the optimization imbalance between the pretrained VLM backbone and the randomly initialized DiT parameter generator and tool-presence head, whose unstable early updates can interfere with the pretrained multimodal representations. Training only the randomly initialized components, while keeping the VLM frozen, already results in stronger and more stable performance. Our full two-stage curriculum performs best overall: it first establishes reliable parameter generation and tool-presence prediction with the VLM frozen, and only then adapts the VLM through joint training, enabling its pretrained representations to specialize toward fine-grained editing decisions.

\begin{wraptable}{r}{0.52\textwidth}
\centering
\vspace{-14pt}
\caption{
Ablation of VLM backbone scale.
}
\vspace{-8pt}
\label{tab:ablation_backbone_size}
\small

\resizebox{\linewidth}{!}{%
\begin{tabular}{lccccc}
\toprule
\textbf{VLM Backbone}
& \textbf{L1} $\downarrow$
& \textbf{L2} $\downarrow$
& \textbf{SC} $\uparrow$
& \textbf{PQ} $\uparrow$
& \textbf{O} $\uparrow$ \\
\midrule
Qwen3.5-0.8B
& 9.58 & 18.65 & 7.27 & 9.61 & 8.25 \\
Qwen3.5-4B
& \best{9.37} & \best{17.78} & \best{7.96} & 9.59 & \best{8.69} \\
Qwen3.5-9B
& 9.44 & 18.20 & 7.54 & \best{9.63} & 8.44 \\
\bottomrule
\end{tabular}%
}
\vspace{-8pt}
\end{wraptable}
\textbf{Scaling the Backbone.}
\cref{tab:ablation_backbone_size} shows that scaling the VLM backbone from 0.8B to 4B generally improves performance, with Qwen3.5-4B achieving the strongest overall results. Further scaling the backbone to 9B provides no additional improvement and slightly degrades most metrics. We therefore adopt Qwen3.5-4B as our default backbone.
 
\section{Conclusion}

We introduced \textbf{FlowTool}, a novel approach to tool-based image editing that replaces autoregressive reasoning with direct structured tool-parameter generation through conditional flow matching. Extensive experiments across multiple tool-based image editing benchmarks demonstrate that FlowTool achieves strong editing performance, outperforming specialized MLLM agents while remaining competitive with frontier proprietary MLLMs.  At the same time, FlowTool provides significantly more efficient inference in terms of both latency and memory usage. 
Overall, our results demonstrate that effective tool-based image editing can be achieved without long autoregressive reasoning trajectories, highlighting conditional flow matching as a promising alternative for efficient and controllable multimodal tool use.

\clearpage

\subsection*{AI Use Statement}

We used large language models (LLMs) in several supporting tasks of this work. 
We used LLMs as assistants during codebase development for coding and debugging. 
All AI-assisted code was reviewed, tested, and validated by the authors. 
We also used LLMs to assist with paraphrasing, grammatical correction, and improving the clarity and readability of our writing.

The research ideas, methodology, experimental design, analysis, scientific claims, and the paper's overall structure and content were developed and determined by the authors. LLMs were not used to autonomously formulate scientific contributions or draw conclusions from experimental results. All AI-assisted outputs were reviewed and, where necessary, revised by the authors. We take full responsibility for the final content of the paper and all associated artifacts.

\subsection*{Ethics Statement}
We use licensed open-source and private datasets. Our work includes a user preference study with 22 human participants across 24 editing cases, evaluating instruction following and overall visual quality. Participants compare model-generated edits with model identities anonymized and presentation order randomized to reduce evaluation bias.

\subsection*{Reproducibility statement}

\cref{sec:prelim_flow_matching,sec:method} describe the method including conceptual design, mathematical formulation. 
\cref{fig:overall,fig:flowtool_architecture} illustrate the model architecture.
\cref{sec:data_preparation} describes the data preparation process.
Computational cost accounting is presented in \cref{app:implementation_details}.
Accessing to datasets, model checkpoints, and code are subjected to internal approval.

\bibliography{iclr2027_conference}
\bibliographystyle{iclr2027_conference}
\clearpage
\appendix

\section{Limitations \& Future Works}
\label{app:limitation}

FlowTool is currently designed for standardized linear parameter spaces, such as parameters normalized to $[-100,100]$, extending the framework to other parameter structures remains an important direction for future work. In particular, many image-editing and generation systems involve cyclic or polar-valued parameters, such as hue, as well as discrete or categorical parameters, such as textual options. Adapting the model to handle these heterogeneous parameter spaces could further broaden its applicability to more general editing tools and interfaces.
\section{Implementation Details}
\label{app:implementation_details}

\textbf{Training Compute.}
Supervised flow-matching training was conducted on a single NVIDIA A100 80GB GPU for approximately 46 hours. Reward-based post-training was performed on a single node with 8 NVIDIA A100 80GB GPUs for approximately 40 hours.

\textbf{Model Architecture.}
We use Qwen3.5-4B as the VLM backbone (0.8B and 9B variants are used in the backbone-scale ablation of \cref{tab:ablation_backbone_size}). The DiT-based tool parameter generator consists of 12 transformer blocks with hidden dimension $1024$ and 16 attention heads. The flow timestep is represented using a $256$-dimensional embedding. Following our structured parameter formulation, we tokenize the editing plan on a per-parameter basis, such that each tool parameter is represented by an individual token before being processed by the DiT. We use no dropout in the DiT blocks. The tool-presence head is enabled throughout training and predicts the active subset of the $D$ editing parameters from the multimodal representation. The VLM is operated in bfloat16 precision with scaled dot-product attention.

\textbf{Supervised Flow-Matching Training.}
We train the SFT model for 4 epochs using a batch size of 64. We use a learning rate of $1\times10^{-4}$ for the newly initialized modules and $2\times10^{-4}$ for the trainable VLM adapters, together with a cosine learning-rate schedule. Flow timesteps are sampled from a Beta distribution with parameters $\alpha=1.5$ and $\beta=1.0$. The tool-presence loss is weighted by $\lambda_{\mathrm{pres}}=1$. Following the two-stage curriculum described in \cref{sec:sft}, we keep the VLM frozen for the first 3 epochs and optimize only the DiT parameter generator and tool-presence head. For the remaining epochs, we enable LoRA adaptation of the VLM and jointly optimize all trainable components. We use LoRA rank $16$, scaling factor $32$, and dropout $0.05$, applied to the query and value projection layers. All input images are resized to $448\times448$.

\textbf{Reward-Based Post-Training.}
We initialize reward-based post-training from the best SFT checkpoint and optimize the model for 1,500 iterations using a learning rate of $1\times10^{-5}$ with a constant learning-rate schedule.
For each update, we sample 8 image--instruction prompts and generate a group of 8 editing plans per prompt. For the reward in \cref{eq:composite_render_reward}, we assign equal weights to the reference-based and VLM-based components, with $\lambda_{\mathrm{ref}}=\lambda_{\mathrm{vlm}}=1$, and average the two reward components after normalization. We use Qwen3.8-27B as the VLM judge with temperature $0$. For DiffusionNFT, we set $\beta=1$, use a KL regularization weight of $10^{-3}$, update the old policy every 10 iterations with decay $0.9$, and perform one inner optimization epoch per rollout group.
\section{More Ablation Experiments}
\label{app:more_ablation}

We conduct two comparisons using baselines with the same Qwen3.5-4B backbone and our curated training data as FlowTool (\cref{sec:experimental_settings}).

\textbf{Flow Matching vs.\ Deterministic Regression.}
We replace FlowTool's flow-matching generator with a deterministic regression head that directly predicts continuous parameters, using the same pretrained backbone and training dataset. Unlike the one-step ODE ablation in \cref{fig:flowtool_step_analysis}, this baseline is trained independently with a regression objective. \cref{tab:same_data} shows that FlowTool outperforms it on all metrics across all benchmarks, demonstrating better reference similarity, instruction following, and overall editing quality. Since both methods produce continuous parameters, these results support the benefit of flow-based modeling over deterministic regression beyond simply avoiding numerical text generation.

\begin{wraptable}{r}{0.42\textwidth}
\centering
\caption{Number of succesful editing plan extraction by autoregressive baseline.}
\label{tab:ar_valid}
\vspace{-6pt}
\small
\begin{tabular}{lc}
\toprule
\textbf{Benchmark} & \textbf{Succesful Plans} \\
\midrule
MMArt-Bench & 166 / 202 (82.2\%) \\
FlowTool-Eval & 188 / 300 (62.7\%) \\
ArtEdit-Bench & 265 / 333 (79.6\%) \\
\bottomrule
\end{tabular}
\vspace{-10pt}
\end{wraptable}
\textbf{Controlling for Training Data.}
We fine-tune Qwen3.5-4B to generate tool selections and parameter values as text using a next-token prediction objective. This baseline shares FlowTool's pretrained backbone and training data, allowing us to assess the contribution of the generation formulation. As shown in \cref{tab:same_data}, FlowTool substantially outperforms this baseline, demonstrating that its gains cannot be explained by training data alone and supporting the effectiveness of flow-based parameter generation over autoregressive prediction. Furthermore, formatting errors prevent the autoregressive baseline from producing executable editing plans in a substantial number of cases (\cref{tab:ar_valid}), highlighting an additional reliability challenge of generating parameters as text.

\begin{table}[h]
\centering
\caption{Comparison with a regression model and an autoregressive model (both on Qwen3.5-4B backbone) trained on the same training data as FlowTool. Metrics follow \cref{tab:main_results}.}
\label{tab:same_data}
\footnotesize
\setlength{\tabcolsep}{2.0pt}
\renewcommand{\arraystretch}{1.1}
\resizebox{\textwidth}{!}{%
\begin{tabular}{l ccccc ccccc ccccc}
\toprule
\multirow{2}{*}{\bf Method}
& \multicolumn{5}{c}{\textbf{MMArt-Bench}}
& \multicolumn{5}{c}{\textbf{FlowTool-Eval}}
& \multicolumn{5}{c}{\textbf{ArtEdit-Bench}}
\\
\cmidrule(lr){2-6}
\cmidrule(lr){7-11}
\cmidrule(lr){12-16}
&
\textbf{L1} $\downarrow$ & \textbf{L2} $\downarrow$ & \textbf{SC} $\uparrow$ & \textbf{PQ} $\uparrow$ & \textbf{O} $\uparrow$
&
\textbf{L1} $\downarrow$ & \textbf{L2} $\downarrow$ & \textbf{SC} $\uparrow$ & \textbf{PQ} $\uparrow$ & \textbf{O} $\uparrow$
&
\textbf{L1} $\downarrow$ & \textbf{L2} $\downarrow$ & \textbf{SC} $\uparrow$ & \textbf{PQ} $\uparrow$ & \textbf{O} $\uparrow$
\\
\midrule
\textbf{FlowTool}
& \best{9.42} & \best{17.98} & \best{8.16} & \best{9.59} & \best{8.81}
& \best{7.87} & \best{13.21} & \best{8.04} & \best{9.64} & \best{8.76}
& \best{7.65} & \best{12.95} & \best{8.34} & 9.75 & \best{8.99}
\\
\midrule
Qwen3.5-4B-Regression
& 10.22 & 21.39 & 6.67 & 9.50 & 7.80
& 10.21 & 21.81 & 6.64 & 9.61 & 7.80
& 9.26 & 18.24 & 7.27 & 9.72 & 8.25
\\
Qwen3.5-4B-Autoregressive
& 10.16 & 20.73 & 6.03 &  9.58 & 7.41
& 9.38 & 18.76 & 5.66 & 9.52 & 7.08
& 8.54 & 16.25 & 6.37 & \best{9.76} & 7.67
\\
\bottomrule
\end{tabular}%
}
\vspace{-14pt}
\end{table}

\section{User Preference Study}
\label{app:user}

We compare FlowTool with two top-performing baselines from our main experiments: JarvisArt, representing specialized MLLMs, and GPT-5.6 Sol, representing general MLLMs. This appendix expands on the user study presented in \cref{sec:result}, providing details of the study design, preference analyses, and participants' rationale.

\subsection{Study Design}
\begin{wrapfigure}{r}{0.55\textwidth}
\vspace{-25pt}
\centering
\includegraphics[width=\linewidth]{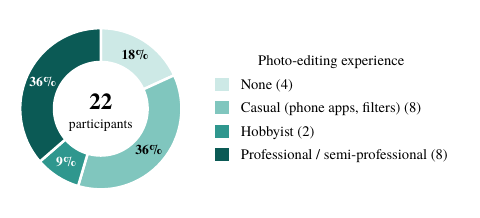}
\vspace{-21pt}
\caption{Self-reported editing experience of users.}
\label{fig:user_study_participants}
\end{wrapfigure}
\textbf{Participants and Editing Tasks.}
The study involves 22 participants and 24 editing cases drawn from the evaluation sets of the four benchmarks. Each case consists of an input image, an editing instruction, and the corresponding outputs from FlowTool, GPT-5.6 Sol, and JarvisArt. Before starting, participants report their photo-editing experience as none, casual (phone apps, filters), regular editing as a hobby, or professional/semi-professional practice. The distribution in \cref{fig:user_study_participants} allows us to examine human preferences and evaluation across different levels of expertise with editing tools.

\textbf{Blinded Pairwise Comparisons.}
Participants complete an online questionnaire containing 48 comparisons: each editing case pairs FlowTool once with GPT-5.6 Sol and once with JarvisArt. Each comparison presents the instruction, the original image, and two outputs labeled \emph{Edit A} and \emph{Edit B}, with model names hidden. Participants are instructed to use a laptop or desktop screen at normal brightness so that color and tonal differences remain visible.

\textbf{Evaluation Criteria.}
For each comparison, participants answer two questions: \emph{Instruction Following} (Which edit better follows the editing request?) and \emph{Overall Look} (Which edit looks better overall?). They must select A, B, or \emph{Tie} for each question and may also provide their explanation and rationale. The two questions assess instruction adherence and visual quality separately: for instance, a strong color or contrast change may fulfill the request while making the image look less natural. All participants answer both questions for every comparison, resulting in 2,112 judgments, or 528 per competitor and criterion.

\subsection{Preference Results}
\textbf{Metrics Explanation.}
\cref{tab:user_study_stats} reports the results at three levels: individual judgments, editing cases, and participants. For the judgments, we count how often participants choose FlowTool (win), neither edit (tie), or the competitor (loss), and define the \emph{gain} as the win share minus the loss share. At the case level, a case counts as a win if more participants choose FlowTool than the competitor, and as a loss if fewer do. The user level is defined in the same way: a participant counts as a win if they choose FlowTool more often than the competitor across the 24 use cases. Because judgments on the same case or from the same participant are not independent, we test significance at the case and user levels with two-sided sign tests that exclude ties.

\begin{table}[h]
\centering
\caption{User preferences by judgment counts, by case, and by user. \emph{Judgments}: number (share) of judgments choosing FlowTool (Win), neither edit (Tie), or the competitor (Loss); Gain is the win share minus the loss share. \emph{Preferences by Case/User}: number of cases or participants in which FlowTool receives more (Win), as many (Tie), or fewer (Loss) judgments than the competitors.}
\label{tab:user_study_stats}
\small
\setlength{\tabcolsep}{3pt}
\resizebox{\textwidth}{!}{%
\begin{tabular}{ll cccc cccc cccc}
\toprule
& & \multicolumn{4}{c}{\textbf{Judgments}} & \multicolumn{4}{c}{\textbf{Preferences by Case}} & \multicolumn{4}{c}{\textbf{Preferences by User}} \\
\cmidrule(lr){3-6} \cmidrule(lr){7-10} \cmidrule(lr){11-14}
\textbf{Competitor} & \textbf{Criterion} & \textbf{Win} & \textbf{Tie} & \textbf{Loss} & \textbf{Gain} & \textbf{Win} & \textbf{Tie} & \textbf{Loss} & $\boldsymbol{p}$ & \textbf{Win} & \textbf{Tie} & \textbf{Loss} & $\boldsymbol{p}$ \\
\midrule
\multirow{2}{*}{GPT-5.6 Sol} & Instruction Following & 246 (46.6\%) & 27 (5.1\%) & 255 (48.3\%) & \drop{1.7} & 9 & 2 & 13 & 0.52 & 7 & 1 & 14 & 0.19 \\
 & Overall Look & 289 (54.7\%) & 40 (7.6\%) & 199 (37.7\%) & \gain{17.0} & 16 & 1 & 7 & 0.09 & 14 & 1 & 7 & 0.19 \\
\midrule
\multirow{2}{*}{JarvisArt} & Instruction Following & 367 (69.5\%) & 16 (3.0\%) & 145 (27.5\%) & \gain{42.0} & 20 & 0 & 4 & 0.002 & 21 & 0 & 1 & $<$0.001 \\
 & Overall Look & 369 (69.9\%) & 25 (4.7\%) & 134 (25.4\%) & \gain{44.5} & 19 & 0 & 5 & 0.007 & 22 & 0 & 0 & $<$0.001 \\
\bottomrule
\end{tabular}%
}
\end{table}

\begin{figure}[h]
\centering
\includegraphics[width=\textwidth]{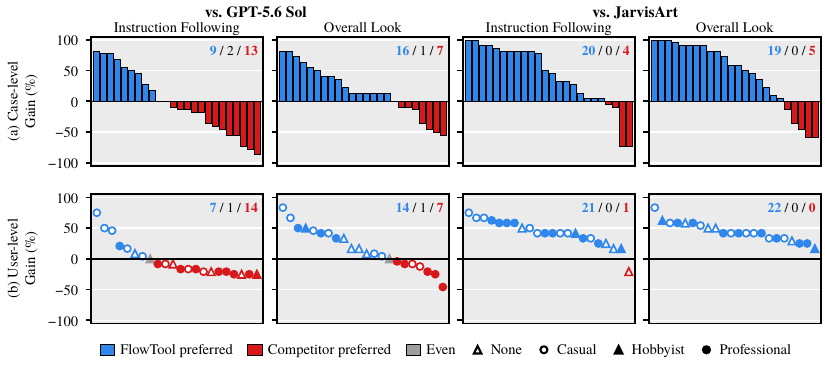}
\vspace{-14pt}
\caption{Case-level (a) and user-level (b) gain, sorted in decreasing order within each panel. Gain is the share of judgments favoring FlowTool minus the share favoring the competitor, so positive values favor FlowTool. The numbers in each panel count the cases or participants with positive / zero / negative gain. Marker shapes in (b) denote the participants' photo-editing experience.}
\label{fig:user_study_breakdown}
\end{figure}

\textbf{Significant preference for FlowTool over JarvisArt.}
FlowTool demonstrates a significant and consistent advantage over JarvisArt in both instruction following and overall look. This preference remains statistically significant when judgments are grouped by case or by user (\cref{tab:user_study_stats}). As shown in \cref{fig:user_study_breakdown}, FlowTool is preferred across most editing cases. The preference is also consistent across users: all participants favor FlowTool for overall look, and all but one favor FlowTool for instruction following.

\textbf{FlowTool remains competitive with GPT-5.6 Sol.}
As shown in~\cref{tab:user_study_stats} and \cref{fig:user_study_breakdown}, instruction-following preferences are closely balanced between the two methods (GPT-5.6 Sol is slightly better), while participants more often favor FlowTool for overall look (54.7\% vs.\ 37.7\%). This preference for FlowTool's visual quality is also reflected when judgments are grouped by case or by user. Although the differences do not reach statistical significance at either level (\(p\geq0.09\)), the results show a consistent trend toward preferring the appearance of FlowTool's edits. In Appendix~\ref{app:user_rationale}, we examine participants' rationales to better understand this pattern, focusing on how they weigh the strength of an edit against its naturalness and visual appeal.

\begin{figure}[h]
\centering
\includegraphics{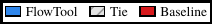}\\[2pt]
\begin{subfigure}[t]{0.48\textwidth}\centering
\includegraphics[width=\linewidth]{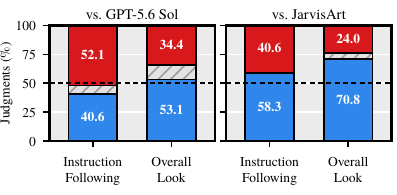}
\caption{No experience ($n=4,j=96$)}
\end{subfigure}\hfill
\begin{subfigure}[t]{0.48\textwidth}\centering
\includegraphics[width=\linewidth]{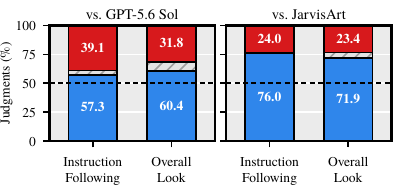}
\caption{Casual ($n=8,j=192$)}
\end{subfigure}\\[4pt]
\begin{subfigure}[t]{0.48\textwidth}\centering
\includegraphics[width=\linewidth]{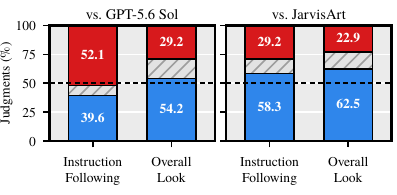}
\caption{Hobbyists ($n=2,j=48$)}
\end{subfigure}\hfill
\begin{subfigure}[t]{0.48\textwidth}\centering
\includegraphics[width=\linewidth]{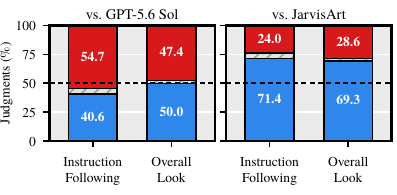}
\caption{Professional ($n=8,j=192$)}
\end{subfigure}
\caption{User preferences by self-reported photo-editing experience. Each bar shows the percentage of judgments favoring FlowTool, the competitor (Baseline), or expressing no preference (Tie), for instruction following or overall look. $n$ denotes the number of participants, and $j$ is the number of judgments per bar.}
\label{fig:user_study_groups}
\end{figure}

\textbf{Preferences by user editing experience.}
\cref{fig:user_study_groups} compares preferences (by judgment counts) across participants with different levels of editing experience. FlowTool is preferred over JarvisArt on both criteria in all four groups, showing broad agreement across experience levels. All groups also favor FlowTool over GPT-5.6 Sol for overall look, although the preference is narrow among professional and semi-professional users. Instruction-following preferences against GPT-5.6 Sol vary across groups: casual users favor FlowTool, while the other three groups favor GPT-5.6 Sol. Thus, the preference for FlowTool's visual quality is consistent across experience levels, while judgments of instruction following are more mixed.

\subsection{User Rationale Analysis}
\label{app:user_rationale}
\begin{figure}[p]
\centering
\includegraphics[width=\textwidth]{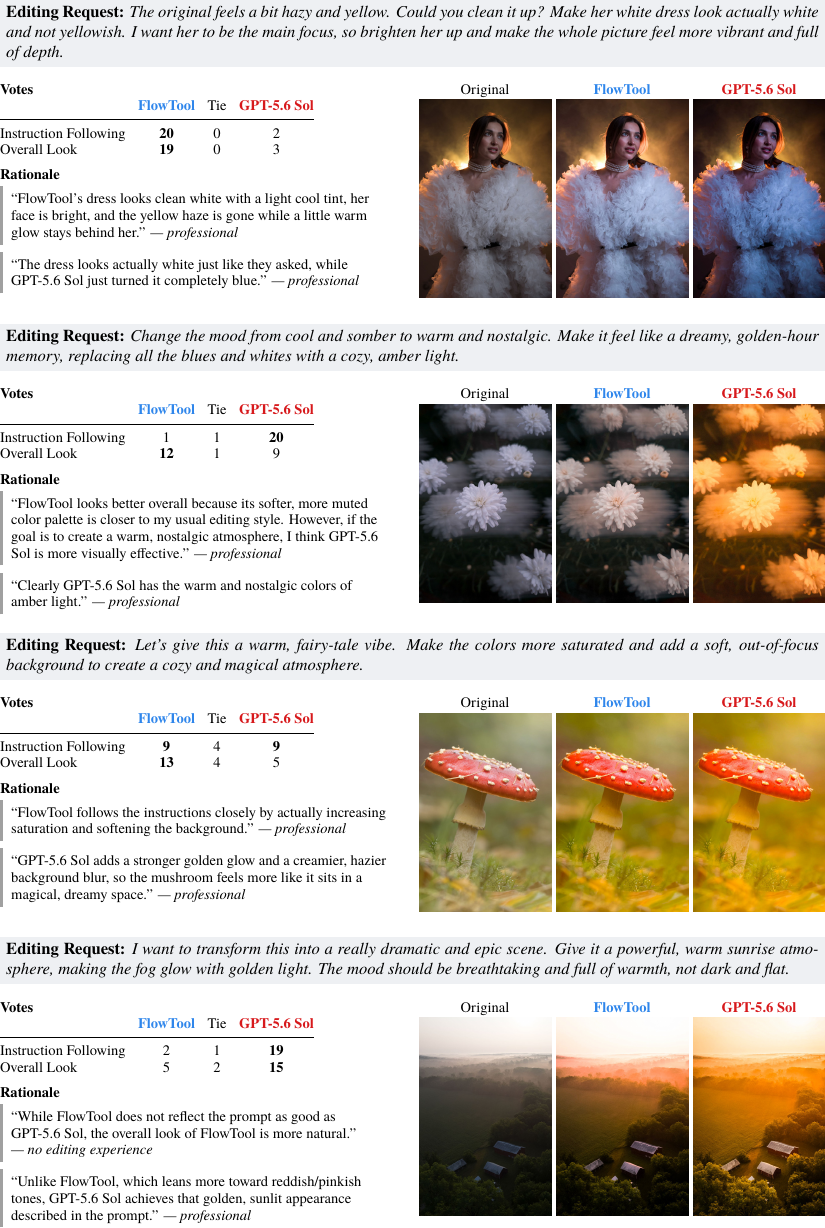}
\caption{Examples from the user study comparing FlowTool with GPT-5.6 Sol. Each example shows the editing request, the original image, and the two edits, with the number of participants (out of 22) who chose FlowTool, a tie, or GPT-5.6 Sol for each criterion (larger count in bold) and excerpts from participants' rationales, in which ``A"/``B" is replaced by the actual model name.}
\label{fig:user_study_examples_gpt}
\end{figure}

\begin{figure}[p]
\centering
\includegraphics[width=\textwidth]{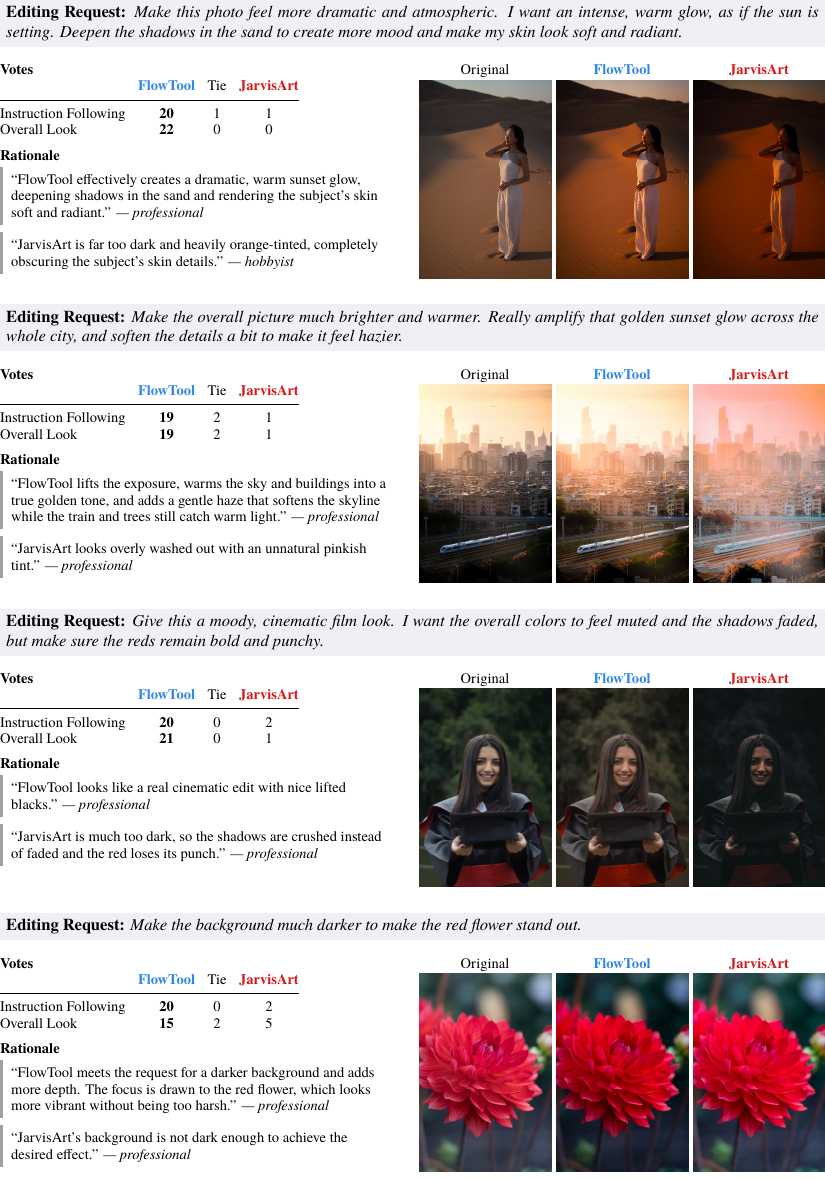}
\caption{Examples from the user study comparing FlowTool with JarvisArt. Each example shows the editing request, the original image, and the two edits, with the number of participants (out of 22) who chose FlowTool, a tie, or JarvisArt for each criterion (larger count in bold) and excerpts from participants' rationales, in which ``A"/``B" is replaced by the actual model name.}
\label{fig:user_study_examples_jarvis}
\end{figure}

Fifteen of the 22 participants provided written rationales for their judgments, resulting in 1,035 responses. We manually analyze these responses to identify recurring themes, including naturalness, tonal balance, and processing artifacts. This qualitative analysis supports the quantitative preference results and helps explain participants' choices. \cref{fig:user_study_examples_gpt,fig:user_study_examples_jarvis} present selected comparisons with vote counts and excerpts from participants' rationales.

\textbf{FlowTool achieve natural and balanced editing compared with GPT-5.6 Sol.}
Participants' rationales highlight FlowTool's ability to follow editing instructions while preserving natural colors, balanced tones, and image detail. Users who favor FlowTool often value these qualities over a stronger transformation that introduces unwanted visual changes. In contrast, criticisms of GPT-5.6 Sol frequently concern excessive color casts, oversaturation, or lost detail.

The examples in \cref{fig:user_study_examples_gpt} illustrate this distinction. When asked to remove a yellow cast, FlowTool produces a more natural result, whereas GPT-5.6 Sol over-corrects and turns the white dress blue; participants prefer FlowTool on both criteria. In the forest example, GPT-5.6 Sol's intense amber treatment receives more instruction-following votes, but FlowTool's more balanced edit is preferred for overall look. Even when the transformations are similar, as in the third example, participants favor FlowTool's cleaner appearance. GPT-5.6 Sol is preferred in the final example, where the request explicitly calls for a ``really dramatic and epic scene," and its stronger golden glow better matches that intent. These examples suggest that FlowTool's appeal lies in achieving the requested transformation while maintaining a natural and visually pleasing result, although its more restrained edits can be less suitable for intense stylization.

\textbf{FlowTool produces more faithful and visually pleasing edits than JarvisArt.}
Participants' rationales highlight major advantages for FlowTool in both fulfilling editing requests and producing visually pleasing results. In \cref{fig:user_study_examples_jarvis}, users criticize JarvisArt's excessive orange or pink casts, washed-out tones, crushed shadows, and insufficient background darkening. These problems compromise the requested appearance, obscure details, or fail to emphasize the intended subject. FlowTool, by comparison, is praised for achieving the requested warmth, mood, and subject emphasis while maintaining balanced colors, appropriate exposure, and visible detail.

These comments show why participants favor FlowTool on both criteria: its edits more successfully realize the intended transformation while preserving image quality. This qualitative evidence reinforces the strong and widespread preference for FlowTool that we discussed above.

\clearpage
\section{Details of Editing Tools}
\label{app:editing_tools}

\cref{tab:editing_tool_catalog} lists the editing parameters that form the continuous parameter space $\mathcal{T}$ of \cref{sec:problem_formulation}, grouped by the adjustment panel they belong to, together with their native ranges before normalization to $[-1,1]$.

{
\small
\setlength{\tabcolsep}{0pt}
\renewcommand{\arraystretch}{1.03}
\setlength{\LTpre}{4pt}
\setlength{\LTpost}{4pt}
\setlength{\LTleft}{\fill}
\setlength{\LTright}{\fill}

\begin{longtable}{
    @{}
    >{\raggedright\arraybackslash}p{0.31\linewidth}
    @{\hspace{0.020\linewidth}}
    >{\raggedright\arraybackslash}p{0.46\linewidth}
    @{\hspace{0.005\linewidth}}
    >{\centering\arraybackslash}p{0.13\linewidth}
    @{}
}
\caption{
Editing parameter catalog used by FlowTool.
}
\label{tab:editing_tool_catalog}\\

\toprule
\textbf{Parameter}
&
\textbf{Functional Description}
&
\textbf{Range}
\\
\midrule
\endfirsthead

\toprule
\textbf{Parameter}
&
\textbf{Functional Description}
&
\textbf{Range}
\\
\midrule
\endhead

\midrule
\multicolumn{3}{r}{\footnotesize Continued on next page}
\\
\endfoot

\bottomrule
\endlastfoot

\multicolumn{3}{l}{\textbf{Basic Adjustments}}
\\
\midrule

Temperature
&
Controls the overall color temperature along the blue--yellow axis.
&
$[2000,10000]$
\\

Tint
&
Controls the green--magenta color balance.
&
$[-150,150]$
\\

Exposure2012
&
Adjusts the overall exposure of the image.
&
$[-5,5]$
\\

Contrast2012
&
Controls the global tonal contrast.
&
$[-100,100]$
\\

Highlights2012
&
Adjusts the brightness of highlight regions.
&
$[-100,100]$
\\

Shadows2012
&
Adjusts the brightness of shadow regions.
&
$[-100,100]$
\\

Whites2012
&
Controls the brightest tonal regions and white point.
&
$[-100,100]$
\\

Blacks2012
&
Controls the darkest tonal regions and black point.
&
$[-100,100]$
\\

Texture
&
Adjusts medium-scale image detail and texture.
&
$[-100,100]$
\\

Clarity2012
&
Adjusts local midtone contrast.
&
$[-100,100]$
\\

Dehaze
&
Controls haze removal or addition.
&
$[-100,100]$
\\

Vibrance
&
Adjusts color intensity with emphasis on less-saturated colors.
&
$[-100,100]$
\\

Saturation
&
Adjusts the overall color saturation.
&
$[-100,100]$
\\

\midrule
\multicolumn{3}{l}{\textbf{Parametric Tone Curve}}
\\
\midrule

ParametricShadows
&
Adjusts the shadow region of the parametric tone curve.
&
$[-100,100]$
\\

ParametricDarks
&
Adjusts the dark tonal region of the parametric tone curve.
&
$[-100,100]$
\\

ParametricLights
&
Adjusts the light tonal region of the parametric tone curve.
&
$[-100,100]$
\\

ParametricHighlights
&
Adjusts the highlight region of the parametric tone curve.
&
$[-100,100]$
\\

\midrule
\multicolumn{3}{l}{\textbf{Detail and Noise Reduction}}
\\
\midrule

Sharpness
&
Controls overall image sharpening.
&
$[0,150]$
\\

LuminanceSmoothing
&
Controls luminance noise reduction.
&
$[0,100]$
\\

ColorNoiseReduction
&
Controls color noise reduction.
&
$[0,100]$
\\

DefringePurpleAmount
&
Controls removal of purple color fringing.
&
$[0,20]$
\\

\midrule
\multicolumn{3}{l}{\textbf{HSL: Hue}}
\\
\midrule

HueAdjustmentRed
&
Adjusts the hue of red colors.
&
$[-100,100]$
\\

HueAdjustmentOrange
&
Adjusts the hue of orange colors.
&
$[-100,100]$
\\

HueAdjustmentYellow
&
Adjusts the hue of yellow colors.
&
$[-100,100]$
\\

HueAdjustmentGreen
&
Adjusts the hue of green colors.
&
$[-100,100]$
\\

HueAdjustmentAqua
&
Adjusts the hue of aqua colors.
&
$[-100,100]$
\\

HueAdjustmentBlue
&
Adjusts the hue of blue colors.
&
$[-100,100]$
\\

HueAdjustmentPurple
&
Adjusts the hue of purple colors.
&
$[-100,100]$
\\

HueAdjustmentMagenta
&
Adjusts the hue of magenta colors.
&
$[-100,100]$
\\

\midrule
\multicolumn{3}{l}{\textbf{HSL: Saturation}}
\\
\midrule

SaturationAdjustmentRed
&
Adjusts the saturation of red colors.
&
$[-100,100]$
\\

SaturationAdjustmentOrange
&
Adjusts the saturation of orange colors.
&
$[-100,100]$
\\

SaturationAdjustmentYellow
&
Adjusts the saturation of yellow colors.
&
$[-100,100]$
\\

SaturationAdjustmentGreen
&
Adjusts the saturation of green colors.
&
$[-100,100]$
\\

SaturationAdjustmentAqua
&
Adjusts the saturation of aqua colors.
&
$[-100,100]$
\\

SaturationAdjustmentBlue
&
Adjusts the saturation of blue colors.
&
$[-100,100]$
\\

SaturationAdjustmentPurple
&
Adjusts the saturation of purple colors.
&
$[-100,100]$
\\

SaturationAdjustmentMagenta
&
Adjusts the saturation of magenta colors.
&
$[-100,100]$
\\

\midrule
\multicolumn{3}{l}{\textbf{HSL: Luminance}}
\\
\midrule

LuminanceAdjustmentRed
&
Adjusts the luminance of red colors.
&
$[-100,100]$
\\

LuminanceAdjustmentOrange
&
Adjusts the luminance of orange colors.
&
$[-100,100]$
\\

LuminanceAdjustmentYellow
&
Adjusts the luminance of yellow colors.
&
$[-100,100]$
\\

LuminanceAdjustmentGreen
&
Adjusts the luminance of green colors.
&
$[-100,100]$
\\

LuminanceAdjustmentAqua
&
Adjusts the luminance of aqua colors.
&
$[-100,100]$
\\

LuminanceAdjustmentBlue
&
Adjusts the luminance of blue colors.
&
$[-100,100]$
\\

LuminanceAdjustmentPurple
&
Adjusts the luminance of purple colors.
&
$[-100,100]$
\\

LuminanceAdjustmentMagenta
&
Adjusts the luminance of magenta colors.
&
$[-100,100]$
\\

\midrule
\multicolumn{3}{l}{\textbf{Split Toning}}
\\
\midrule

SplitToningShadowHue
&
Controls the hue applied to shadow regions.
&
$[0,359]$
\\

SplitToningShadowSaturation
&
Controls the saturation of the shadow toning.
&
$[0,100]$
\\

SplitToningHighlightHue
&
Controls the hue applied to highlight regions.
&
$[0,359]$
\\

SplitToningHighlightSaturation
&
Controls the saturation of the highlight toning.
&
$[0,100]$
\\

\midrule
\multicolumn{3}{l}{\textbf{Effects}}
\\
\midrule

PostCropVignetteAmount
&
Controls the strength and direction of the post-crop vignette.
&
$[-100,100]$
\\

VignetteAmount
&
Controls the overall vignette amount.
&
$[-100,100]$
\\

ShadowTint
&
Adjusts the color tint of shadow regions.
&
$[-100,100]$
\\

GrainAmount
&
Controls the amount of synthetic image grain.
&
$[0,100]$
\\

\midrule
\multicolumn{3}{l}{\textbf{Color Grading}}
\\
\midrule

ColorGradeMidtoneHue
&
Controls the color-grading hue of midtones.
&
$[0,360]$
\\

ColorGradeMidtoneSat
&
Controls the color-grading saturation of midtones.
&
$[0,100]$
\\

ColorGradeMidtoneLum
&
Adjusts luminance in the color-graded midtones.
&
$[-100,100]$
\\

ColorGradeShadowLum
&
Adjusts luminance in the color-graded shadows.
&
$[-100,100]$
\\

ColorGradeHighlightLum
&
Adjusts luminance in the color-graded highlights.
&
$[-100,100]$
\\

ColorGradeGlobalHue
&
Controls the global color-grading hue.
&
$[0,360]$
\\

ColorGradeGlobalSat
&
Controls the global color-grading saturation.
&
$[0,100]$
\\

ColorGradeGlobalLum
&
Controls global luminance within the color-grading module.
&
$[-100,100]$
\\

ColorGradeBlending
&
Controls blending between the color-graded tonal regions.
&
$[0,100]$
\\

\midrule
\multicolumn{3}{l}{\textbf{Camera Calibration}}
\\
\midrule

RedHue
&
Adjusts the hue of the red calibration primary.
&
$[-100,100]$
\\

RedSaturation
&
Adjusts the saturation of the red calibration primary.
&
$[-100,100]$
\\

GreenHue
&
Adjusts the hue of the green calibration primary.
&
$[-100,100]$
\\

GreenSaturation
&
Adjusts the saturation of the green calibration primary.
&
$[-100,100]$
\\

BlueHue
&
Adjusts the hue of the blue calibration primary.
&
$[-100,100]$
\\

BlueSaturation
&
Adjusts the saturation of the blue calibration primary.
&
$[-100,100]$
\\

\midrule
\multicolumn{3}{l}{\textbf{Incremental White Balance}}
\\
\midrule

IncrementalTemperature
&
Applies a relative adjustment to color temperature.
&
$[-100,100]$
\\

IncrementalTint
&
Applies a relative adjustment to tint.
&
$[-100,100]$
\\

\midrule
\multicolumn{3}{l}{\textbf{Noise Reduction Sub-Controls}}
\\
\midrule

LuminanceNoiseReductionDetail
&
Controls detail preservation during luminance noise reduction.
&
$[0,100]$
\\

LuminanceNoiseReductionContrast
&
Controls contrast preservation during luminance noise reduction.
&
$[0,100]$
\\

ColorNoiseReductionDetail
&
Controls detail preservation during color noise reduction.
&
$[0,100]$
\\

ColorNoiseReductionSmoothness
&
Controls smoothness of color noise reduction.
&
$[0,100]$
\\

\midrule
\multicolumn{3}{l}{\textbf{Grain Shape}}
\\
\midrule

GrainSize
&
Controls the size of the synthetic grain.
&
$[0,100]$
\\

GrainFrequency
&
Controls the spatial frequency of the synthetic grain.
&
$[0,100]$
\\

\midrule
\multicolumn{3}{l}{\textbf{Post-Crop Vignette Shape}}
\\
\midrule

PostCropVignetteMidpoint
&
Controls the midpoint of the post-crop vignette.
&
$[0,100]$
\\

PostCropVignetteFeather
&
Controls the softness of the vignette boundary.
&
$[0,100]$
\\

PostCropVignetteRoundness
&
Controls the shape and roundness of the vignette.
&
$[-100,100]$
\\

PostCropVignetteHighlightContrast
&
Controls highlight preservation within the vignette.
&
$[0,100]$
\\

\midrule
\multicolumn{3}{l}{\textbf{B\&W Channel Mixer}}
\\
\midrule

GrayMixerRed
&
Controls the contribution of red tones in grayscale conversion.
&
$[-100,100]$
\\

GrayMixerOrange
&
Controls the contribution of orange tones in grayscale conversion.
&
$[-100,100]$
\\

GrayMixerYellow
&
Controls the contribution of yellow tones in grayscale conversion.
&
$[-100,100]$
\\

GrayMixerGreen
&
Controls the contribution of green tones in grayscale conversion.
&
$[-100,100]$
\\

GrayMixerAqua
&
Controls the contribution of aqua tones in grayscale conversion.
&
$[-100,100]$
\\

GrayMixerBlue
&
Controls the contribution of blue tones in grayscale conversion.
&
$[-100,100]$
\\

GrayMixerPurple
&
Controls the contribution of purple tones in grayscale conversion.
&
$[-100,100]$
\\

GrayMixerMagenta
&
Controls the contribution of magenta tones in grayscale conversion.
&
$[-100,100]$
\\

\end{longtable}
}
\clearpage

\end{document}